%% file: FSM_benchmark_arXiv.tex
\PassOptionsToPackage{table,xcdraw}{xcolor}
\documentclass[]{article}

\usepackage{caption}
\usepackage{todonotes}
\usepackage{tabu}
\usepackage{longtable}
\usepackage{booktabs}
\usepackage{adjustbox}
\usepackage{array}
\usepackage{tabularx}
\usepackage{dcolumn}
\usepackage{lscape}
\usepackage[colorlinks=true, allcolors=teal]{hyperref}
\usepackage[nolist]{acronym}
\usepackage[doublespacing]{setspace}
\usepackage{natbib}
\usepackage[a4paper, left=2.3cm, right=2.3cm, top=2cm]{geometry}

\setcitestyle{authoryear}

\newcolumntype{R}[2]{%
	>{\adjustbox{angle=#1,lap=\width-(#2)}\bgroup}%
	l%
	<{\egroup}%
}
\newcommand*\rot{\multicolumn{1}{R{90}{.3em}}}

\newcommand*{\numFSMs}{58\ }

\begin{document}

\title{Filter Methods for Feature Selection in Supervised Machine Learning Applications---Review and Benchmark}

\author{Konstantin Hopf, Sascha Reifenrath}
\date{%
\textit{University of Bamberg \\
Chair of Information Systems and Energy Efficient Systems\\
Kapuzinerstr. 16, D-96047 Bamberg\\
contact: \href{mailto:konstantin.hopf@uni-bamberg.de}{konstantin.hopf@uni-bamberg.de}}\\[2ex]%
\today}

\maketitle

\begin{abstract}
  The amount of data for machine learning (ML) applications is constantly growing. Not only the number of observations, especially the number of measured variables (features) increases with ongoing digitization. Selecting the most appropriate features for predictive modeling is an important lever for the success of ML applications in business and research. Feature selection methods (FSM) that are independent of a certain ML algorithm---so-called filter methods---have been numerously suggested, but little guidance for researchers and quantitative modelers exists to choose appropriate approaches for typical ML problems. This review synthesizes the substantial literature on feature selection benchmarking and evaluates the performance of \numFSMs methods in the widely used R environment. For concrete guidance, we consider four typical dataset scenarios that are challenging for ML models (noisy, redundant, imbalanced data and cases with more features than observations). Drawing on the experience of earlier benchmarks, which have considered much fewer FSMs, we compare the performance of the methods according to four criteria (predictive performance, number of relevant features selected, stability of the feature sets and runtime). We found methods relying on the random forest approach, the double input symmetrical relevance filter (DISR) and the joint impurity filter (JIM) were well-performing candidate methods for the given dataset scenarios.
\end{abstract}

\begin{acronym}
	\acro{AUC}{Area Under ROC Curve}
	\acro{CRAN}{Comprehensive R Archive Network}
	\acro{FSM}{Feature Selection Method}
	\acro{ML}{Machine Learning}
	\acro{NB}{Na\"ive Bayes}
	\acro{RF}{Random Forest}
	\acro{ROC}{Receiver Operating Characteristic}
	\acro{SVM}{Support Vector Machine}
\end{acronym}

\section{Introduction}

\ac{ML} 
is a core technology of artificial intelligence \citep{russell_artificial_2016}, which has rendered current outstanding applications of speech processing, image recognition, self-driving cars, and others possible. 
Next to these remarkable technological developments, \ac{ML} and related techniques also alter the data processing in business and research. \citet{davenport_competing_2007} were right in their prediction that firms today build competitive advantages through analytics \citep{kraus_deep_2020, muller_effect_2018, wu_data_2019}. Firms do so by employing effective data analysis and \ac{ML} to extract relevant information faster and make more informed decisions. Likewise, researchers benefits from \ac{ML}-related techniques allowing them to explore new patterns in data and thus to discover new insights or theories \citep{berente_research_2019, shmueli_predictive_2011}.

\subsection{Feature selection as a core activity in pursuing ML data analysis}
A core task in setting up \ac{ML} is to identify the relevant features from the wide variety of input data. Feature selection mitigates the ``curse of dimensionality" problem \citep{jain_feature_1997, keogh_curse_2011}, which is known to lower the quality of ML predictions. It allows learning less complex models, which are less likely to be over-fitted to the data, lowers training times, and requires less storage space of data and models \citep{guyon_introduction_2003, kudo_comparison_2000}. Feature selection, thus, reduces computing costs of data analyses \citep{li_feature_2017}. 

The necessity of feature selection has increased in recent years: Whereas in the last decades, a number of 50 to 100 features was called a ``large" feature set \citep[p. 25]{kudo_comparison_2000}, today we are confronted with hundreds or even thousands \citep{hua_performance_2009} of features that are measured on every entity. Further digitization will aggravate this, as it allows measuring even more variables in business operations through connected devices and novel sensors \citep{feng_big_data_production_2017}. Although some studies argue that modern neural networks (i.e., deep learning) can be applied directly to raw data without requiring feature selection \citep{kraus_deep_2020, lecun_deep_2015}, it seems that  their predictive performance can still be improved in some areas by selecting suitable features before applying \ac{ML} \citep{semwal_robust_2017, borisov_cancelout_2019}. Moreover, not for all application areas of \ac{ML} are deep learning approaches the most suitable ones \citep{fernandez-delgado_we_2014}. Thus, feature selection remains an essential task in setting up \ac{ML}.

When \ac{ML} is adopted by wider ranges of applications, related data analysis techniques, like \acp{FSM}, similarly have to cope with the challenges of \ac{ML} use \citep{lheureux_machine_2017, bosu_taxonomy_2013, kaur_systematic_2019, branco_survey_2016}. Those challenges are, for example, missing values, low data quality (e.g., redundancies and noise in the data), and small datasets (e.g., few observations but many variables, and the infrequent occurence of interesting events, which leads to imbalanced datasets). 

\subsection{The multitude of available feature selection methods complicates their choice}

Several software libraries offer an extensive number of \acp{FSM} to researchers, \ac{ML} users, and data scientists with ready-to-use interfaces \citep{corelearn, mlr, fselector, praznik, li_feature_2017}. 
With regard to the feature selection strategy, literature typically classifies the approaches into three categories \citep{blum_selection_1997, guyon_introduction_2003, li_feature_2017}: Wrapper, embedded, and filter methods. Wrappers utilize \ac{ML} models that are trained and tested to assess the predictive power of single feature sets and to find feature sets with a local maximum of the model performance. Embedded methods are bound to specific ML algorithms. Filter methods, by contrast, obtain a score for each feature based on the dataset, independent of the applied \ac{ML} algorithm. They either select a set of the best features or they rank features based on their estimated merit based on this scoring. 

Due to the large number of available \acp{FSM}, it is difficult to choose the most appropriate method for a given \ac{ML} problem. There is a series of reviews that describe the various available \acp{FSM} available \citep{li_feature_2017, bommert_benchmark_2020, chandrashekar_survey_2014, guyon_introduction_2003, liu_feature_2010, saeys_review_2007}. More or less comprehensive \ac{FSM} benchmark studies also investigated how well existing \acp{FSM} behave---in combination or without \ac{ML} algorithms. This research found that filter methods show a comparable good performance in comparison to other approaches \citep{haury_influence_2011, saeys_robust_2008}. Filters also usually need less computational resources than wrapper methods and the selection of features does not depend on the choice of the learning algorithm or the evaluation metric. Given that filter methods evaluate features independently of any particular classifier, the extracted features are “generic, having incorporated few assumptions” \citep{brown_conditional_2012}. 

When focusing on studies that compare filtering methods for feature selection, we notice that, while there are some comparative studies, they provide only a partial overview. One reason is that existing studies consider different sets of \acp{FSM} and that many of them use domain-specific datasets (e.g., bioinformatics, text analysis, energy retailing). Another reason is that these works often only focus on the high dimensionality of data, but not other dataset properties, like noise, skewed class distributions or redundant features. Finally, the studies come to different conclusions. This gives rise to the assumption that the no free lunch theorem \citep{wolpert_lack_1996} applies for feature selection as well. This theorem states, in basic words, that different optimization problem strategies perform equally well when averaged over all possible problems \citep{gomez_empirical_2016}. Yet, when applying ML and related methods, it is of little relevance that there is no all-surpassing method in theory. In the data analysis practice, usually not all theoretical dataset problems occur at the same time, but analysts are often confronted with few but severe dataset problems (noisy data, skewed data, small datasets, etc.), which are manifested to varying degrees. Therefore, it is rather important to know those \acp{FSM} which have turned out to be the most helpful for specific problem classes.

\subsection{Objectives of this work}

Our work extends previous efforts to obtain actionable knowledge for researchers and \ac{ML} through benchmark studies of \ac{ML} algorithms \citep{baumann_comparative_2019, fitzpatrick_empirical_2016, gartner_machine_2015} and ML platforms \citep{roy_performance_2019}. Thus, we seek to address the lack of an overview to effective FSMs in particular problem classes. In doing so, our paper makes two contributions. First, we provide an overview to the current state of benchmark experiments on \acp{FSM} and conduct a meta-review of 33 studies. This overview will help \ac{ML} users as a reference to more specific analyses in already existing benchmark studies. We also derive a synthesized benchmark method, which we apply in the latter part of our study. Second, we report on an extensive benchmark study involving \numFSMs available in the widely used statistical programming environment GNU R, which represents---to the best of our knowledge---the most comprehensive benchmark of \acp{FSM} so far. Our experiment builds on learnings from the earlier benchmark studies. Thereby, we evaluated the performance of \acp{FSM} in four dataset scenarios that are typical \ac{ML} problems:  
We use two problems from the area of \textit{low data quality} \citep{lee_machine_2020, bosu_taxonomy_2013, alonso_challenges_2015}, namely (I) noise in the data (class and attribute noise) and (II) redundancy.
Besides, we use two problems from the area of \textit{small datasets} \citep{lheureux_machine_2017, bosu_taxonomy_2013}, namely (III) few observations with many features and (IV) few observations of a minority class (i.e., imbalanced classes).
Moreover, we examine whether methods that are described as particularly well suited for specific dataset problems (e.g. noisy and imbalanced data) also perform better than other methods in these scenarios. 

With the no free lunch theorem in mind, our benchmark study cannot be expected to lead to a selection of methods that always outperform others. Rather than aiming for a general recommendation, we attempt to provide assistance for concrete dataset scenarios and focus on our benchmark analysis on four scenarios.

\subsection{Structure of the paper}

This paper proceeds with our survey of earlier \acp{FSM} benchmark studies. \autoref{sec:fsm_benchmark_inR} continues with presenting our benchmark of currently available \acp{FSM} in the GNU R environment. Thereby, we compile a holistic benchmark approach for \acp{FSM} as a result of the literature survey and apply it to \numFSMs \acp{FSM} in the widely-used GNU R environment. We report the performance of each method for four dataset challenges. We conclude this paper with a summary and discussion our results.

\pagebreak
\section{Background}

Several software libraries offer an extensive number of \acp{FSM} to researchers, \ac{ML} users, and data scientists with ready-to-use interfaces \citep{corelearn, mlr, fselector, praznik, li_feature_2017}. This section gives an overview to \acp{FSM} and reviews earlier benchmark studies that compare different approaches.

\subsection{Overview to FSMs}

With regard to the feature selection strategy, literature typically classifies the approaches into three categories \citep{blum_selection_1997, guyon_introduction_2003, li_feature_2017}: Wrapper, embedded, and filter methods. Wrappers utilize \ac{ML} models that are trained and tested to assess the predictive power of single feature sets and to find feature sets with a local maximum of the model performance. Embedded methods are bound to specific ML algorithms. Filter methods, by contrast, obtain a score for each feature based on the dataset, independent of the applied \ac{ML} algorithm. They either select a set of the best features or they rank features based on their estimated merit based on this scoring. 

Due to the large number of available \acp{FSM}, it is difficult to choose the most appropriate method for a given \ac{ML} problem. There is a series of reviews that describe the various available \acp{FSM} available \citep{li_feature_2017, bommert_benchmark_2020, chandrashekar_survey_2014, guyon_introduction_2003, liu_feature_2010, saeys_review_2007}. More or less comprehensive \ac{FSM} benchmark studies also investigated how well existing \acp{FSM} behave---in combination or without \ac{ML} algorithms. This research found that filter methods show a comparable good performance in comparison to other approaches \citep{haury_influence_2011, saeys_robust_2008}. Filters also usually need less computational resources than wrapper methods and the selection of features does not depend on the choice of the learning algorithm or the evaluation metric. Given that filter methods evaluate features independently of any particular classifier, the extracted features are “generic, having incorporated few assumptions” \citep{brown_conditional_2012}. 

When focusing on studies that compare filtering methods for feature selection, we notice that, while there are some comparative studies, they provide only a partial overview. One reason is that existing studies consider different sets of \acp{FSM} and that many of them use domain-specific datasets (e.g., bioinformatics, text analysis, energy retailing). Another reason is that these works often only focus on the high dimensionality of data, but not other dataset properties, like noise, skewed class distributions or redundant features. Finally, the studies come to different conclusions. This gives rise to the assumption that the no free lunch theorem \citep{wolpert_lack_1996} applies for feature selection as well. This theorem states, in basic words, that different optimization problem strategies perform equally well when averaged over all possible problems \citep{gomez_empirical_2016}. Yet, when applying ML and related methods, it is of little relevance that there is no all-surpassing method in theory. In the data analysis practice, usually not all theoretical dataset problems occur at the same time, but analysts are often confronted with few but severe dataset problems (noisy data, skewed data, small datasets, etc.), which are manifested to varying degrees. Therefore, it is rather important to know those \acp{FSM} which have turned out to be the most helpful for specific problem classes.
Yet, with the no free lunch theorem in mind, our benchmark study cannot be expected to lead to a selection of methods that always outperform others. Rather than aiming for a general recommendation, we attempt to provide assistance for concrete dataset scenarios and focus on our benchmark analysis on four scenarios.

\subsection{Earlier benchmark studies for FSMs}
\label{sec:related_work}

To obtain an overview to existing benchmark studies on \acp{FSM}, we queried premier scientific databases (e.g., ACM digital library, Google Scholar, EBSCO Business Source Ultimate). We used a combination of the search terms \textit{feature selection} or \textit{variable selection} with one of the words \textit{benchmark}, \textit{comparison}, \textit{experiment}. Based on the title, abstract, keywords, and if necessary the full-text, we decided whether the respective paper is relevant to our review. Based on the initial set of studies, we also included studies that the most recent articles referenced (backward search). We found several studies that suggest new \acp{FSM} and compare their method(s) to existing ones, merely to demonstrate their superiority. We included such studies only when they considered more than five other methods in their comparison (as, for example, \citet{brown_conditional_2012} and \citet{meyer_informationtheoretic_2008} did). In total, our literature search resulted in a collection of 33 studies.

We reviewed all identified studies and pulled out the number and type of considered \acp{FSM}, the used data for the benchmark (i.e., number of datasets, if they are real or artificial data, the number of features and observations), and the application subject (e.g., general \ac{ML}, biomedical, text analysis). We also analyzed which specific dataset challenges the works are concerned with at their core (i.e., more features than observations, imbalanced class distributions, noisy data, redundant features, certain variable types). Finally, we surveyed which evaluation metrics the papers used in evaluating \acp{FSM}. 

\autoref{tab:literature} shows the results of our review with the study characteristics.
Except from \citet{bommert_benchmark_2020}, who compare 22 methods with 16 datasets and \citet{hopf_predictive_2019}, who compares 43 methods with one dataset from electricity retailing, studies include only some, at most eleven, feature filters in their benchmark. Many works use only a limited set of performance criteria and mostly focus on prediction accuracy alone. 
\pagebreak

			
	\begin{tiny}
		\rowcolors{2}{gray!15}{white}
		\begin{longtable}{p{1.9cm}p{1.5cm}p{.7cm}rr|lllllp{0.9cm}p{0.8cm}llllll}
			\caption{Earlier FSM benchmark studies (in chronological order)}\label{tab:literature}\\
			\hiderowcolors
			\toprule
			&  & \multicolumn{3}{l|}{Dataset} & \multicolumn{5}{l}{Dataset challenge} &&& \multicolumn{6}{l}{Metrics} \\
			\cmidrule{4-6}
			\cmidrule{7-11}
			\cmidrule{13-18}
			Ref. & Subject & 	\rot{Number of datasets$^a$} & 
			\multicolumn{1}{ R{90}{-2.2em}}{Max num. of Features} &
			\multicolumn{1}{ R{90}{-2.2em}}{Training samples} &
			\multicolumn{1}{|R{90}{0.2em}}{High-dimensional} &
			\multicolumn{1}{ R{90}{0.2em}}{Class imbalance}  &
			\multicolumn{1}{ R{90}{0.2em}}{Noisy data} & 
			\multicolumn{1}{ R{90}{0.2em}}{Redundant features} & 
			\multicolumn{1}{ R{90}{0.2em}}{Variable type (cat., num.)} &
			\multicolumn{1}{ R{90}{0.3em}}{Classifier$^b$}&
			\multicolumn{1}{ R{90}{0.3em}}{FSMs in benchmark$^c$}& 
			\multicolumn{1}{ R{90}{0.3em}}{Predictive performance$^d$}&
			\multicolumn{1}{ R{90}{0.3em}}{Stability} & 
			\multicolumn{1}{ R{90}{0.2em}}{Correct chosen features}& 
			\multicolumn{1}{ R{90}{0.2em}}{Number of selected features} &
			\multicolumn{1}{ R{90}{0.2em}}{Comparison of F/W/E} & 
			\multicolumn{1}{ R{90}{0.2em}}{Runtime performance}\\
			\midrule
			\showrowcolors
			\cite{dash_feature_1997} & General ML&
			3 (S)	&       12 &      122 &  &  & X& X&  &NB, DT    &  8 (F, W)&  &  & X&  & X&  \\

			\cite{kudo_comparison_2000}& Diverse domains&
			8 (R, S)&       65 &  1{,}000 &  &  &  &  &  &IB        &  9 (W)   & A&  & & X&  & X\\

			\cite{liu_comparative_2002}	& Biomedical &
			2 (R)	& 15{,}154 &      327 & X&  &  &  &  &IB, DT, NB&  6 (F)   & A&  & & X&  &  \\
			
			\cite{forman_extensive_2003} & Text classification&
			1 (R)	&       16 &      229 & X&  &  &  &  &NB, DT, LR, SVM& 11 (F) & AFR&  & & X&  &  \\
			
			\cite{reunanen_overfitting_2003} & Diverse datasets &
			7 (R)	&      112 &  1{,}000 &  &  &  & X&  &IB        &  2 (W)   & A&  & & X&  &  \\
			
			\cite{robnik-sikonja_experiments_2003} & General ML&
			12(S)	&       25 &  1{,}000 &  & X&  &  &  &-         & 13 (F)   &  &  & X&  &  &  \\
			
			\cite{hall_benchmarking_2003} & Diverse domains&
			18 (R)	&  1{,}555 & 29{,}598 & X&  &  &  &  &DT, NB    &  6 (F, W)& A&  &  & X& X& X\\
			
			\cite{inza_filter_2004} & Biomedical data &
			2 (R)	&  7{,}129 &       72 &  &  &  &  &  &IB, NB, DT&  7 (F, W)& A&  & &  & X&  \\
			
			\cite{peng_feature_2005} & Medical data and handwritten digits&
			4 (R, S)&  9{,}703 &  2{,}000 & X&  &  &  & X&NB, SVM, LDA &  3 (F)& A&  &  & X&  &  \\
			
			\cite{lai_comparison_2006} & Biomedical data &
			7 (R)	&  5{,}963 &      102 & X&  &  & X&  &SVM, Fisher, IB&  7 (F, W, E)& A&  & &  & X&  \\
			
			\cite{sanchez_filter_2007} & General ML &
			2 (S)	&      200 &      400 & X&  & X& X&  &        - &  4 (F)   &  &  & X&  &  &  \\
			
			\cite{kalousis_stability_2007} & Biomedical data &
			11 (R)	&  4{,}031 &  7{,}977 & X&  &  &  &  &        - &  5 (F)   &  & X& &  &  &  \\
			
			\cite{saeys_robust_2008} & Biomedical data &
			6 (R)	& 15{,}154 &      322 &  &  &  &  &  &RF, IB, SVM&  4 (F, E)& A& X& &  & X&  \\
			
			\cite{meyer_informationtheoretic_2008} & Biomedical data &
			12(R, S)&       15 &      308 &  &  & X&  &  &SVM, IB   &  6 (F)   & A&  & X&  &  & X\\
			
			\cite{dougherty_performance_2009} & Biomedical data &
			2 (S,R)	&       20 &      295 & X&  & X&  &  &LDA, IB   &  4 (F, W)& A&  &  & X& X&  \\
			
			\cite{hua_performance_2009} & Biomedical data &
			1 (S)	&       20 &      180 & X&  & X&  &  &IB, SVM   &  8 (F,W) & A&  &  & X& X&  \\
			
			\cite{haury_influence_2011} & Biomedical data &
			4 (R)	&      100 &      286 & X&  &  &  &  &NC, SVM, NB, LDA&8 (F, W, E)& R& X& &  & X&  \\
			
			\cite{alelyani_dilemma_2011} & General ML&
			5 (R)	& 22{,}283 & 718      &  &  &  &  &  & -        & 5        &  & X& &  & X&  \\
			
			\cite{brown_conditional_2012} & Diverse domains&
			15 (R)	&      256 &  6{,}435 & X&  &  &  &  &IB        &  9 (F)   & A& X& &  &  &  \\
			
			\cite{bolon-canedo_review_2013} & General ML&
			11(S, R)&  4{,}060 &  2{,}915 & X&  & X& X&  &DT, SVM, NB, IB&12 (F, W, E)& A&  & X&  & X&  \\
			
			\cite{chandrashekar_survey_2014} & Electrical engineering &
			6 (R)	&       34 &     n.a. &  &  &  &  &  &SVM, ANN  &  4 (F, W)& A&  & &  & X&  \\
			
			\cite{bolon-canedo_review_2014} & Biomedical data &
			9 (R)	&       24 &      254 & X&  &  &  &  &DT, SVM, NB&  7 (F)   & AS&  & &  &  &  \\
			
			\cite{aphinyanaphongs_comprehensive_2014} & Text&
			20 (R)	&  7{,}549 & 11{,}162 &  &  &  &  &  &SVM, LR, NB, AdaBoost&   F, W& R&  & &  &  &  \\
			
			\cite{khoshgoftaar_comparing_2015} & Software defects&
			3 (R)	&      209 &      661 & X& X&  &  &  &SVM, ANN  &5 (F, W, E)& R&  & &  & X&  \\
			
			\cite{xue_comprehensive_2015} & Diverse domains&
			24 (R)	&      617 &  6{,}435 &  &  &  &  & X&IB, SVM, DT, NB& 10 (F, W)& A&  &  & X& X& X\\
			
			\cite{adams_empirical_2017} & General ML &
			10		&       64 &       20 &  &  &  &  &  &DT, IB, RF&  5 (F, W)& A&  & &  & X&  \\
			
			\cite{liu_cost-sensitive_2018} & General ML &
			9 (S)	&      120 &  1{,}650 &  & X&  &  &  &SVM       &  9 (F)   & AFR&  & & X&  &  \\
			
			\cite{darshan_performance_2018} & Malware data&
			2 (R)	&  1{,}877 &      600 & X& X&  &  &  &LR, RF, DT&  4 (F)   & AS&  & &  &  &  \\
			
			\cite{wah_feature_2018} & Biomedical data&
			5 (S)	&       57 &  1{,}000 & X&  & X&  &  &LR        &  8 (F, W)& ASR& & X&  & X&  \\
			
			\cite{nogueira_stability_2018} & General ML &
			1 (S)	&      100 &  2{,}000 &  &  &  &  &  & LR       & 2 (E)    &  &X &  &  &  &  \\
			
			\cite{hopf_predictive_2019} & Energy retailing &
			1 (R)	&      308 &      451 &  &  &  &  &  & LR       & 43 (F)   &A &X &  &X &  & X\\
			
			\cite{bommert_benchmark_2020} & General ML&
			16 (R, S)& 22{,}283 &  7{,}000 	  & X&  &  & X& X&IB, LR, SVM& 22 (F)  & A& X& &  &  & X\\

			\cite{kou_2020_evaluation} & Text &
			10 (R)	& 27{,}973	&  1{,}000 	  & X&  &  &  &  & SVM, IB, NB & 10 (F)& AFR& X&  &   &   & \\
			
			\midrule
			\multicolumn{2}{l}{This study} & 
			13 (S)	& 1{,}000 &  2{,}500 	  & X& X& X& X&  & SVM, RF, NB & \numFSMs (F)  & R& X& X& X &  & X\\
			\hiderowcolors
			\midrule
			\multicolumn{17}{l}{$a$) S: Synthetic, R: Real} \\
			\multicolumn{17}{p{13cm}}{$b$) 
				ANN: Artificial Neural Networks, 
				DT: Decision trees (e.g., C4.5), 
				IB: Instance based learners (e.g., k nearest neighbor, nearest centroid, IB1),
				LDA: Linear Discriminant Analysis,
				LR: Logistic Regression,
				NB: Na\"ive Bayes, 
				RF: Random Forest, 
				SVM: Support Vector Machines} \\
			\multicolumn{17}{l}{$c$) F: Filter, W: Wrapper, E: Embedded}\\
			\multicolumn{17}{p{13cm}}{$d$) 
				A: Accuracy or error rate, 
				F: F-Measure, 
				R: ROC or AUC, 
				S: sensitivity and specificity} \\
			
		\end{longtable}
	\end{tiny}

In detail, we identified three streams of literature. The first one \citep[and others]{dash_feature_1997, kudo_comparison_2000, reunanen_overfitting_2003} compares wrapper and filter methods on the basis of different datasets in terms of classification accuracy. From these studies, no single method was found to be superior to others. A major drawback of these works is that \ac{FSM} benchmarks only considered the classification performance. However, some evidence suggests that filter methods show a comparable good performance in comparison to other approaches, or can even outperform more computationally complex wrapper or embedded methods \citep{haury_influence_2011, saeys_robust_2008}.
The second stream compares filter methods with regard to the stability of their selection and whether they identify relevant features \citep{alelyani_dilemma_2011, kalousis_stability_2007, robnik-sikonja_theoretical_2003, sanchez_filter_2007}. While these evaluation results are independent of any classification algorithm, the actual performance gains of the \acp{FSM} for classification tasks cannot be drawn.
A third and more recent stream suggestst to also include the stability of \acp{FSM} (similarity of selected feature sets by slightly varied data caused by randomized subsampling) and the algorithm runtimes in the comparison of \acp{FSM} \citep{haury_influence_2011, saeys_robust_2008, bommert_benchmark_2020, hopf_predictive_2019}. 

To sum up our review, studies with comprehensive benchmark experiments (i.e., considering multiple criteria such as predictive performance, stability, and runtime) cover only few \acp{FSM}. Those studies with larger numbers of \acp{FSM} do not focus on different dataset challenges that we, for example, selected for our study. 
Furthermore, the fields of applications are dominated by studies in the area of bioinformatics with microarray, mass spectrometry datasets, etc. with domain-specific prediction problems \citep{bolon-canedo_review_2014, dougherty_performance_2009, haury_influence_2011, hua_performance_2009, kalousis_stability_2007, lai_comparison_2006, saeys_review_2007, saeys_robust_2008}.


\section{Benchmark of FSMs in GNU R}
\label{sec:fsm_benchmark_inR}


Given the suggestions and the limitations of earlier benchmark studies, we developed a holistic benchmark approach for \acp{FSM} that we present below. This benchmark method draws on approaches of earlier works and considers four evaluation criteria.

For the benchmark of the \acp{FSM} we first generated synthetic data covering four scenarios of dataset challenges (noisy and redundant data, few observations with many features and imbalanced classes). Then, we applied the \acp{FSM} to the data under these scenarios and computed evaluation metrics based on the outcomes. Finally, we analyzed the results statistically. 

\subsection{Evaluation scenarios and datasets}
\label{sec:datasets}

According to the four scenarios of datset challenges, we created 13 artificial datasets that form the base of our benchmark experiments (see \autoref{tab:datasets}).  For each dataset, we specified which features are relevant and varied the dataset characteristics like the class distribution, the number of observations, the number of (ir-)relevant, and redundant features. For dataset generation, we used the scikit-learn method ``make\_classification" \citep{pedregosa_scikit-learn_2011}, which is an adaption of Guyon’s (\citeyear{guyon_introduction_2003}) algorithm that generates binary classification problems with numerical features of a continuous value range. The method first generates relevant features using clusters with normally distributed points around corners of a multidimensional hypercube whose dimension equals to the specified number of features. For each class, it generates the same number of clusters. Then, the method adds interdependence between features and redundant features through linear combinations of relevant features together with randomly generated noise. Finally, the values of the features are shifted and randomly re-scaled.
Datasets that were generated with this method were used in the NIPS 2003 Feature Selection Challange and were used in earlier benchmark studies for FSM \citep{bommert_benchmark_2020, bolon-canedo_review_2013, nogueira_stability_2018}. 

\begin{table}[h]
	\caption{Summary of the datasets and their properties}\label{tab:datasets}
	\begin{small}
		\rowcolors{2}{gray!15}{white}
		\begin{tabu}{Xrrrrrrr}
			\toprule
			&& \multicolumn{3}{l}{Features} & \multicolumn{2}{l}{Noise} &\\
			\cmidrule{3-5}
			\cmidrule{6-7}
			Name & Observations & Total & Relevant & Redundant & Class & Attribute & Minority class\\
			\midrule
			Baseline 	& 2{,}500	&  100	& 10	&  0 & 0 & 0 & 0.5\\
			ClassNoise\_1 & 2{,}500	&  100	& 10	&  0 & 0.1 & 0 & 0.5\\
			ClassNoise\_2 & 2{,}500	&  100	& 10	&  0 & 0.2 & 0 & 0.5\\
			ClassNoise\_3 & 2{,}500	&  100	& 10	&  0 & 0.3 & 0 & 0.5\\
			AttNoise\_1	& 2{,}500	&  100	& 10	&  0 & 0 & 0.1 & 0.5\\
			AttNoise\_2	& 2{,}500	&  100	& 10	&  0 & 0 & 0.2 & 0.5\\
			AttNoise\_3	& 2{,}500	&  100	& 10	&  0 & 0 & 0.3 & 0.5\\
			Redundant\_1	& 2{,}500	&  100	& 10	& 10 & 0 & 0 & 0.5\\
			Redundant\_2	& 2{,}500	&  100	& 10	& 20 & 0 & 0 & 0.5\\
			Imbalanced\_1	& 2{,}500	&  100	& 10	& 10 & 0 & 0 & 0.2\\
			Imbalanced\_1	& 2{,}500	&  100	& 10	& 10 & 0 & 0 & 0.1\\
			Dimensionality\_1	& 2{,}500	&  500	& 15	&  0 & 0 & 0 & 0.5\\
			Dimensionality\_2	&   500	& 1000	& 30	&  0 & 0 & 0 & 0.5\\
			\bottomrule
			
		\end{tabu}
	\end{small}
\end{table}

We used the dataset \textit{Baseline} for all later analyses and combine this dataset without special characteristics with the other datasets. For \textit{ClassNoise\_*}, we added noise to the dependent variable of the baseline dataset in that we swapped the values of one class with the others for the degree of noise \citep{zhu_class_2004}. For \textit{AttNoise\_*}, we added noise to the features by adding a random number to the original value of each feature. 
The datasets with noise were only used during feature selection and classifier training, the baseline dataset was used for evaluation \citep{saez2013tackling}. 
We used the datasets \textit{Redundant\_*} to investigate the suitability of filter methods in the presence of redundant features. The two datasets each contained ten relevant features and ten and twenty redundant features, respectively. 
The datasets \textit{Imbalanced\_*} had a class distribution where the smaller class has a relative frequency of 0.2 and 0.1 respectively. This allowed to evaluate the suitability of filter methods on imbalanced datasets.
\textit{Dimensionality\_*} were datasets with many features compared to the number of observations (i.e., large $p$ small $n$ problem). In \textit{Dimensionality\_1}, we increased the number of features to 500 and in \textit{Dimensionality\_2}, we set the number of features higher than the number of observations, which is a typical problem of small datasets. For each dataset, we randomized the order of columns and rows to avoid unintended disturbance of our experiments.

\subsection{Measurement of performance metrics}

We used four performance metrics to evaluate the \acp{FSM}. Each metric is explained below.

\subsubsection{Predictive performance}


All earlier studies that evaluate \acp{FSM} in combination with \ac{ML} algorithms consider the predictive performance, mostly as the core criterion (see \autoref{tab:literature}). Thereby, a wide variety of \ac{ML} algorithms are used to obtain predictions. Most frequently, \ac{SVM} (17 studies), instance-based learners (15 studies), \ac{NB} (12 studies) were used.

Our study employs three well-known classification algorithms that belong to different categories of learners\footnote{Another quite popular class of \ac{ML} algorithms is neural networks. Following our argumentation in the introduction, we do not consider them in our study. In addition to the already reviewed organizational issues, two reasons led us to focus on other \ac{ML} methods: First, neural networks depend on large amounts of training data. Since business data analysis often has to work with small data sets, the use of neural networks is often limited. Second, in earlier benchmark experiments with real data sets \citep{fernandez-delgado_we_2014}, neural networks did not clearly outperform the predictive performance of RF and SVM.}: \ac{NB}, \ac{RF}, and \ac{SVM}.
\ac{NB} is a classifier based on the application of Bayes' theorem and assumes, for simplicity, that the features are stochastically independent of each other \citep{dougherty2013estimating}. Based on the feature values, the classifier calculates conditional probabilities for each new observation for each class and assigns the class with the highest probability to the observation. We use the implementation of \citet{e1071}. 
The \ac{RF} algorithm composes several decision trees whose results are aggregated to determine the final outcome. This process of training multiple versions of a classifier on bootstrap samples and then aggregating them is called bootstrap aggregation or ``bagging" \citep{breiman_random_2001}. To reduce the correlation of the resulting trees, \ac{RF} uses just a random sample of the feature sets at every split in the tree. We use the implementation of \citet{randomForest} with the standard parameter of 500 trees.
\acp{SVM} \citep{vapnik_statistical_1998} are a class of learning algorithms that search for a hyperplane in the feature space as a decision boundary that separates the data points with largest possible distance. The complete separation of data points by a hyperplane is only possible in the case of linearly separable data. In the case of nonlinearly separable data, a kernel function is used to  transform the data into a higher-dimensional space. We use the \ac{SVM} implementation of \citet{e1071} with a radial basis function kernel, with the parameters $cost=1$ and $\gamma=(number~of~features)^{-1}$.

To assess the predictive performance, we compare the true classes of the test set with the predicted classes and use the \ac{ROC} curve. The curve graphically represents the trade-off between the true positive and the false postive rate for different classification thresholds in a unit square. Based on the curve, the \ac{AUC} is a reliable classification performance metric that ranges between 0 and 1, where 0.5 represents random classification and values above indicate a better predictive performance. The majority of earlier \ac{FSM} benchmark studies (see \autoref{tab:literature}) use weaker metrics that simply count correct or wrong classified examples (e.g., accuracy, precision, recall, F-measure). We decided to use \ac{AUC}, because it is not biased by the class distribution of the dependent variable \citep{fawcett_introduction_2006}. Therefore, \ac{AUC} values can be compared across different classification problems.

We used stratified 10-fold cross-validation to estimate performance metrics \citep{kohavi_study_1995}. With this apporach, the data is divided into ten equally sized subsets, taking into account the class distribution. In ten iterations, each of these subsets were used as the test dataset and the remaining nine subsets as the training dataset for feature selection and model training. We estimate the predictive performance by computing the mean of the ten individual values. Given that the observations are allocated into the ten folds using random sampling, the model performance is affected by this sampling. To lower this effect, we repeated the stratified tenfold cross-validation five times. This means that each filter method was applied a total of 50 times per dataset and the 50 values of the performance metric were averaged.

\subsubsection{Number of relevant features selected}

As a second criterion, we propose to consider the number of relevant features that each \ac{FSM} selected. This metric is common to all studies that do not consider classification algorithms in their benchmark of \acp{FSM} (see \autoref{tab:literature}). Given that we employ synthetic data for the comparison of filter methods, we know in advance what features are relevant, redundant or irrelevant.

\subsubsection{Stability of selected feature sets}

Due to random components in the \acp{FSM}, and given that only subsets of data are used to evaluate them, several runs of \acp{FSM} might produce different feature sets that cause variations in the \ac{ML} models. To measure the stability, 
we suggest to use the distance metric recently suggested by \citet{nogueira_stability_2018}. They suggest a metric that overcomes several undesirable properties of other pairwise similarity or frequency based metrics that were used in earlier studies that compare \acp{FSM}. 

Given $X_1, \ldots, X_p$ available features ($p$ is the number of available features) in $m$ datasets, a \ac{FSM} obtains $V_1, \ldots, V_m$  sets of chosen features based on the datasets. 
%
%
When $h_j$ denotes the number of sets that contain feature $X_j$ so that $h_j$ is the absolute frequency with which feature $X_j$ is chosen and $q = \sum_{j=1}^p h_j = \sum_{i=1}^m |V_i|$ is the average number of features chosen for the $m$ datasets, then

$$\hat{\mathcal{N}}(V) = 1 - \frac{\frac{1}{p} \sum_{j=1}^p \frac{m}{m-1} \frac{h_j}{m} \left(1 - \frac{h_j}{m}\right)} {\frac{q}{mp} (1 - \frac{q}{mp})}$$

is a stability metric for a \ac{FSM} that is fully defined, is strict monotonic, has upper and lower bounds, has its maximum value when all feature sets are equal, and is corrected for chance \citep{nogueira_stability_2018}.

\subsubsection{Runtime}

Finally, the time (e.g., in seconds) that the execution of each \ac{FSM} takes in an analysis is a practically relevant criterion. Especially in the case of large datasets this matters as a selection criterion for practical use. 

Our study focuses on selected dataset characteristics, but not on the size of the datasets, as this was already subject of earlier studies \citep{bommert_benchmark_2020, saeys_robust_2008, aphinyanaphongs_comprehensive_2014}. Thus, we include the criterion to complete our picture on the investigated methods. 

For the benchmark experiments, we used an instance of Amazon Elastic Compute Cloud (Amazon EC2) and kept the hardware configuration similar throughout all experiments. The instances were of type \textit{m5.2xlarge}, had eight virtual CPUs and 32 gibibibytes of RAM. The software environment was Ubuntu 18.04 LTS and we used R version 3.6.0. 

\section{Filter methods in GNU R}

The focus of our benchmark study is the statistical programming environment GNU R, which is open source software and has gained high popularity in academia and practice\footnote{See Robert A. Muenchen's website for latest popularity figures: \url{http://r4stats.com/articles/popularity/} (last access on August 18, 2020)}. We searched the \ac{CRAN}, a repository for open source R software libraries, for implemented \acp{FSM} that could be used by analysts and \ac{ML} users. We found the packages \textit{CORElearn}, \textit{FSelector}, \textit{mlr}, and \textit{praznik} that in total contain \numFSMs unique methods. 
We considered the \acp{FSM} only once, even if they were implemented in multiple methods. One exception is the method \textit{gain ratio}, because \textit{FSelector} and \textit{CORElearn} use different data discretization methods. We excluded the \textit{m:variance} filter from our benchmark, which identifies relevant features based on their variance. As our experiments are based on synthetically generated data, this method almost perfectly separated relevant from irrelevant ones which we consider is not the true behavior of the method with real data.
\autoref{tab:fsm_description} lists all considered \acp{FSM} with a brief description and, if we found, a literature reference. We denote the package that contains a method with the leading letters \textit{C, F, m,} and \textit{p} respectively. If there is no literature reference given, we point readers to the package documentation in the respective library.

\tiny
\rowcolors{2}{gray!15}{white}
\input{tables/fsm_desc.tex}
\normalsize

\doublespacing

With regard to the investigated scenarios of dataset challenges of low data quality and small datasets, we determined whether the methods are suitable for applying it to noisy data (i.e., class and attribute noise) and imbalanced class distributions, based on the method documentation. For this, we reviewed the method documentations in order to find respective descriptions. In addition, we determined whether each methods considers features individually or in combination (multivariate). Multivariate methods can be assumed to be more suitable to exclude redundant features. In total, we found 16 methods with special support for class noise, 15 methods for attribute noise, 9 methods that are cost-sensitive which means they have special support for imbalanced class problems, and 27 methods that are multivariate, whereas 31 were not.

The nine cost-sensitive methods require a cost matrix as an additional parameter to be considered in the feature selection. We followed the recommendation of \citet{robnik-sikonja_experiments_2003} and weighted the misclassification of the minority class 20 times more than the misclassification of the majority class. When data sets are imbalanced, there is usually a greater interest in correctly classifying the minority class, so higher costs are assigned to misclassification of the minority class. The nine cost-sensitive methods were used exclusively for the two datasets with an imbalanced class distribution, since the application of these methods with a cost matrix is not useful when the class distribution is balanced \citet{robnik-sikonja_experiments_2003}.

\section{Results}
\label{sec:results}

We organize the presentation of the results in three steps. First, we present the results of each evaluation with respect to the four dataset scenarios that are 
(I) class and attribute noise, 
(II) redundant features (both are data quality problems in the data),
(III) imbalanced data, and
(IV) more features than observations (both are problems of small datasets). Second, we test whether methods that expose special support in their documentation for one of the dataset scenario outperform other methods with this respect. Finally, we make a multi-criteria comparison of the \acp{FSM} considering all evaluation criteria to obtain an overall comparison of the benchmarked methods. Before we start describing the results, we briefly introduce the method used to analyze the results.

\subsection{Analysis of benchmark results}

We analyze the benchmark results using a multiple linear regression with ordinary least squares estimation and robust standard errors \citep{white_heteroskedasticity-consistent_1980, zeileis_econometric_2004}. This approach is common when investigating several control variables on an outcome variable \citep{rawlings_applied_1998}. For the analyses, we use the following model specification:

$$Y^\gamma = \beta_0^\gamma + \beta_1^\gamma \ast FSM + \beta_2^\gamma \ast C_1 + \ldots + \beta_m^\gamma \ast C_{m-1} + \epsilon$$

The dependent variable is one of the evaluation criteria $\gamma$, namely $\gamma = \lbrace$ \textit{predictive performance (\ac{AUC}), number of relevant features, stability, runtime} $\rbrace$. We dedicate separate subsections to each of the dependent variables in the following. 

\textit{FSM} is a categorical variable with a dummy-encoding that contains the name of the \ac{FSM} (see \autoref{tab:fsm_description} for the full list). In total, our benchmark comprises \numFSMs methods. Nine of them are variants of other methods that add cost-sensitivity to better handle class imbalance \citep{robnik-sikonja_experiments_2003}. Given that in a uniform class distribution, the results of these are similar to the original methods, we only computed results for the imbalanced dataset scenario for these cost-sensitive variants.
For the analysis of the results, we consider no feature selection as a reference case. This means that the coefficients of the estimates labeled with ``method*" are deviations from a prediction model without feature selection. 

$C_1, \ldots, C_{m-1}$ are control variables that help us to quantify the impact of model characteristics that we vary during the simulation experiments. \autoref{tab:controls} lists all used control variables together with their values. These values result from the generated datasets (see \autoref{sec:datasets}), bold values are the respective baseline value in the models. 

\begin{table}[h]
	\caption{Control variables in the statistical models for benchmark analyses}\label{tab:controls}
	\small
	\begin{tabu}{llXl}
		\toprule
		\multicolumn{2}{l}{Control variable} 		& Explanation & Values\\
		\midrule
		$C_1$ & classnoise 				& Percent of noise added to the dependent variable & \textbf{0}, 0.1, 0.2, 0.3\\
		$C_2$ & attributenoise			& Percent of noise added to the features & \textbf{0}, 0.1, 0.2, 0.3\\
		$C_3$ & classifier				& Supervised \ac{ML} algorithm used & \textbf{\ac{NB}}, \ac{SVM}, \ac{RF} \\
		$C_4$ & num\_redundant\_features& Number of redundant features & \textbf{0}, 10, 20\\
		$C_5$ & minClassDev 			& Deviation of the equal distribution in percentage points, multiplied with factor 10. & \textbf{0}, 3, 4\\
		$C_6$ & relFeatObs				& Relation of the number of features $p$ to the number of observations $n$ & $\mathbf{\frac{100}{2500}}, \frac{500}{2500}, \frac{1000}{500}$\\
		\bottomrule
	\end{tabu} 
\end{table}

The unit of analysis for the regression models are the performances of each method under a certain combination of control variables and one fold of the 10-fold cross validation. For all analyses, we kept the same random allocation of data examples in the cross-folds to ensure comparability. For regression analysis this leads to relatively large samples (745-5250 measurements), which quickly lead to statistically significant results. For completeness, we report the levels of significance in this paper, but focus our evaluation on effect sizes and ranks. 

The analysis method allows to compare the performance metrics with the reference case (no feature selection), but cannot express the performance differences between single \acp{FSM}. This is, however, without detriment given that such a comparison would allow little conclusion to be drawn due to the relatively high variance and closely divergent results of several methods. In addition, such a comparison is also not recommended as it misleads conclusions \citep{demsar_statistical_2006}.

\subsection{Predictive performance}

We computed linear models based on the predictive performance for the four dataset scenarios and show the results in \autoref{tab:models_auc}. The reference case---a model trained with the \ac{NB} classifier, without feature selection, using the Baseline data for training---shows classification performance that is higher than $AUC=0.5$ in all four scenarios, which indicates a classification performance that is better than random. 

The estimated coefficients in \autoref{tab:models_auc} show the performance deviation from the base case, which means that all methods with a positive estimate that is significantly different from zero (as indicated by the star notation of the significance levels) showed a better performance than no feature selection. The overall best performance in this analysis showed \textit{F:random.forest.importance, p:DISR} in all scenarios.  \textit{p:ranger\_impurity, p:ranger\_permutation}, and \textit{p:JMI} perform very well in some scenarios.

The models explain between 65 and 77 percent of variance (Adjusted $R^2$) and the control variables show meaningful estimates: 
The use of \ac{RF} and \ac{SVM} improves the classification by several percentage points compared to the base case of \ac{NB}. This effect is particularly strong in scenario II (Redundant).
In Scenario I (Noise), the performance decreases strongly with noise (if the dependent variable is noisy much more than if the features are noisy), with very strong noise the estimated \ac{AUC} decreases as expected below 0.5. In Scenario III (Imbalanced), the performance decreases with increasing skew of the dependent variable distribution. In scenario IV (Dimensionality), the performance decreases with an increasing ratio of the number of features to the number of observations.
All in all, the estimates of the control variables are as we expected and the model quality is sufficient. This allows us to analyze the effects of the \acp{FSM} on the model performance.

\begin{table}
	\caption{Models for the criterion 'predictive performance' (AUC) in the four dataset scenarios}\label{tab:models_auc}
	\tiny
	\centering
	\rowcolors{2}{gray!15}{white}
	\input{tables/models_auc.tex}
\end{table}

\subsection{Percentage of relevant features}

\begin{figure}[h]
	\small
	\sffamily
	\input{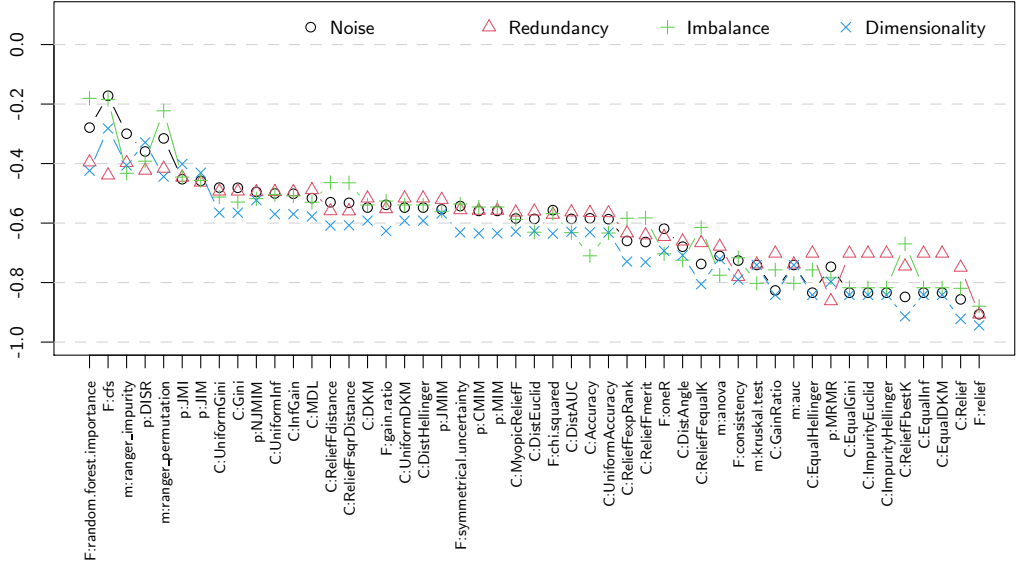}
	\vspace{-2em}
	\caption{Loss in percentage of relevant features selected per method and dataset scenario}\label{fig:coefs_method_relevant}
\end{figure}

For the percentage of relevant features that each \ac{FSM} selects, we computed linear models for each dataset scenario. We list the complete model in \autoref{tab:models_relev} in the appendix (on \autopageref{tab:models_relev}) and illustrate the estimated loss of performance (here: percentage of relevant features selected) compared with no feature selection in \autoref{fig:coefs_method_relevant}. In order to compare the number of relevant features selected in the four dataset scenarios, we computed the rank of each \ac{FSM} for each scenario separately and we ordered the methods in the figure according to their average rank---from left (best) to right (worst). 

Among the top five methods according to this evaluation criterion, there is no clear picture which one is best overall, as it depends on the dataset scenario. However, the filters based on \ac{RF} feature importance measures (\textit{F:random.forest-importance} and \textit{m:ranger\_impurity}) together with \textit{F:cfs} seem to perform particularly well in selecting relevant features with noisy and imbalanced data. \textit{p:DISR} is also among the top five methods and performs well across all dataset scenarios.

\subsection{Stabiliy of selected feature sets}

\begin{figure}[h]
	\small
	\sffamily
	\input{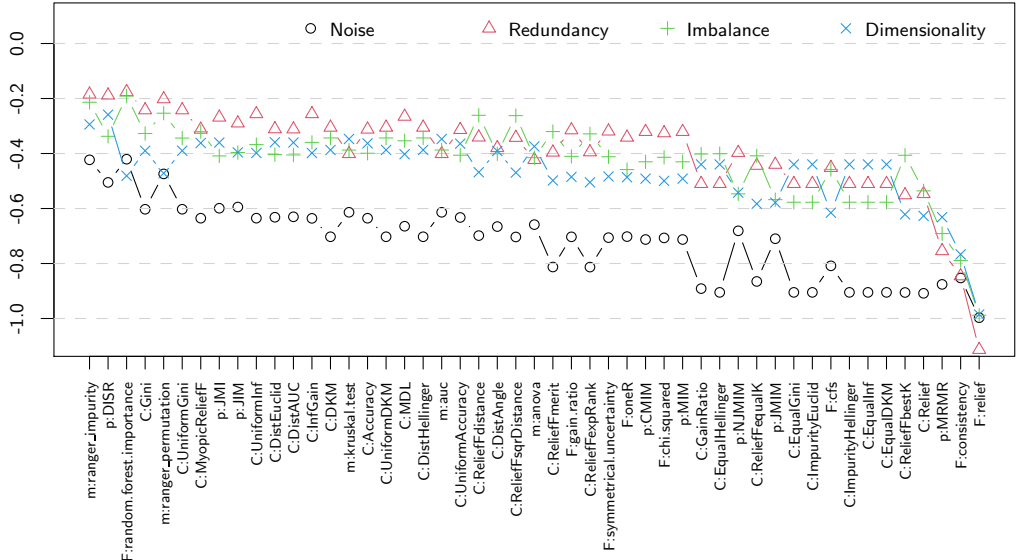}
	\vspace{-2em}
	\caption{Stability deviation from no selection per method and dataset scenario}\label{fig:coefs_method_stability}
\end{figure}

We proceeded similarly for the stability of selected feature sets, which we measured in the distance $\hat{\mathcal{N}}$ of \citet{nogueira_stability_2018}, and computed linear models for each dataset scenario (see \autoref{tab:models_stab} in the appendix on \autopageref{tab:models_stab}). We illustrate the estimated stability loss per method, compared to no feature selection, ordered by average rank, in \autoref{fig:coefs_method_stability}.

The four dataset scenarios have a strong influence on the stability of selected features sets, especially in the noise scenario, selected feature sets were less stable than in others. Similarly to the earlier evaluation metrics, there is no method that clearly outperforms all others, but apparently, the \ac{RF}-based methods (\textit{m:ranger\_impurity, F:random.forest.importance}, and \textit{m:ranger\_permutation}) rank again among the top performing methods. \textit{p:DISR} is also among the top five methods.

\subsection{Runtime}

For the method runtime, we estimated linear models that we show in \autoref{tab:models_runtime}. Given that we ran all tests on the same R instance, we considered no control variable in the models. The methods in the first three scenarios (noise, redundant, imbalanced) have a very low runtime. Thus, only some methods have higher than one second runtime (they need significantly more computational resources). In the dataset scenario with more features than observations, especially the methods \textit{F:cfs, F:consistency}, and \textit{F:relief} showed higher execution times.
Methods with the lowest runtime (on average below 5 seconds in all four dataset scenarios) are \textit{m:auc, p:MRMR, p:MIM}, and \textit{p: JIM}.

\begin{table}[h]
	\caption{Models for the criterion 'method runtime' in the four dataset scenarios}\label{tab:models_runtime}
	\tiny
	\centering
	\rowcolors{2}{gray!15}{white}
	\input{tables/models_runtime.tex}
\end{table}

\begin{figure}[h]
	\footnotesize
	\sffamily
	\input{figures/runtime_auc.tex}
	\caption{Runtime and predictive performance for each method over all experiments}\label{fig:runtime_auc}
\end{figure}
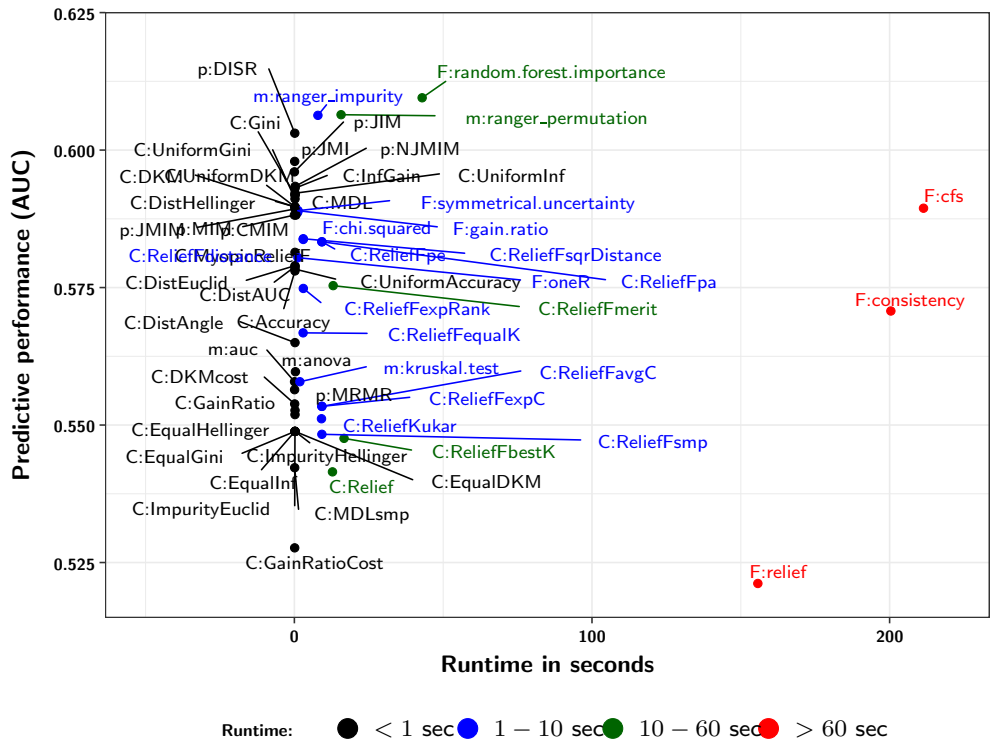

In addition, we compare the runtime of all methods with the predictive performance, as we would intuitively assume that computationally more expensive methods lead to higher predictive performance. Our experiments, however, could not support this hypothesis, as we found no correlation between method runtime and predictive performance (Pearson's $r(56) = -.08$, $p = .543$). This result is also illustrated in \autoref{fig:runtime_auc}.

\subsection{FSM characteristics}

In our review of the \acp{FSM}, we found three properties that make methods appear particularly suitable for certain dataset scenarios. These are robustness to noise, cost sensitivity to counteract class imbalance and the consideration of feature combinations (multivariate methods) for the detection of redundant features. In the following we investigate whether the method properties contribute with statistical evidence to a better handling of the respective dataset challenges. \autoref{tab:stat_tests} gives an overview to these results.

\begin{table}[b]
	\caption{Mean and standard deviation (in brackets) of predictive performance for three method characteristics}\label{tab:stat_tests}
	\small
	\begin{tabu}{lrX[r]X[r]X[r]}
		\toprule
		Subset of methods 		&Methods & $AUC_{NB}$ & $AUC_{RF}$ & $AUC_{SVM}$\\
		\midrule
		Robust for attribute noise	&10& 0.552 (0.02) &  0.577 (0.04) &  0.577 (0.04) \\
		Other					&39& 0.564 (0.02) &  0.605 (0.06) &  0.612 (0.06) \\
		\midrule
		Robust for class noise	&11& 0.552 (0.02) &  0.577 (0.04) &  0.577 (0.04) \\
		Other					&38& 0.564 (0.02) &  0.606 (0.06) &  0.613 (0.06) \\
		\midrule
		Cost-sensitive			& 9& 0.555 (0.02) &  0.548 (0.03) &  0.539 (0.03) \\
		Other					&49& 0.568 (0.02) &  0.606 (0.05) &  0.607 (0.06) \\
		\midrule
		Multivariate			&20& 0.626 (0.03) &  0.749 (0.07) &  0.766 (0.08) \\
		Univariate				&29& 0.618 (0.02) &  0.714 (0.06) &  0.727 (0.07) \\
		\midrule
		
	\end{tabu}
\end{table}

\paragraph{Performance of noise robust methods with noisy data:}

We tested whether the methods that specified that they support attribute or class noise perform better than other methods. We found neither support for the 11 methods for attribute noise ($t(47) \leq -1.44$, $p \geq .922$, $d \leq -0.51$), nor for the 10 class noise resistant methods ($t(47) \leq -1.65$, $p \geq .947$, $d \leq -0.57$). Given that our experiments with noise do not include class imbalance, we did not include the cost-sensitive variants of methods in this comparison. The statistics of attribute and noise robust methods are almost similar, because there is only one method difference in the set of methods. 

\paragraph{Appropriateness of cost-senstive methods on imbalanced data:}

According to the method descriptions, cost-sensitive filter methods are particularly designed for imbalanced datasets \citep{robnik-sikonja_experiments_2003}. In our experiments with the imbalanced class dataset scenario, however, the methods had a lower average predictive performance than other methods. The differences are statistically significant with $t(56) \geq 1.80$, $p \leq .038$, $d \geq 0.65$.

\paragraph{Performance of multi-dimensional methods on redundant features:}

In the case of datasets with correlations and redundant features, filter methods that take into account dependencies between features (multivariate methods) should perform significantly better than univariate methods. We find support for this hypothesis based on our experiments, as the multivariate methods performed significantly better than others in our redundant feature dataset scenario with the \ac{RF} and \ac{SVM} classifier ($t(47) \geq 1.81$, $p \leq .036$, $d \geq 0.53$).

\subsection{Multi-criteria benchmark of methods}

Finally, we compare the benchmark results for the methods with all evaluation criteria in \autoref{tab:benchmark_overview}. This table denotes the predictive performance with a positive / negative sign when the \ac{AUC} is significantly higher / lower than no feature selection. The top / bottom five methods are denoted with $++$ / $--$. We denote percentage of relevant features and the stability with the symbols based on the 10, 25, 50, 75, an 90 percent quantile. The runtime performance is denoted with a $++$ when the runtime was lower than 0.1 seconds, with $+$ when the runtime was less than 1 seconds, with a $0$ when the runtime was between 1 and 10 seconds, with $-$ when it is between 10 and 60 seconds, and with $--$ when it exceeded 60 seconds.

\begin{table}
	\caption{Benchmark results overview}\label{tab:benchmark_overview}
	\begin{footnotesize}
		\rowcolors{2}{gray!15}{white}
		\input{tables/benchmark_summary.tex}
	\end{footnotesize}
\end{table}

\section{Discussion and Implications}

This study provided an overview to existing benchmark studies on \acp{FSM} and compared the peformance of \numFSMs \acp{FSM} in four typical dataset scenarios that are challenging for ML models, namely: Noisy data, redundant data, imbalanced datasets, and datasets with more features than observations. To examine the performance of \acp{FSM} in these scenarios, we created 13 synthetic datasets with characteristics that represent these dataset challenges. 
We measured the performance of the methods using four criteria that earlier studies have found to be reasonable. The criteria were (1) predictive performance (we used \ac{AUC} as a reliable metric and tested three \ac{ML} algorithms), (2) the number of selected relevant features, (3) the stability of the feature sets (for this purpose we used a recently developed distance metric for feature sets \citep{nogueira_stability_2018} which overcomes limitations of previous metrics), and (4) the method runtime. 

Over all dataset scenarios and criteria, those methods that are based on random forest rankings \\ (\textit{F:random.forest.importance}, \textit{m:ranger\_impurity}, \textit{m:ranger\_permutation}) show a good performance, however their runtime  was higher than average in our experiments. Similarly well performed the methods \textit{p:DISR} (a normalized version of \textit{p:JMI}, which also showed a good performance) and \textit{p:JIM}, considering all criteria and dataset scenarios. We compiled a summary of the results for each dataset scenario and criterion in \autoref{tab:benchmark_overview}.

\subsection{Implications of the results}

Based on our empirical comparison, we can recommend researchers and quantitative modelers to start their search for an appropriate \ac{FSM} with these filter methods.

We also tested three characteristics that the respective documentation of the methods exhibited as special features. The results of these statistical analyses were suprising: Methods that were described as \textit{noise robust} did not perform better with noisy data than other methods (neither for class nor for attribute noise). Also, the use of cost-senstive methods, which implement approaches to encounter \textit{class imbalance} \citep{robnik-sikonja_experiments_2003}, did not achieve better results on imbalanced data than other methods. One reason for this result may be that methods that claim to be particularly suitable for noisy data or imbalanced datasets were developed at a time when the methods that performed well in our benchmark did not exist. Based on this finding, future research should further develop well-performing approaches in our benchmark for these specific dataset scenarios. Conversely, developers and maintainers of methods should be careful in labeling their methods as particularly suitable for a certain dataset scenario.

Only multi-dimensional methods showed a better performance with \textit{redundant features} than other methods. As a consequence of this result, users do not need to apply special methods for noisy data and imbalanced datasets and can use those that have performed well in our benchmark in these scenarios or in general. This saves effort and resources during implementation and when training experts. Only for datasets with redundant features, we recommend to use multi-dimensional methods. 

\subsection{Limitations and future research}

We consider four limitations of our study, which we believe are areas for future research. 
First, our benchmark is based on synthetic data. This allowed us to deliberately evaluate \acp{FSM} regarding the dataset characteristics in the focus of our study, under laboratory conditions and independent of any application field. Although we have used established data generation methods from the \ac{ML}-field, the data generation is subject to assumptions from which reality will differ. Earlier studies on the comparison of \acp{FSM} used synthetic as well as real datasets. Therefore, we motivate future research that validates our results with data from various application domains.
Second, we measured the runtime for all our experiments with the same hardware and the same datasets to be valid internally. This measurement is likely to be different in magnitude when the employed hardware and software environment is modified. 
Given that the focus of our study was the comparison of \acp{FSM} with respect to different dataset scenarios, we refer readers who are interested in more detailed analysis of the algorithms runtime with very large datasets to previous studies, e.g., \cite{bommert_benchmark_2020}, \cite{aphinyanaphongs_comprehensive_2014}, and \cite{xue_comprehensive_2015}. 
Third, we considered three well-known \ac{ML} algorithms to obtain the predictive performance. A promising approach for business data analytics is deep learning which is capable to use raw data directly without the need of feature selection \citep{kraus_deep_2020}. However, such neural networks are ``data hungry" \citep[p. 6]{marcus2018deep} and therefore only usable in some applications. We motivate future research to examine how classical \ac{ML} algorithms together with \acp{FSM} perform in comparison with deep learning approaches. 
Fourth, the dataset challenges considered in this study are only a selection of challenges and further ones should be addressed. For example, the problem of missing values usually makes data analysis difficult. Even though there are numerous basic approaches how to deal with missing values, we could not find \acp{FSM} that explicitly address this problem. We motivate future research to further develop the methods in this respect.

\section{Conclusion}

Feature selection is an important task in setting up \ac{ML} models in business analytics. Identifying most appropriate features for predictive modeling enables the creation of less complex models that are more interpretable in decision situations, can improve model performance, and saves computational effort in later analysis steps. These are particular requirements in big data analytics tasks, not only today, given that digitization will allow measuring a wide variety of new variables in the future. 

As feature selection is important, many methods have been suggested so far: In the widely used statistical software environment GNU R, we found over fifty different methods---ready to use by analysts and researchers. A comprehensive comparison of these methods, however, is currently missing, which makes the decision on which method to use difficult. 

Our study narrowed this gap and first provided a comprehensive literature review to benchmarks of \acp{FSM}. We outlined limitations of current studies and suggested---based on the learnings of earlier benchmark studies---a multi-criteria benchmark method for \acp{FSM}, which we applied in the second part of our paper. There, we provided a benchmark of \numFSMs currently available feature filter methods in the GNU R environment. For the comparison of methods, we used four dataset challenges that are typical in \ac{ML} applications. The method overview and comparison guides the automatic feature selection task of researchers and quantitative modelers. Our benchmark method provides a base for further studies that employ it to case-specific datasets.

\section*{Acknowledgements}
We appreciate Andreas Weigert's support in implementing early stages of the benchmark analysis and his insightful comments to earlier versions of this manuscript. We thank Thorsten Staake for his feedback on the initial  study design. 

\section*{Appendices}

Additional statistical models are listed in  \autoref{tab:models_relev} and \autoref{tab:models_stab}.

\input{tables/models_relev.tex}

\input{tables/models_stab.tex}

\clearpage

\bibliographystyle{plainnat}
\bibliography{FSM_literature}

\end{document}

%% file: tables/fsm_desc.tex
\begin{longtable}{p{1.8cm}p{6.0cm}p{2.5cm}cccc}
	\caption{Feature filter methods considered in the benchmark study}\label{tab:fsm_description}\\
	\toprule
	\hiderowcolors
	Name & Description & Ref. & \multicolumn{2}{c}{Noise} & Cost 	  & Multi- \\ 
	\cmidrule{4-5}
	& 			   &          & Class & Attribute 		  & sensitive & variate \\ 
	\showrowcolors
	\midrule
	\endfirsthead
	\hiderowcolors
	\toprule
	Name & Description & Ref. & \multicolumn{2}{c}{Noise} & Cost 	  & Multi- \\ 
	\cmidrule{4-5}
	& 			   &          & Class & Attribute 		  & sensitive & variate \\ 
	\showrowcolors
	\midrule
	\endhead 
	
	\bottomrule
	\endfoot

	\bottomrule
	\endlastfoot
	
  C:Accuracy & Filter based on Accuracy of resulting split &  &  &  &  &  \\ 
  C:DistAngle & Cosine of angular distance between splits &  &  &  &  &  \\ 
  C:DistAUC & AUC distance between splits &  &  &  &  &  \\ 
  C:DistEuclid & Euclidean distance between splits &  &  &  &  &  \\ 
  C:DistHellinger & Hellinger distance between class distributions in branches &  &  &  &  &  \\ 
  C:DKM & A measure that is suitable for two class problems & \citep{Dietterich_Applyingweaklearning_1996} &  &  &  &  \\ 
  C:DKMcost & Cost-sensitive variant of DKM & \citep{robnik-sikonja_experiments_2003} &  &  & $\times$ &  \\ 
  C:EqualDKM & DKM with equal weights for splits &  &  &  &  &  \\ 
  C:EqualGini & Gini index with equal weights for splits &  &  &  &  &  \\ 
  C:EqualHellinger & Two equally weighted splits based Hellinger distance &  &  &  &  &  \\ 
  C:EqualInf & Information gain with equal weights for splits & \citep{hunt_experiments_1966} &  &  &  &  \\ 
  C:GainRatio & Gain ratio, which is normalized information gain to prevent bias to multi-valued attributes & \citep{quinlan_induction_1986} &  &  &  &  \\ 
  C:GainRatioCost & Cost-sensitive variant of GainRatio & \citep{robnik-sikonja_experiments_2003} &  &  & $\times$ &  \\ 
  C:Gini & Gini-index of the attributes & \citep{breiman_random_2001} &  &  &  &  \\ 
  C:ImpurityEuclid & Euclidean distance as impurity function on within node class distributions &  &  &  &  &  \\ 
  C:ImpurityHellinger & Hellinger distance as impurity function on within node class distributions &  &  &  &  &  \\ 
  C:InfGain & Information Gain as used in the original decision tree & \citep{quinlan_induction_1986} &  &  &  &  \\ 
  C:MDL & Minimum Description Length, a method with favorable bias for multi-valued and multi-class problems & \citep{kononenko_estimating_1994} &  &  &  &  \\ 
  C:MDLsmp & Cost-sensitive variant of MDL where costs are introduced through sampling & \citep{robnik-sikonja_experiments_2003} &  &  & $\times$ &  \\ 
  C:MyopicReliefF & Myopic version of the 'ReliefF' algorithm resulting from assumption of no local dependencies and attribute dependencies upon class & \citep{kononenko_estimating_1994} & $\times$ & $\times$ &  &  \\ 
  C:Relief & Calculates scores for each feature, based on the Euclidean distance to nearest neighbor training instance pairs & \citep{kira_practical_1992} & $\times$ & $\times$ &  & $\times$ \\ 
  C:ReliefFavgC & Cost-sensitive 'ReliefF' version with average costs & \citep{robnik-sikonja_experiments_2003} & $\times$ & $\times$ & $\times$ & $\times$ \\ 
  C:ReliefFbestK & 'ReliefF' variant testing all possible k nearest instances for each feature and returns the highest score & \citep{robnik-sikonja_experiments_2003} & $\times$ & $\times$ &  & $\times$ \\ 
  C:ReliefFdistance & 'ReliefF' variant where k nearest instances are weighed directly with its inverse distance from the selected instance & \citep{robnik-sikonja_experiments_2003} & $\times$ & $\times$ &  & $\times$ \\ 
  C:ReliefFequalK & 'ReliefF' algorithm where k nearest instances have equal weight & \citep{robnik-sikonja_experiments_2003} & $\times$ & $\times$ &  & $\times$ \\ 
  C:ReliefFexpC & Cost-sensitive 'ReliefF' algorithm with expected costs & \citep{robnik-sikonja_experiments_2003} & $\times$ & $\times$ & $\times$ & $\times$ \\ 
  C:ReliefFexpRank & 'ReliefF' algorithm where k nearest instances have weight exponentially de-creasing with increasing rank; ranks are determined by the increasing (Manhattan) distance from the selected instance; conditional dependencies among attributes are taken into account &  &  &  &  & $\times$ \\ 
  C:ReliefFmerit & 'ReliefF' algorithm where for each random instance the merit of each feature is normalized by the sum of differences in all attributes & \citep{robnik-sikonja_experiments_2003} & $\times$ & $\times$ &  & $\times$ \\ 
  C:ReliefFpa & Cost-sensitive 'ReliefF' algorithm with average probability & \citep{robnik-sikonja_experiments_2003} & $\times$ & $\times$ & $\times$ & $\times$ \\ 
  C:ReliefFpe & Cost-sensitive 'ReliefF' algorithm with expected probability & \citep{robnik-sikonja_experiments_2003} & $\times$ & $\times$ & $\times$ & $\times$ \\ 
  C:ReliefFsmp & Cost-sensitive 'ReliefF' algorithm with cost sensitive sampling & \citep{robnik-sikonja_experiments_2003} & $\times$ & $\times$ & $\times$ & $\times$ \\ 
  C:ReliefFsqr\-Distance & 'ReliefF' variant where k nearest instances are weighed with its inverse square distance from the selected instance & \citep{robnik-sikonja_experiments_2003} & $\times$ & $\times$ &  & $\times$ \\ 
  C:ReliefKukar & Cost-sensitive 'Relief' variant & \citep{kukar_analysing_1999} & $\times$ & $\times$ & $\times$ & $\times$ \\ 
  C:UniformAccuracy & Accuracy with uniform priors &  &  &  &  &  \\ 
  C:UniformDKM & DKM measure with uniform priors &  &  &  &  &  \\ 
  C:UniformGini & Gini index with uniform priors &  &  &  &  &  \\ 
  C:UniformInf & Information gain with uniform priors &  &  &  &  &  \\ 
  F:cfs & The algorithm finds attribute subset using correlation and entropy measures for continuous and discrete data & \citep{hall_correlation-based_1999} & $\times$ & $\times$ &  & $\times$ \\ 
  F:chi.squared & The algorithm finds weights of discrete attributes basing on a chi-squared test & \citep{liu_chi2_1995} &  &  &  &  \\ 
  F:consistency & The algorithm finds attribute subset using consistency measure for continuous and discrete data & \citep{dash_consistency_2000} & $\times$ &  &  & $\times$ \\ 
  F:gain.ratio & The algorithms find weights of discrete attributes basing on their correlation with continuous class attribute & \citep{quinlan_induction_1986} &  &  &  &  \\ 
  F:oneR & Find weights of discrete attributes basing on very simple association rules involving only one attribute in condition part & \citep{holte_very_1993} &  &  &  &  \\ 
  F:random.forest.\-importance & The algorithm creates a weighting of the features using the Random Forest algorithm. & \citep{breiman_random_2001} &  &  &  & $\times$ \\ 
  F:relief & The algorithm calculates scores for each feature, based on the Euclidean distance to nearest neighbor training instance pairs & \citep{kira_practical_1992} & $\times$ & $\times$ &  & $\times$ \\ 
  F:symmetrical.\-uncertainty & Weighting of discrete features based on the correlation to the target variable; unlike information gain, no preference for features with many values & \citep{yu_feature_2003} &  &  &  &  \\ 
  m:anova & The algorithm finds weights of features based on an analysis of variance. & \citep{bommert_benchmark_2020} &  &  &  &  \\ 
  m:auc & AUC filter for binary classification problems. Weight of a feature is based on achieved AUC when used directly and separately from other features for prediction. & \citep{bommert_benchmark_2020} &  &  &  &  \\ 
  m:kruskal.test & The algorithm finds weights of features based on a Kruskal test. & \citep{bommert_benchmark_2020} &  &  &  &  \\ 
  m:ranger\_impurity & Variable importance based on ranger impurity importance &  &  &  &  & $\times$ \\ 
  m:ranger\_per\-mutation & Variable importance based on ranger permutation importance &  &  &  &  & $\times$ \\ 
  p:CMIM & First selects the feature with the greatest maximum mutual information with the target variable, then featires are iteratively selected, which supply most information about the target variable, whereby the information of the features already selected is considered & \citep{fleuret_fast_2004} &  &  &  & $\times$ \\ 
  p:DISR & Normalized version of JMI & \citep{meyer_use_2006} &  &  &  & $\times$ \\ 
  p:JIM & Joint impurity filter &  &  &  &  & $\times$ \\ 
  p:JMI & First selects feature with the greatest maximum mutual information with the target variable, then features are selected iteratively, which maximize the cumulative summation of the common mutual information with the already selected feature subset & \citep{yang_data_2000} &  &  &  & $\times$ \\ 
  p:JMIM & Modification of JMI using minimal common information about already selected features instead of a sum & \citep{bennasar_feature_2015} &  &  &  & $\times$ \\ 
  p:MIM & Calculates the mutual information between all features and the target variable & \citep{battiti_using_1994} &  &  &  & $\times$ \\ 
  p:MRMR & The method first selects the feature with the greatest maximum mutual information with the target variable. Afterwards features are iteratively selected on the basis of measures of relevance and redundancy & \citep{peng_feature_2005} &  &  &  & $\times$ \\ 
  p:NJMIM & Normalized version of JMIM & \citep{bennasar_feature_2015} &  &  &  & $\times$ \\ 
\end{longtable}

%% file: tables/models_auc.tex
\begin{tabular}{l D{)}{)}{9)3} D{)}{)}{9)3} D{)}{)}{9)3} D{)}{)}{9)3}}
\toprule
 & \multicolumn{1}{c}{Noise} & \multicolumn{1}{c}{Redundant} & \multicolumn{1}{c}{Imbalanced} & \multicolumn{1}{c}{Dimensionality} \\
\midrule
(Intercept)                            & 0.60 \; (0.00)^{***}  & 0.56 \; (0.01)^{***}  & 0.62 \; (0.01)^{***}  & 0.58 \; (0.00)^{***}  \\
FSMs                                   &                       &                       &                       &                       \\
\quad C:Accuracy                       & 0.02 \; (0.00)^{***}  & 0.03 \; (0.01)^{***}  & -0.01 \; (0.01)       & 0.06 \; (0.01)^{***}  \\
\quad C:DistAngle                      & -0.01 \; (0.00)^{***} & -0.01 \; (0.01)       & -0.02 \; (0.01)^{**}  & 0.03 \; (0.01)^{***}  \\
\quad C:DistAUC                        & 0.02 \; (0.00)^{***}  & 0.03 \; (0.01)^{***}  & 0.00 \; (0.01)        & 0.06 \; (0.01)^{***}  \\
\quad C:DistEuclid                     & 0.02 \; (0.00)^{***}  & 0.03 \; (0.01)^{***}  & 0.00 \; (0.01)        & 0.06 \; (0.01)^{***}  \\
\quad C:DistHellinger                  & 0.04 \; (0.00)^{***}  & 0.06 \; (0.01)^{***}  & 0.04 \; (0.01)^{***}  & 0.08 \; (0.01)^{***}  \\
\quad C:DKM                            & 0.04 \; (0.00)^{***}  & 0.06 \; (0.01)^{***}  & 0.04 \; (0.01)^{***}  & 0.08 \; (0.01)^{***}  \\
\quad C:DKMcost                        &                       &                       & 0.01 \; (0.01)        &                       \\
\quad C:EqualDKM                       & -0.05 \; (0.00)^{***} & -0.04 \; (0.01)^{***} & -0.04 \; (0.01)^{***} & -0.01 \; (0.01)       \\
\quad C:EqualGini                      & -0.05 \; (0.00)^{***} & -0.04 \; (0.01)^{***} & -0.04 \; (0.01)^{***} & -0.01 \; (0.01)       \\
\quad C:EqualHellinger                 & -0.05 \; (0.00)^{***} & -0.04 \; (0.01)^{***} & -0.03 \; (0.01)^{**}  & -0.01 \; (0.01)       \\
\quad C:EqualInf                       & -0.05 \; (0.00)^{***} & -0.04 \; (0.01)^{***} & -0.04 \; (0.01)^{***} & -0.01 \; (0.01)       \\
\quad C:GainRatio                      & -0.04 \; (0.00)^{***} & -0.04 \; (0.01)^{***} & -0.03 \; (0.01)^{**}  & -0.01 \; (0.01)       \\
\quad C:GainRatioCost                  &                       &                       & -0.03 \; (0.01)^{***} &                       \\
\quad C:Gini                           & 0.05 \; (0.00)^{***}  & 0.06 \; (0.01)^{***}  & 0.04 \; (0.01)^{***}  & 0.07 \; (0.01)^{***}  \\
\quad C:ImpurityEuclid                 & -0.05 \; (0.00)^{***} & -0.04 \; (0.01)^{***} & -0.04 \; (0.01)^{***} & -0.01 \; (0.01)       \\
\quad C:ImpurityHellinger              & -0.05 \; (0.00)^{***} & -0.04 \; (0.01)^{***} & -0.04 \; (0.01)^{***} & -0.01 \; (0.01)       \\
\quad C:InfGain                        & 0.05 \; (0.00)^{***}  & 0.06 \; (0.01)^{***}  & 0.05 \; (0.01)^{***}  & 0.08 \; (0.01)^{***}  \\
\quad C:MDL                            & 0.05 \; (0.00)^{***}  & 0.06 \; (0.01)^{***}  & 0.05 \; (0.01)^{***}  & 0.08 \; (0.01)^{***}  \\
\quad C:MDLsmp                         &                       &                       & -0.00 \; (0.01)       &                       \\
\quad C:MyopicReliefF                  & 0.02 \; (0.00)^{***}  & 0.03 \; (0.01)^{***}  & 0.01 \; (0.01)        & 0.06 \; (0.01)^{***}  \\
\quad C:Relief                         & -0.05 \; (0.01)^{***} & -0.00 \; (0.02)       & -0.04 \; (0.01)^{***} & -0.05 \; (0.01)^{***} \\
\quad C:ReliefFavgC                    &                       &                       & -0.01 \; (0.01)       &                       \\
\quad C:ReliefFbestK                   & -0.04 \; (0.01)^{***} & -0.00 \; (0.02)       & -0.01 \; (0.01)       & -0.05 \; (0.01)^{***} \\
\quad C:ReliefFdistance                & 0.04 \; (0.01)^{***}  & 0.06 \; (0.01)^{***}  & 0.05 \; (0.01)^{***}  & 0.05 \; (0.01)^{***}  \\
\quad C:ReliefFequalK                  & -0.01 \; (0.00)^{***} & 0.03 \; (0.01)^{*}    & 0.01 \; (0.01)        & -0.00 \; (0.01)       \\
\quad C:ReliefFexpC                    &                       &                       & -0.01 \; (0.01)       &                       \\
\quad C:ReliefFexpRank                 & 0.00 \; (0.00)        & 0.04 \; (0.01)^{***}  & 0.02 \; (0.01)^{*}    & 0.02 \; (0.01)^{**}   \\
\quad C:ReliefFmerit                   & 0.00 \; (0.00)        & 0.04 \; (0.01)^{***}  & 0.02 \; (0.01)^{**}   & 0.02 \; (0.01)^{***}  \\
\quad C:ReliefFpa                      &                       &                       & 0.04 \; (0.01)^{***}  &                       \\
\quad C:ReliefFpe                      &                       &                       & 0.04 \; (0.01)^{***}  &                       \\
\quad C:ReliefFsmp                     &                       &                       & -0.02 \; (0.01)       &                       \\
\quad C:ReliefFsqrDistance             & 0.04 \; (0.01)^{***}  & 0.06 \; (0.01)^{***}  & 0.05 \; (0.01)^{***}  & 0.05 \; (0.01)^{***}  \\
\quad C:ReliefKukar                    &                       &                       & -0.01 \; (0.01)       &                       \\
\quad C:UniformAccuracy                & 0.02 \; (0.00)^{***}  & 0.03 \; (0.01)^{**}   & 0.00 \; (0.01)        & 0.06 \; (0.01)^{***}  \\
\quad C:UniformDKM                     & 0.04 \; (0.00)^{***}  & 0.06 \; (0.01)^{***}  & 0.04 \; (0.01)^{***}  & 0.08 \; (0.01)^{***}  \\
\quad C:UniformGini                    & 0.05 \; (0.00)^{***}  & 0.06 \; (0.01)^{***}  & 0.05 \; (0.01)^{***}  & 0.07 \; (0.01)^{***}  \\
\quad C:UniformInf                     & 0.05 \; (0.00)^{***}  & 0.06 \; (0.01)^{***}  & 0.05 \; (0.01)^{***}  & 0.08 \; (0.01)^{***}  \\
\quad F:cfs                            & 0.03 \; (0.00)^{***}  & 0.02 \; (0.01)^{*}    & 0.03 \; (0.01)^{***}  & 0.04 \; (0.01)^{***}  \\
\quad F:chi.squared                    & 0.04 \; (0.00)^{***}  & 0.06 \; (0.01)^{***}  & 0.04 \; (0.01)^{***}  & 0.05 \; (0.01)^{***}  \\
\quad F:consistency                    & -0.00 \; (0.00)       & -0.04 \; (0.01)^{***} & -0.00 \; (0.01)       & 0.02 \; (0.01)^{***}  \\
\quad F:gain.ratio                     & 0.04 \; (0.00)^{***}  & 0.06 \; (0.01)^{***}  & 0.05 \; (0.01)^{***}  & 0.05 \; (0.01)^{***}  \\
\quad F:oneR                           & 0.02 \; (0.00)^{***}  & 0.04 \; (0.01)^{***}  & -0.01 \; (0.01)       & 0.03 \; (0.01)^{***}  \\
\quad F:random.forest.importance       & \textbf{0.11} \; (0.01)^{***}  & \textbf{0.10} \; (0.01)^{***}  & \textbf{0.13} \; (0.01)^{***}  & \textbf{0.12} \; (0.01)^{***}  \\
\quad F:relief                         & -0.06 \; (0.01)^{***} & -0.13 \; (0.01)^{***} & -0.06 \; (0.01)^{***} & -0.06 \; (0.01)^{***} \\
\quad F:symmetrical.uncertainty        & 0.04 \; (0.00)^{***}  & 0.06 \; (0.01)^{***}  & 0.04 \; (0.01)^{***}  & 0.05 \; (0.01)^{***}  \\
\quad m:anova                          & -0.02 \; (0.00)^{***} & -0.03 \; (0.01)^{*}   & -0.04 \; (0.01)^{***} & 0.03 \; (0.01)^{***}  \\
\quad m:auc                            & -0.03 \; (0.00)^{***} & -0.04 \; (0.01)^{**}  & -0.04 \; (0.01)^{***} & 0.02 \; (0.01)^{*}    \\
\quad m:kruskal.test                   & -0.03 \; (0.00)^{***} & -0.04 \; (0.01)^{**}  & -0.04 \; (0.01)^{***} & 0.02 \; (0.01)^{**}   \\
\quad m:ranger\_impurity               & \textbf{0.10} \; (0.01)^{***}  & \textbf{0.09} \; (0.01)^{***}  & 0.06 \; (0.01)^{***}  & \textbf{0.12} \; (0.01)^{***}  \\
\quad m:ranger\_permutation            & \textbf{0.10} \; (0.01)^{***}  & \textbf{0.09} \; (0.01)^{***}  & \textbf{0.12} \; (0.01)^{***}  & \textbf{0.11} \; (0.01)^{***}  \\
\quad p:CMIM                           & 0.04 \; (0.00)^{***}  & 0.06 \; (0.01)^{***}  & 0.04 \; (0.01)^{***}  & 0.05 \; (0.01)^{***}  \\
\quad p:DISR                           & \textbf{0.10} \; (0.01)^{***}  & \textbf{0.09} \; (0.01)^{***}  & \textbf{0.08} \; (0.01)^{***}  & \textbf{0.13} \; (0.02)^{***}  \\
\quad p:JIM                            & \textbf{0.07} \; (0.01)^{***}  & 0.08 \; (0.01)^{***}  & 0.06 \; (0.01)^{***}  & \textbf{0.11} \; (0.01)^{***}  \\
\quad p:JMI                            & \textbf{0.07} \; (0.01)^{***}  & \textbf{0.09} \; (0.01)^{***}  & \textbf{0.07} \; (0.01)^{***}  & \textbf{0.12} \; (0.02)^{***}  \\
\quad p:JMIM                           & 0.04 \; (0.00)^{***}  & 0.07 \; (0.01)^{***}  & 0.03 \; (0.01)^{**}   & 0.07 \; (0.01)^{***}  \\
\quad p:MIM                            & 0.04 \; (0.00)^{***}  & 0.06 \; (0.01)^{***}  & 0.04 \; (0.01)^{***}  & 0.05 \; (0.01)^{***}  \\
\quad p:MRMR                           & -0.02 \; (0.00)^{***} & -0.08 \; (0.01)^{***} & -0.04 \; (0.01)^{***} & 0.01 \; (0.01)        \\
\quad p:NJMIM                          & 0.06 \; (0.01)^{***}  & 0.08 \; (0.01)^{***}  & 0.05 \; (0.01)^{***}  & 0.08 \; (0.01)^{***}  \\
Controls                               &                       &                       &                       &                       \\
\quad classnoise                       & -0.44 \; (0.01)^{***} &                       &                       &                       \\
\quad attributenoise                   & -0.23 \; (0.01)^{***} &                       &                       &                       \\
\quad classifierRandom Forest          & 0.04 \; (0.00)^{***}  & 0.11 \; (0.00)^{***}  & 0.03 \; (0.00)^{***}  & 0.05 \; (0.00)^{***}  \\
\quad classifierSupport Vector Machine & 0.04 \; (0.00)^{***}  & 0.12 \; (0.00)^{***}  & 0.03 \; (0.00)^{***}  & 0.05 \; (0.00)^{***}  \\
\quad num\_redundant\_features         &                       & 0.00 \; (0.00)^{***}  &                       &                       \\
\quad minClassDev                      &                       &                       & -0.03 \; (0.00)^{***} &                       \\
\quad relFeatObs                       &                       &                       &                       & -0.06 \; (0.00)^{***} \\
\midrule
R$^2$                                  & 0.75                  & 0.78                  & 0.65                  & 0.66                  \\
Adj. R$^2$                             & 0.74                  & 0.77                  & 0.65                  & 0.65                  \\
Num. obs.                              & 5250                  & 2250                  & 2520                  & 2235                  \\
F statistic                            & 288.71                & 148.30                & 76.42                 & 80.89                 \\
\bottomrule
\multicolumn{5}{l}{\scriptsize{Asterisks indicate statistical significance ($^{***}p<0.001$; $^{**}p<0.01$; $^{*}p<0.05$), standard}}\\
\multicolumn{5}{l}{\scriptsize{errors are in parentheses}}\\
\end{tabular}

%% file: tables/models_runtime.tex
\begin{tabular}{l D{)}{)}{10)3} D{)}{)}{10)3} D{)}{)}{10)3} D{)}{)}{13)3}}
\toprule
 & \multicolumn{1}{c}{Noise} & \multicolumn{1}{c}{Redundant} & \multicolumn{1}{c}{Imbalanced} & \multicolumn{1}{c}{Dimensionality} \\
\midrule
(Intercept)                      & 0.00 \; (0.00)^{***}   & 0.00 \; (0.00)^{***}   & 0.00 \; (0.00)^{***}   & 0.00 \;   (0.00)^{***}   \\
FSMs                             &                        &                        &                        &                          \\
\quad C:Accuracy                 & 0.09 \; (0.00)^{***}   & 0.08 \; (0.00)^{***}   & 0.08 \; (0.00)^{***}   & 0.66 \;   (0.11)^{***}   \\
\quad C:DistAngle                & 0.10 \; (0.00)^{***}   & 0.09 \; (0.00)^{***}   & 0.09 \; (0.00)^{***}   & 0.72 \;   (0.12)^{***}   \\
\quad C:DistAUC                  & 0.09 \; (0.00)^{***}   & 0.08 \; (0.00)^{***}   & 0.08 \; (0.00)^{***}   & 0.70 \;   (0.12)^{***}   \\
\quad C:DistEuclid               & 0.09 \; (0.00)^{***}   & 0.09 \; (0.00)^{***}   & 0.09 \; (0.00)^{***}   & 0.69 \;   (0.11)^{***}   \\
\quad C:DistHellinger            & 0.09 \; (0.00)^{***}   & 0.08 \; (0.00)^{***}   & 0.09 \; (0.00)^{***}   & 0.69 \;   (0.11)^{***}   \\
\quad C:DKM                      & 0.09 \; (0.00)^{***}   & 0.08 \; (0.00)^{***}   & 0.08 \; (0.00)^{***}   & 0.74 \;   (0.12)^{***}   \\
\quad C:DKMcost                  &                        &                        & 0.10 \; (0.01)^{***}   &                          \\
\quad C:EqualDKM                 & 0.09 \; (0.00)^{***}   & 0.09 \; (0.00)^{***}   & 0.08 \; (0.00)^{***}   & 0.86 \;   (0.21)^{***}   \\
\quad C:EqualGini                & 0.09 \; (0.00)^{***}   & 0.09 \; (0.00)^{***}   & 0.08 \; (0.00)^{***}   & 0.69 \;   (0.11)^{***}   \\
\quad C:EqualHellinger           & 0.09 \; (0.00)^{***}   & 0.08 \; (0.00)^{***}   & 0.08 \; (0.00)^{***}   & 0.70 \;   (0.12)^{***}   \\
\quad C:EqualInf                 & 0.09 \; (0.00)^{***}   & 0.09 \; (0.00)^{***}   & 0.08 \; (0.00)^{***}   & 0.69 \;   (0.12)^{***}   \\
\quad C:GainRatio                & 0.10 \; (0.00)^{***}   & 0.09 \; (0.00)^{***}   & 0.09 \; (0.00)^{***}   & 0.71 \;   (0.12)^{***}   \\
\quad C:GainRatioCost            &                        &                        & 0.10 \; (0.00)^{***}   &                          \\
\quad C:Gini                     & 0.09 \; (0.00)^{***}   & 0.09 \; (0.00)^{***}   & 0.09 \; (0.00)^{***}   & 0.70 \;   (0.12)^{***}   \\
\quad C:ImpurityEuclid           & 0.09 \; (0.00)^{***}   & 0.09 \; (0.00)^{***}   & 0.08 \; (0.00)^{***}   & 0.69 \;   (0.11)^{***}   \\
\quad C:ImpurityHellinger        & 0.09 \; (0.00)^{***}   & 0.08 \; (0.00)^{***}   & 0.08 \; (0.00)^{***}   & 0.70 \;   (0.11)^{***}   \\
\quad C:InfGain                  & 0.09 \; (0.00)^{***}   & 0.09 \; (0.00)^{***}   & 0.10 \; (0.00)^{***}   & 0.71 \;   (0.12)^{***}   \\
\quad C:MDL                      & 0.12 \; (0.00)^{***}   & 0.11 \; (0.00)^{***}   & 0.11 \; (0.00)^{***}   & 0.78 \;   (0.13)^{***}   \\
\quad C:MDLsmp                   &                        &                        & 0.10 \; (0.00)^{***}   &                          \\
\quad C:MyopicReliefF            & 0.10 \; (0.00)^{***}   & 0.09 \; (0.00)^{***}   & 0.09 \; (0.00)^{***}   & 0.71 \;   (0.11)^{***}   \\
\quad C:Relief                   & 9.61 \; (0.08)^{***}   & 9.25 \; (0.01)^{***}   & 9.15 \; (0.01)^{***}   & 27.44 \;   (7.20)^{***}  \\
\quad C:ReliefFavgC              &                        &                        & 9.24 \; (0.01)^{***}   &                          \\
\quad C:ReliefFbestK             & 11.76 \; (0.08)^{***}  & 11.43 \; (0.03)^{***}  & 11.61 \; (0.04)^{***}  & 38.53 \;  (10.28)^{***}  \\
\quad C:ReliefFdistance          & 1.93 \; (0.02)^{***}   & 1.87 \; (0.02)^{***}   & 1.85 \; (0.02)^{***}   & 7.67 \;   (2.06)^{***}   \\
\quad C:ReliefFequalK            & 1.90 \; (0.01)^{***}   & 1.85 \; (0.01)^{***}   & 1.83 \; (0.01)^{***}   & 7.54 \;   (2.04)^{***}   \\
\quad C:ReliefFexpC              &                        &                        & 9.24 \; (0.01)^{***}   &                          \\
\quad C:ReliefFexpRank           & 2.00 \; (0.02)^{***}   & 1.88 \; (0.01)^{***}   & 1.88 \; (0.01)^{***}   & 7.55 \;   (1.99)^{***}   \\
\quad C:ReliefFmerit             & 9.76 \; (0.08)^{***}   & 9.40 \; (0.01)^{***}   & 9.33 \; (0.01)^{***}   & 27.98 \;   (7.33)^{***}  \\
\quad C:ReliefFpa                &                        &                        & 9.25 \; (0.01)^{***}   &                          \\
\quad C:ReliefFpe                &                        &                        & 9.24 \; (0.01)^{***}   &                          \\
\quad C:ReliefFsmp               &                        &                        & 9.24 \; (0.01)^{***}   &                          \\
\quad C:ReliefFsqrDistance       & 1.92 \; (0.02)^{***}   & 1.86 \; (0.01)^{***}   & 1.84 \; (0.01)^{***}   & 7.73 \;   (2.10)^{***}   \\
\quad C:ReliefKukar              &                        &                        & 9.10 \; (0.01)^{***}   &                          \\
\quad C:UniformAccuracy          & 0.09 \; (0.00)^{***}   & 0.09 \; (0.00)^{***}   & 0.09 \; (0.00)^{***}   & 0.69 \;   (0.11)^{***}   \\
\quad C:UniformDKM               & 0.10 \; (0.00)^{***}   & 0.09 \; (0.00)^{***}   & 0.09 \; (0.00)^{***}   & 0.72 \;   (0.12)^{***}   \\
\quad C:UniformGini              & 0.09 \; (0.00)^{***}   & 0.08 \; (0.00)^{***}   & 0.08 \; (0.00)^{***}   & 0.70 \;   (0.12)^{***}   \\
\quad C:UniformInf               & 0.10 \; (0.00)^{***}   & 0.10 \; (0.00)^{***}   & 0.09 \; (0.00)^{***}   & 0.71 \;   (0.12)^{***}   \\
\quad F:cfs                      & 15.04 \; (0.37)^{***}  & 6.68 \; (1.04)^{***}   & 14.80 \; (0.88)^{***}  & 1500.61 \; (479.84)^{**} \\
\quad F:chi.squared              & 0.61 \; (0.00)^{***}   & 0.60 \; (0.00)^{***}   & 0.60 \; (0.00)^{***}   & 3.13 \;   (0.49)^{***}   \\
\quad F:consistency              & 157.66 \; (1.14)^{***} & 153.38 \; (1.06)^{***} & 143.97 \; (2.76)^{***} & 403.67 \;  (61.52)^{***} \\
\quad F:gain.ratio               & 0.68 \; (0.00)^{***}   & 0.68 \; (0.00)^{***}   & 0.68 \; (0.00)^{***}   & 3.43 \;   (0.52)^{***}   \\
\quad F:oneR                     & 0.67 \; (0.00)^{***}   & 0.66 \; (0.00)^{***}   & 0.66 \; (0.00)^{***}   & 3.53 \;   (0.56)^{***}   \\
\quad F:random.forest.importance & 37.76 \; (0.06)^{***}  & 36.97 \; (0.15)^{***}  & 37.37 \; (0.08)^{***}  & 66.49 \;  (13.66)^{***}  \\
\quad F:relief                   & 82.42 \; (0.22)^{***}  & 83.11 \; (0.28)^{***}  & 82.78 \; (0.29)^{***}  & 472.34 \;  (72.48)^{***} \\
\quad F:symmetrical.uncertainty  & 0.68 \; (0.00)^{***}   & 0.68 \; (0.00)^{***}   & 0.69 \; (0.01)^{***}   & 3.48 \;   (0.54)^{***}   \\
\quad m:anova                    & 0.20 \; (0.00)^{***}   & 0.20 \; (0.01)^{***}   & 0.20 \; (0.01)^{***}   & 1.00 \;   (0.18)^{***}   \\
\quad m:auc                      & 0.05 \; (0.00)^{***}   & 0.05 \; (0.00)^{***}   & 0.05 \; (0.00)^{***}   & 0.13 \;   (0.02)^{***}   \\
\quad m:kruskal.test             & 1.34 \; (0.01)^{***}   & 1.33 \; (0.01)^{***}   & 1.35 \; (0.01)^{***}   & 3.97 \;   (0.57)^{***}   \\
\quad m:ranger\_impurity         & 7.85 \; (0.01)^{***}   & 7.55 \; (0.04)^{***}   & 7.34 \; (0.09)^{***}   & 8.96 \;   (1.60)^{***}   \\
\quad m:ranger\_permutation      & 14.10 \; (0.02)^{***}  & 13.66 \; (0.07)^{***}  & 13.72 \; (0.06)^{***}  & 23.40 \;   (4.53)^{***}  \\
\quad p:CMIM                     & 0.09 \; (0.00)^{***}   & 0.09 \; (0.00)^{***}   & 0.10 \; (0.00)^{***}   & 2.30 \;   (0.52)^{***}   \\
\quad p:DISR                     & 0.05 \; (0.00)^{***}   & 0.05 \; (0.00)^{***}   & 0.05 \; (0.00)^{***}   & 0.50 \;   (0.09)^{***}   \\
\quad p:JIM                      & \textbf{0.01} \; (0.00)^{***}   & \textbf{0.01} \; (0.00)^{***}   & \textbf{0.01} \; (0.00)^{***}   & \textbf{0.02} \;   (0.00)^{***}   \\
\quad p:JMI                      & 0.04 \; (0.00)^{***}   & 0.04 \; (0.00)^{***}   & 0.04 \; (0.00)^{***}   & 0.38 \;   (0.07)^{***}   \\
\quad p:JMIM                     & 0.06 \; (0.00)^{***}   & 0.06 \; (0.00)^{***}   & 0.06 \; (0.00)^{***}   & 0.87 \;   (0.16)^{***}   \\
\quad p:MIM                      & \textbf{0.02} \; (0.00)^{***}   & \textbf{0.02} \; (0.00)^{***}   & \textbf{0.02} \; (0.00)^{***}   & \textbf{0.03} \;   (0.00)^{***}   \\
\quad p:MRMR                     & 0.04 \; (0.00)^{***}   & 0.03 \; (0.00)^{***}   & 0.03 \; (0.00)^{***}   & 0.25 \;   (0.04)^{***}   \\
\quad p:NJMIM                    & 0.07 \; (0.00)^{***}   & 0.06 \; (0.00)^{***}   & 0.07 \; (0.00)^{***}   & 1.16 \;   (0.23)^{***}   \\
\midrule
R$^2$                            & 1.00                   & 1.00                   & 1.00                   & 0.52                     \\
Adj. R$^2$                       & 1.00                   & 1.00                   & 1.00                   & 0.48                     \\
Num. obs.                        & 1750                   & 750                    & 840                    & 745                      \\
F statistic                      & 20426.65               & 12355.62               & 2913.56                & 15.25                    \\
\bottomrule
\multicolumn{5}{l}{\scriptsize{Asterisks indicate statistical significance ($^{***}p<0.001$; $^{**}p<0.01$; $^{*}p<0.05$), standard}}\\
\multicolumn{5}{l}{\scriptsize{errors are in parentheses}}\\
\end{tabular}

%% file: figures/runtime_auc.tex
\begin{tikzpicture}[x=1pt,y=1pt, scale = 0.8, every node/.style={scale=0.8}]
\definecolor{fillColor}{RGB}{255,255,255}
\path[use as bounding box,fill=fillColor,fill opacity=0.00] (0,0) rectangle (469.76,361.35);
\begin{scope}
\path[clip] (  0.00,  0.00) rectangle (469.76,361.35);
\definecolor{drawColor}{RGB}{255,255,255}
\definecolor{fillColor}{RGB}{255,255,255}

\path[draw=drawColor,line width= 0.6pt,line join=round,line cap=round,fill=fillColor] (  0.00,  0.00) rectangle (469.76,361.35);
\end{scope}
\begin{scope}
\path[clip] ( 50.81, 70.56) rectangle (464.26,355.85);
\definecolor{fillColor}{RGB}{255,255,255}

\path[fill=fillColor] ( 50.81, 70.56) rectangle (464.26,355.85);
\definecolor{drawColor}{gray}{0.92}

\path[draw=drawColor,line width= 0.3pt,line join=round] ( 50.81,128.91) --
	(464.26,128.91);

\path[draw=drawColor,line width= 0.3pt,line join=round] ( 50.81,193.75) --
	(464.26,193.75);

\path[draw=drawColor,line width= 0.3pt,line join=round] ( 50.81,258.59) --
	(464.26,258.59);

\path[draw=drawColor,line width= 0.3pt,line join=round] ( 50.81,323.43) --
	(464.26,323.43);

\path[draw=drawColor,line width= 0.3pt,line join=round] ( 69.60, 70.56) --
	( 69.60,355.85);

\path[draw=drawColor,line width= 0.3pt,line join=round] (208.81, 70.56) --
	(208.81,355.85);

\path[draw=drawColor,line width= 0.3pt,line join=round] (348.02, 70.56) --
	(348.02,355.85);

\path[draw=drawColor,line width= 0.6pt,line join=round] ( 50.81, 96.49) --
	(464.26, 96.49);

\path[draw=drawColor,line width= 0.6pt,line join=round] ( 50.81,161.33) --
	(464.26,161.33);

\path[draw=drawColor,line width= 0.6pt,line join=round] ( 50.81,226.17) --
	(464.26,226.17);

\path[draw=drawColor,line width= 0.6pt,line join=round] ( 50.81,291.01) --
	(464.26,291.01);

\path[draw=drawColor,line width= 0.6pt,line join=round] ( 50.81,355.85) --
	(464.26,355.85);

\path[draw=drawColor,line width= 0.6pt,line join=round] (139.20, 70.56) --
	(139.20,355.85);

\path[draw=drawColor,line width= 0.6pt,line join=round] (278.41, 70.56) --
	(278.41,355.85);

\path[draw=drawColor,line width= 0.6pt,line join=round] (417.62, 70.56) --
	(417.62,355.85);
\definecolor{drawColor}{RGB}{0,0,0}
\definecolor{fillColor}{RGB}{0,0,0}

\path[draw=drawColor,line width= 0.4pt,line join=round,line cap=round,fill=fillColor] (139.47,234.01) circle (  1.96);

\path[draw=drawColor,line width= 0.4pt,line join=round,line cap=round,fill=fillColor] (139.49,200.20) circle (  1.96);

\path[draw=drawColor,line width= 0.4pt,line join=round,line cap=round,fill=fillColor] (139.49,236.07) circle (  1.96);

\path[draw=drawColor,line width= 0.4pt,line join=round,line cap=round,fill=fillColor] (139.49,236.49) circle (  1.96);

\path[draw=drawColor,line width= 0.4pt,line join=round,line cap=round,fill=fillColor] (139.49,264.21) circle (  1.96);

\path[draw=drawColor,line width= 0.4pt,line join=round,line cap=round,fill=fillColor] (139.50,264.21) circle (  1.96);

\path[draw=drawColor,line width= 0.4pt,line join=round,line cap=round,fill=fillColor] (139.34,171.28) circle (  1.96);

\path[draw=drawColor,line width= 0.4pt,line join=round,line cap=round,fill=fillColor] (139.52,158.33) circle (  1.96);

\path[draw=drawColor,line width= 0.4pt,line join=round,line cap=round,fill=fillColor] (139.48,158.33) circle (  1.96);

\path[draw=drawColor,line width= 0.4pt,line join=round,line cap=round,fill=fillColor] (139.49,166.26) circle (  1.96);

\path[draw=drawColor,line width= 0.4pt,line join=round,line cap=round,fill=fillColor] (139.48,158.33) circle (  1.96);

\path[draw=drawColor,line width= 0.4pt,line join=round,line cap=round,fill=fillColor] (139.50,168.27) circle (  1.96);

\path[draw=drawColor,line width= 0.4pt,line join=round,line cap=round,fill=fillColor] (139.34,103.41) circle (  1.96);

\path[draw=drawColor,line width= 0.4pt,line join=round,line cap=round,fill=fillColor] (139.49,269.76) circle (  1.96);

\path[draw=drawColor,line width= 0.4pt,line join=round,line cap=round,fill=fillColor] (139.48,158.33) circle (  1.96);

\path[draw=drawColor,line width= 0.4pt,line join=round,line cap=round,fill=fillColor] (139.48,158.33) circle (  1.96);

\path[draw=drawColor,line width= 0.4pt,line join=round,line cap=round,fill=fillColor] (139.50,273.04) circle (  1.96);

\path[draw=drawColor,line width= 0.4pt,line join=round,line cap=round,fill=fillColor] (139.54,269.98) circle (  1.96);

\path[draw=drawColor,line width= 0.4pt,line join=round,line cap=round,fill=fillColor] (139.34,141.21) circle (  1.96);

\path[draw=drawColor,line width= 0.4pt,line join=round,line cap=round,fill=fillColor] (139.50,242.78) circle (  1.96);
\definecolor{drawColor}{RGB}{0,100,0}
\definecolor{fillColor}{RGB}{0,100,0}

\path[draw=drawColor,line width= 0.4pt,line join=round,line cap=round,fill=fillColor] (157.02,139.25) circle (  1.96);
\definecolor{drawColor}{RGB}{0,0,255}
\definecolor{fillColor}{RGB}{0,0,255}

\path[draw=drawColor,line width= 0.4pt,line join=round,line cap=round,fill=fillColor] (152.06,170.14) circle (  1.96);
\definecolor{drawColor}{RGB}{0,100,0}
\definecolor{fillColor}{RGB}{0,100,0}

\path[draw=drawColor,line width= 0.4pt,line join=round,line cap=round,fill=fillColor] (162.44,155.05) circle (  1.96);
\definecolor{drawColor}{RGB}{0,0,255}
\definecolor{fillColor}{RGB}{0,0,255}

\path[draw=drawColor,line width= 0.4pt,line join=round,line cap=round,fill=fillColor] (143.35,248.86) circle (  1.96);

\path[draw=drawColor,line width= 0.4pt,line join=round,line cap=round,fill=fillColor] (143.29,204.81) circle (  1.96);

\path[draw=drawColor,line width= 0.4pt,line join=round,line cap=round,fill=fillColor] (152.07,170.14) circle (  1.96);

\path[draw=drawColor,line width= 0.4pt,line join=round,line cap=round,fill=fillColor] (143.38,225.75) circle (  1.96);
\definecolor{drawColor}{RGB}{0,100,0}
\definecolor{fillColor}{RGB}{0,100,0}

\path[draw=drawColor,line width= 0.4pt,line join=round,line cap=round,fill=fillColor] (157.34,227.08) circle (  1.96);
\definecolor{drawColor}{RGB}{0,0,255}
\definecolor{fillColor}{RGB}{0,0,255}

\path[draw=drawColor,line width= 0.4pt,line join=round,line cap=round,fill=fillColor] (152.08,247.75) circle (  1.96);

\path[draw=drawColor,line width= 0.4pt,line join=round,line cap=round,fill=fillColor] (152.06,247.75) circle (  1.96);

\path[draw=drawColor,line width= 0.4pt,line join=round,line cap=round,fill=fillColor] (152.07,156.96) circle (  1.96);

\path[draw=drawColor,line width= 0.4pt,line join=round,line cap=round,fill=fillColor] (143.36,249.20) circle (  1.96);

\path[draw=drawColor,line width= 0.4pt,line join=round,line cap=round,fill=fillColor] (151.88,164.30) circle (  1.96);
\definecolor{drawColor}{RGB}{0,0,0}
\definecolor{fillColor}{RGB}{0,0,0}

\path[draw=drawColor,line width= 0.4pt,line join=round,line cap=round,fill=fillColor] (139.49,234.88) circle (  1.96);

\path[draw=drawColor,line width= 0.4pt,line join=round,line cap=round,fill=fillColor] (139.50,264.21) circle (  1.96);

\path[draw=drawColor,line width= 0.4pt,line join=round,line cap=round,fill=fillColor] (139.49,267.98) circle (  1.96);

\path[draw=drawColor,line width= 0.4pt,line join=round,line cap=round,fill=fillColor] (139.50,270.66) circle (  1.96);
\definecolor{drawColor}{RGB}{255,0,0}
\definecolor{fillColor}{RGB}{255,0,0}

\path[draw=drawColor,line width= 0.4pt,line join=round,line cap=round,fill=fillColor] (433.48,263.59) circle (  1.96);
\definecolor{drawColor}{RGB}{0,0,255}
\definecolor{fillColor}{RGB}{0,0,255}

\path[draw=drawColor,line width= 0.4pt,line join=round,line cap=round,fill=fillColor] (140.71,261.55) circle (  1.96);
\definecolor{drawColor}{RGB}{255,0,0}
\definecolor{fillColor}{RGB}{255,0,0}

\path[draw=drawColor,line width= 0.4pt,line join=round,line cap=round,fill=fillColor] (418.20,215.11) circle (  1.96);
\definecolor{drawColor}{RGB}{0,0,255}
\definecolor{fillColor}{RGB}{0,0,255}

\path[draw=drawColor,line width= 0.4pt,line join=round,line cap=round,fill=fillColor] (140.87,262.43) circle (  1.96);

\path[draw=drawColor,line width= 0.4pt,line join=round,line cap=round,fill=fillColor] (140.88,240.26) circle (  1.96);
\definecolor{drawColor}{RGB}{0,100,0}
\definecolor{fillColor}{RGB}{0,100,0}

\path[draw=drawColor,line width= 0.4pt,line join=round,line cap=round,fill=fillColor] (198.96,315.67) circle (  1.96);
\definecolor{drawColor}{RGB}{255,0,0}
\definecolor{fillColor}{RGB}{255,0,0}

\path[draw=drawColor,line width= 0.4pt,line join=round,line cap=round,fill=fillColor] (355.99, 86.61) circle (  1.96);
\definecolor{drawColor}{RGB}{0,0,255}
\definecolor{fillColor}{RGB}{0,0,255}

\path[draw=drawColor,line width= 0.4pt,line join=round,line cap=round,fill=fillColor] (140.88,262.44) circle (  1.96);
\definecolor{drawColor}{RGB}{0,0,0}
\definecolor{fillColor}{RGB}{0,0,0}

\path[draw=drawColor,line width= 0.4pt,line join=round,line cap=round,fill=fillColor] (139.69,186.48) circle (  1.96);

\path[draw=drawColor,line width= 0.4pt,line join=round,line cap=round,fill=fillColor] (139.29,181.76) circle (  1.96);
\definecolor{drawColor}{RGB}{0,0,255}
\definecolor{fillColor}{RGB}{0,0,255}

\path[draw=drawColor,line width= 0.4pt,line join=round,line cap=round,fill=fillColor] (141.75,181.76) circle (  1.96);

\path[draw=drawColor,line width= 0.4pt,line join=round,line cap=round,fill=fillColor] (150.21,307.39) circle (  1.96);
\definecolor{drawColor}{RGB}{0,100,0}
\definecolor{fillColor}{RGB}{0,100,0}

\path[draw=drawColor,line width= 0.4pt,line join=round,line cap=round,fill=fillColor] (161.04,307.69) circle (  1.96);
\definecolor{drawColor}{RGB}{0,0,0}
\definecolor{fillColor}{RGB}{0,0,0}

\path[draw=drawColor,line width= 0.4pt,line join=round,line cap=round,fill=fillColor] (139.91,260.36) circle (  1.96);

\path[draw=drawColor,line width= 0.4pt,line join=round,line cap=round,fill=fillColor] (139.39,298.97) circle (  1.96);

\path[draw=drawColor,line width= 0.4pt,line join=round,line cap=round,fill=fillColor] (139.22,280.74) circle (  1.96);

\path[draw=drawColor,line width= 0.4pt,line join=round,line cap=round,fill=fillColor] (139.35,285.67) circle (  1.96);

\path[draw=drawColor,line width= 0.4pt,line join=round,line cap=round,fill=fillColor] (139.50,263.26) circle (  1.96);

\path[draw=drawColor,line width= 0.4pt,line join=round,line cap=round,fill=fillColor] (139.23,260.36) circle (  1.96);

\path[draw=drawColor,line width= 0.4pt,line join=round,line cap=round,fill=fillColor] (139.31,178.00) circle (  1.96);

\path[draw=drawColor,line width= 0.4pt,line join=round,line cap=round,fill=fillColor] (139.58,273.91) circle (  1.96);

\path[draw=drawColor,line width= 0.6pt,line join=round,line cap=round] (134.25,216.24) -- (139.47,234.01);

\node[text=drawColor,anchor=base,inner sep=0pt, outer sep=0pt, scale=  1.10] at (134.14,207.14) {C:Accuracy};

\path[draw=drawColor,line width= 0.6pt,line join=round,line cap=round] (113.80,210.06) -- (139.49,200.20);

\node[text=drawColor,anchor=base,inner sep=0pt, outer sep=0pt, scale=  1.10] at ( 82.76,206.81) {C:DistAngle};

\path[draw=drawColor,line width= 0.6pt,line join=round,line cap=round] (129.78,228.71) -- (139.49,236.07);

\node[text=drawColor,anchor=base,inner sep=0pt, outer sep=0pt, scale=  1.10] at (116.03,219.60) {C:DistAUC};

\path[draw=drawColor,line width= 0.6pt,line join=round,line cap=round] (116.71,229.60) -- (139.49,236.49);

\node[text=drawColor,anchor=base,inner sep=0pt, outer sep=0pt, scale=  1.10] at ( 84.21,225.34) {C:DistEuclid};

\path[draw=drawColor,line width= 0.6pt,line join=round,line cap=round] (130.37,266.81) -- (139.49,264.21);

\node[text=drawColor,anchor=base,inner sep=0pt, outer sep=0pt, scale=  1.10] at ( 91.64,263.26) {C:DistHellinger};

\path[draw=drawColor,line width= 0.6pt,line join=round,line cap=round] ( 92.89,279.21) -- (139.50,264.21);

\node[text=drawColor,anchor=base,inner sep=0pt, outer sep=0pt, scale=  1.10] at ( 72.30,275.69) {C:DKM};

\path[draw=drawColor,line width= 0.6pt,line join=round,line cap=round] (125.14,183.95) -- (139.34,171.28);

\node[text=drawColor,anchor=base,inner sep=0pt, outer sep=0pt, scale=  1.10] at ( 95.01,181.39) {C:DKMcost};

\path[draw=drawColor,line width= 0.6pt,line join=round,line cap=round] (194.53,135.55) -- (139.52,158.33);

\node[text=drawColor,anchor=base,inner sep=0pt, outer sep=0pt, scale=  1.10] at (229.15,131.45) {C:EqualDKM};

\path[draw=drawColor,line width= 0.6pt,line join=round,line cap=round] (114.74,148.01) -- (139.48,158.33);

\node[text=drawColor,anchor=base,inner sep=0pt, outer sep=0pt, scale=  1.10] at ( 83.22,143.60) {C:EqualGini};

\node[text=drawColor,anchor=base,inner sep=0pt, outer sep=0pt, scale=  1.10] at ( 95.21,156.18) {C:EqualHellinger};

\path[draw=drawColor,line width= 0.6pt,line join=round,line cap=round] (123.82,140.26) -- (139.48,158.33);

\node[text=drawColor,anchor=base,inner sep=0pt, outer sep=0pt, scale=  1.10] at (119.94,131.15) {C:EqualInf};

\node[text=drawColor,anchor=base,inner sep=0pt, outer sep=0pt, scale=  1.10] at (106.66,168.83) {C:GainRatio};

\node[text=drawColor,anchor=base,inner sep=0pt, outer sep=0pt, scale=  1.10] at (148.61, 93.26) {C:GainRatioCost};

\path[draw=drawColor,line width= 0.6pt,line join=round,line cap=round] (122.28,299.61) -- (139.49,269.76);

\node[text=drawColor,anchor=base,inner sep=0pt, outer sep=0pt, scale=  1.10] at (120.86,301.12) {C:Gini};

\path[draw=drawColor,line width= 0.6pt,line join=round,line cap=round] (139.36,123.32) -- (139.48,158.33);

\node[text=drawColor,anchor=base,inner sep=0pt, outer sep=0pt, scale=  1.10] at ( 95.60,118.73) {C:ImpurityEuclid};

\path[draw=drawColor,line width= 0.6pt,line join=round,line cap=round] (146.38,152.96) -- (139.48,158.33);

\node[text=drawColor,anchor=base,inner sep=0pt, outer sep=0pt, scale=  1.10] at (154.39,143.85) {C:ImpurityHellinger};

\path[draw=drawColor,line width= 0.6pt,line join=round,line cap=round] (154.55,279.04) -- (139.50,273.04);

\node[text=drawColor,anchor=base,inner sep=0pt, outer sep=0pt, scale=  1.10] at (180.01,276.06) {C:InfGain};

\node[text=drawColor,anchor=base,inner sep=0pt, outer sep=0pt, scale=  1.10] at (161.54,263.40) {C:MDL};

\path[draw=drawColor,line width= 0.6pt,line join=round,line cap=round] (141.15,121.61) -- (139.34,141.21);

\node[text=drawColor,anchor=base,inner sep=0pt, outer sep=0pt, scale=  1.10] at (170.73,116.47) {C:MDLsmp};

\node[text=drawColor,anchor=base,inner sep=0pt, outer sep=0pt, scale=  1.10] at (115.01,238.23) {C:MyopicReliefF};
\definecolor{drawColor}{RGB}{0,100,0}

\node[text=drawColor,anchor=base,inner sep=0pt, outer sep=0pt, scale=  1.10] at (170.43,129.25) {C:Relief};
\definecolor{drawColor}{RGB}{0,0,255}

\path[draw=drawColor,line width= 0.6pt,line join=round,line cap=round] (245.09,186.81) -- (152.06,170.14);

\node[text=drawColor,anchor=base,inner sep=0pt, outer sep=0pt, scale=  1.10] at (281.24,183.10) {C:ReliefFavgC};
\definecolor{drawColor}{RGB}{0,100,0}

\path[draw=drawColor,line width= 0.6pt,line join=round,line cap=round] (193.98,149.55) -- (162.44,155.05);

\node[text=drawColor,anchor=base,inner sep=0pt, outer sep=0pt, scale=  1.10] at (232.61,145.58) {C:ReliefFbestK};
\definecolor{drawColor}{RGB}{0,0,255}

\node[text=drawColor,anchor=base,inner sep=0pt, outer sep=0pt, scale=  1.10] at ( 95.25,238.03) {C:ReliefFdistance};

\path[draw=drawColor,line width= 0.6pt,line join=round,line cap=round] (173.18,204.57) -- (143.29,204.81);

\node[text=drawColor,anchor=base,inner sep=0pt, outer sep=0pt, scale=  1.10] at (214.08,200.77) {C:ReliefFequalK};

\path[draw=drawColor,line width= 0.6pt,line join=round,line cap=round] (193.14,174.41) -- (152.07,170.14);

\node[text=drawColor,anchor=base,inner sep=0pt, outer sep=0pt, scale=  1.10] at (229.45,170.66) {C:ReliefFexpC};

\path[draw=drawColor,line width= 0.6pt,line join=round,line cap=round] (151.59,219.10) -- (143.38,225.75);

\node[text=drawColor,anchor=base,inner sep=0pt, outer sep=0pt, scale=  1.10] at (196.71,213.37) {C:ReliefFexpRank};
\definecolor{drawColor}{RGB}{0,100,0}

\path[draw=drawColor,line width= 0.6pt,line join=round,line cap=round] (244.38,217.23) -- (157.34,227.08);

\node[text=drawColor,anchor=base,inner sep=0pt, outer sep=0pt, scale=  1.10] at (281.16,213.37) {C:ReliefFmerit};
\definecolor{drawColor}{RGB}{0,0,255}

\path[draw=drawColor,line width= 0.6pt,line join=round,line cap=round] (284.71,229.89) -- (152.08,247.75);

\node[text=drawColor,anchor=base,inner sep=0pt, outer sep=0pt, scale=  1.10] at (314.42,226.04) {C:ReliefFpa};

\path[draw=drawColor,line width= 0.6pt,line join=round,line cap=round] (158.06,244.38) -- (152.06,247.75);

\node[text=drawColor,anchor=base,inner sep=0pt, outer sep=0pt, scale=  1.10] at (187.62,238.47) {C:ReliefFpe};

\path[draw=drawColor,line width= 0.6pt,line join=round,line cap=round] (273.09,154.31) -- (152.07,156.96);

\node[text=drawColor,anchor=base,inner sep=0pt, outer sep=0pt, scale=  1.10] at (306.82,150.50) {C:ReliefFsmp};

\path[draw=drawColor,line width= 0.6pt,line join=round,line cap=round] (218.84,242.37) -- (143.36,249.20);

\node[text=drawColor,anchor=base,inner sep=0pt, outer sep=0pt, scale=  1.10] at (270.78,238.53) {C:ReliefFsqrDistance};

\node[text=drawColor,anchor=base,inner sep=0pt, outer sep=0pt, scale=  1.10] at (188.52,158.22) {C:ReliefKukar};
\definecolor{drawColor}{RGB}{0,0,0}

\path[draw=drawColor,line width= 0.6pt,line join=round,line cap=round] (158.48,230.17) -- (139.49,234.88);

\node[text=drawColor,anchor=base,inner sep=0pt, outer sep=0pt, scale=  1.10] at (207.69,226.03) {C:UniformAccuracy};

\path[draw=drawColor,line width= 0.6pt,line join=round,line cap=round] (126.22,274.39) -- (139.50,264.21);

\node[text=drawColor,anchor=base,inner sep=0pt, outer sep=0pt, scale=  1.10] at (109.00,275.89) {C:UniformDKM};

\path[draw=drawColor,line width= 0.6pt,line join=round,line cap=round] (129.06,291.12) -- (139.49,267.98);

\node[text=drawColor,anchor=base,inner sep=0pt, outer sep=0pt, scale=  1.10] at ( 91.63,288.38) {C:UniformGini};

\path[draw=drawColor,line width= 0.6pt,line join=round,line cap=round] (207.01,279.77) -- (139.50,270.66);

\node[text=drawColor,anchor=base,inner sep=0pt, outer sep=0pt, scale=  1.10] at (240.72,276.05) {C:UniformInf};
\definecolor{drawColor}{RGB}{255,0,0}

\node[text=drawColor,anchor=base,inner sep=0pt, outer sep=0pt, scale=  1.10] at (442.94,266.41) {F:cfs};
\definecolor{drawColor}{RGB}{0,0,255}

\node[text=drawColor,anchor=base,inner sep=0pt, outer sep=0pt, scale=  1.10] at (177.84,251.21) {F:chi.squared};
\definecolor{drawColor}{RGB}{255,0,0}

\node[text=drawColor,anchor=base,inner sep=0pt, outer sep=0pt, scale=  1.10] at (427.32,217.53) {F:consistency};
\definecolor{drawColor}{RGB}{0,0,255}

\path[draw=drawColor,line width= 0.6pt,line join=round,line cap=round] (206.02,254.82) -- (140.87,262.43);

\node[text=drawColor,anchor=base,inner sep=0pt, outer sep=0pt, scale=  1.10] at (235.67,250.95) {F:gain.ratio};

\path[draw=drawColor,line width= 0.6pt,line join=round,line cap=round] (244.95,229.76) -- (140.88,240.26);

\node[text=drawColor,anchor=base,inner sep=0pt, outer sep=0pt, scale=  1.10] at (263.93,225.91) {F:oneR};
\definecolor{drawColor}{RGB}{0,100,0}

\path[draw=drawColor,line width= 0.6pt,line join=round,line cap=round] (209.97,323.40) -- (198.96,315.67);

\node[text=drawColor,anchor=base,inner sep=0pt, outer sep=0pt, scale=  1.10] at (259.58,324.91) {F:random.forest.importance};
\definecolor{drawColor}{RGB}{255,0,0}

\node[text=drawColor,anchor=base,inner sep=0pt, outer sep=0pt, scale=  1.10] at (365.41, 89.13) {F:relief};
\definecolor{drawColor}{RGB}{0,0,255}

\path[draw=drawColor,line width= 0.6pt,line join=round,line cap=round] (183.72,267.14) -- (140.88,262.44);

\node[text=drawColor,anchor=base,inner sep=0pt, outer sep=0pt, scale=  1.10] at (248.68,263.40) {F:symmetrical.uncertainty};
\definecolor{drawColor}{RGB}{0,0,0}

\node[text=drawColor,anchor=base,inner sep=0pt, outer sep=0pt, scale=  1.10] at (149.60,189.06) {m:anova};

\path[draw=drawColor,line width= 0.6pt,line join=round,line cap=round] (126.36,196.69) -- (139.29,181.76);

\node[text=drawColor,anchor=base,inner sep=0pt, outer sep=0pt, scale=  1.10] at (110.44,194.15) {m:auc};
\definecolor{drawColor}{RGB}{0,0,255}

\path[draw=drawColor,line width= 0.6pt,line join=round,line cap=round] (172.81,188.80) -- (141.75,181.76);

\node[text=drawColor,anchor=base,inner sep=0pt, outer sep=0pt, scale=  1.10] at (208.11,185.26) {m:kruskal.test};

\path[draw=drawColor,line width= 0.6pt,line join=round,line cap=round] (154.22,312.36) -- (150.21,307.39);

\node[text=drawColor,anchor=base,inner sep=0pt, outer sep=0pt, scale=  1.10] at (155.04,313.87) {m:ranger\_impurity};
\definecolor{drawColor}{RGB}{0,100,0}

\path[draw=drawColor,line width= 0.6pt,line join=round,line cap=round] (204.96,307.21) -- (161.04,307.69);

\node[text=drawColor,anchor=base,inner sep=0pt, outer sep=0pt, scale=  1.10] at (262.00,303.41) {m:ranger\_permutation};
\definecolor{drawColor}{RGB}{0,0,0}

\node[text=drawColor,anchor=base,inner sep=0pt, outer sep=0pt, scale=  1.10] at (120.60,250.67) {p:CMIM};

\path[draw=drawColor,line width= 0.6pt,line join=round,line cap=round] (127.26,329.28) -- (139.39,298.97);

\node[text=drawColor,anchor=base,inner sep=0pt, outer sep=0pt, scale=  1.10] at (107.82,326.32) {p:DISR};

\path[draw=drawColor,line width= 0.6pt,line join=round,line cap=round] (162.16,304.23) -- (139.22,280.74);

\node[text=drawColor,anchor=base,inner sep=0pt, outer sep=0pt, scale=  1.10] at (178.15,301.21) {p:JIM};

\node[text=drawColor,anchor=base,inner sep=0pt, outer sep=0pt, scale=  1.10] at (153.71,288.61) {p:JMI};

\path[draw=drawColor,line width= 0.6pt,line join=round,line cap=round] ( 93.81,254.58) -- (139.50,263.26);

\node[text=drawColor,anchor=base,inner sep=0pt, outer sep=0pt, scale=  1.10] at ( 72.76,250.61) {p:JMIM};

\path[draw=drawColor,line width= 0.6pt,line join=round,line cap=round] (115.05,254.87) -- (139.23,260.36);

\node[text=drawColor,anchor=base,inner sep=0pt, outer sep=0pt, scale=  1.10] at ( 96.83,250.78) {p:MIM};

\node[text=drawColor,anchor=base,inner sep=0pt, outer sep=0pt, scale=  1.10] at (167.01,172.75) {p:MRMR};

\path[draw=drawColor,line width= 0.6pt,line join=round,line cap=round] (172.80,291.87) -- (139.58,273.91);

\node[text=drawColor,anchor=base,inner sep=0pt, outer sep=0pt, scale=  1.10] at (197.99,288.61) {p:NJMIM};
\definecolor{drawColor}{gray}{0.20}

\path[draw=drawColor,line width= 0.6pt,line join=round,line cap=round] ( 50.81, 70.56) rectangle (464.26,355.85);
\end{scope}
\begin{scope}
\path[clip] (  0.00,  0.00) rectangle (469.76,361.35);
\definecolor{drawColor}{RGB}{0,0,0}

\node[text=drawColor,anchor=base east,inner sep=0pt, outer sep=0pt, scale=  0.96] at ( 45.86, 93.17) {\bfseries 0.525};

\node[text=drawColor,anchor=base east,inner sep=0pt, outer sep=0pt, scale=  0.96] at ( 45.86,158.01) {\bfseries 0.550};

\node[text=drawColor,anchor=base east,inner sep=0pt, outer sep=0pt, scale=  0.96] at ( 45.86,222.85) {\bfseries 0.575};

\node[text=drawColor,anchor=base east,inner sep=0pt, outer sep=0pt, scale=  0.96] at ( 45.86,287.69) {\bfseries 0.600};

\node[text=drawColor,anchor=base east,inner sep=0pt, outer sep=0pt, scale=  0.96] at ( 45.86,352.53) {\bfseries 0.625};
\end{scope}
\begin{scope}
\path[clip] (  0.00,  0.00) rectangle (469.76,361.35);
\definecolor{drawColor}{gray}{0.20}

\path[draw=drawColor,line width= 0.6pt,line join=round] ( 48.06, 96.49) --
	( 50.81, 96.49);

\path[draw=drawColor,line width= 0.6pt,line join=round] ( 48.06,161.33) --
	( 50.81,161.33);

\path[draw=drawColor,line width= 0.6pt,line join=round] ( 48.06,226.17) --
	( 50.81,226.17);

\path[draw=drawColor,line width= 0.6pt,line join=round] ( 48.06,291.01) --
	( 50.81,291.01);

\path[draw=drawColor,line width= 0.6pt,line join=round] ( 48.06,355.85) --
	( 50.81,355.85);
\end{scope}
\begin{scope}
\path[clip] (  0.00,  0.00) rectangle (469.76,361.35);
\definecolor{drawColor}{gray}{0.20}

\path[draw=drawColor,line width= 0.6pt,line join=round] (139.20, 67.81) --
	(139.20, 70.56);

\path[draw=drawColor,line width= 0.6pt,line join=round] (278.41, 67.81) --
	(278.41, 70.56);

\path[draw=drawColor,line width= 0.6pt,line join=round] (417.62, 67.81) --
	(417.62, 70.56);
\end{scope}
\begin{scope}
\path[clip] (  0.00,  0.00) rectangle (469.76,361.35);
\definecolor{drawColor}{RGB}{0,0,0}

\node[text=drawColor,anchor=base,inner sep=0pt, outer sep=0pt, scale=  0.96] at (139.20, 58.96) {\bfseries 0};

\node[text=drawColor,anchor=base,inner sep=0pt, outer sep=0pt, scale=  0.96] at (278.41, 58.96) {\bfseries 100};

\node[text=drawColor,anchor=base,inner sep=0pt, outer sep=0pt, scale=  0.96] at (417.62, 58.96) {\bfseries 200};
\end{scope}
\begin{scope}
\path[clip] (  0.00,  0.00) rectangle (469.76,361.35);
\definecolor{drawColor}{RGB}{0,0,0}

\node[text=drawColor,anchor=base,inner sep=0pt, outer sep=0pt, scale=  1.40] at (257.53, 44.68) {\bfseries Runtime in seconds};
\end{scope}
\begin{scope}
\path[clip] (  0.00,  0.00) rectangle (469.76,361.35);
\definecolor{drawColor}{RGB}{0,0,0}

\node[text=drawColor,rotate= 90.00,anchor=base,inner sep=0pt, outer sep=0pt, scale=  1.40] at ( 15.16,213.20) {\bfseries Predictive performance (AUC)};
\end{scope}
\begin{scope}
\path[clip] (  0.00,  0.00) rectangle (469.76,361.35);
\definecolor{fillColor}{RGB}{255,255,255}

\path[fill=fillColor] ( 99.63,  5.50) rectangle (415.44, 30.95);
\end{scope}
\begin{scope}
\path[clip] (  0.00,  0.00) rectangle (469.76,361.35);
\definecolor{drawColor}{RGB}{0,0,0}

\node[text=drawColor,anchor=base west,inner sep=0pt, outer sep=0pt, scale=  1.00] at (105.13, 14.78) {\bfseries Runtime:};
\end{scope}
\begin{scope}
\path[clip] (  0.00,  0.00) rectangle (469.76,361.35);
\definecolor{fillColor}{RGB}{255,255,255}

\path[fill=fillColor] (156.91, 11.00) rectangle (171.37, 25.45);
\end{scope}
\begin{scope}
\path[clip] (  0.00,  0.00) rectangle (469.76,361.35);
\definecolor{drawColor}{RGB}{0,0,0}
\definecolor{fillColor}{RGB}{0,0,0}

\path[draw=drawColor,line width= 0.4pt,line join=round,line cap=round,fill=fillColor] (164.14, 18.23) circle (  4.64);
\end{scope}
\begin{scope}
\path[clip] (  0.00,  0.00) rectangle (469.76,361.35);
\definecolor{drawColor}{RGB}{0,0,0}

\node[text=drawColor,anchor=base,inner sep=0pt, outer sep=0pt, scale=  1.14] at (164.14, 14.31) {a};
\end{scope}
\begin{scope}
\path[clip] (  0.00,  0.00) rectangle (469.76,361.35);
\definecolor{fillColor}{RGB}{255,255,255}

\path[fill=fillColor] (213.09, 11.00) rectangle (227.54, 25.45);
\end{scope}
\begin{scope}
\path[clip] (  0.00,  0.00) rectangle (469.76,361.35);
\definecolor{drawColor}{RGB}{0,0,255}
\definecolor{fillColor}{RGB}{0,0,255}

\path[draw=drawColor,line width= 0.4pt,line join=round,line cap=round,fill=fillColor] (220.31, 18.23) circle (  4.64);
\end{scope}
\begin{scope}
\path[clip] (  0.00,  0.00) rectangle (469.76,361.35);
\definecolor{drawColor}{RGB}{0,0,255}

\node[text=drawColor,anchor=base,inner sep=0pt, outer sep=0pt, scale=  1.14] at (220.31, 14.31) {a};
\end{scope}
\begin{scope}
\path[clip] (  0.00,  0.00) rectangle (469.76,361.35);
\definecolor{fillColor}{RGB}{255,255,255}

\path[fill=fillColor] (280.92, 11.00) rectangle (295.38, 25.45);
\end{scope}
\begin{scope}
\path[clip] (  0.00,  0.00) rectangle (469.76,361.35);
\definecolor{drawColor}{RGB}{0,100,0}
\definecolor{fillColor}{RGB}{0,100,0}

\path[draw=drawColor,line width= 0.4pt,line join=round,line cap=round,fill=fillColor] (288.15, 18.23) circle (  4.64);
\end{scope}
\begin{scope}
\path[clip] (  0.00,  0.00) rectangle (469.76,361.35);
\definecolor{drawColor}{RGB}{0,100,0}

\node[text=drawColor,anchor=base,inner sep=0pt, outer sep=0pt, scale=  1.14] at (288.15, 14.31) {a};
\end{scope}
\begin{scope}
\path[clip] (  0.00,  0.00) rectangle (469.76,361.35);
\definecolor{fillColor}{RGB}{255,255,255}

\path[fill=fillColor] (353.76, 11.00) rectangle (368.22, 25.45);
\end{scope}
\begin{scope}
\path[clip] (  0.00,  0.00) rectangle (469.76,361.35);
\definecolor{drawColor}{RGB}{255,0,0}
\definecolor{fillColor}{RGB}{255,0,0}

\path[draw=drawColor,line width= 0.4pt,line join=round,line cap=round,fill=fillColor] (360.99, 18.23) circle (  4.64);
\end{scope}
\begin{scope}
\path[clip] (  0.00,  0.00) rectangle (469.76,361.35);
\definecolor{drawColor}{RGB}{255,0,0}

\node[text=drawColor,anchor=base,inner sep=0pt, outer sep=0pt, scale=  1.14] at (360.99, 14.31) {a};
\end{scope}
\begin{scope}
\path[clip] (  0.00,  0.00) rectangle (469.76,361.35);
\definecolor{drawColor}{RGB}{0,0,0}

\node[text=drawColor,anchor=base west,inner sep=0pt, outer sep=0pt, scale=  1.40] at (176.37, 14.78) {$<1$ sec};
\end{scope}
\begin{scope}
\path[clip] (  0.00,  0.00) rectangle (469.76,361.35);
\definecolor{drawColor}{RGB}{0,0,0}

\node[text=drawColor,anchor=base west,inner sep=0pt, outer sep=0pt, scale=  1.40] at (232.54, 14.78) {$1-10$ sec};
\end{scope}
\begin{scope}
\path[clip] (  0.00,  0.00) rectangle (469.76,361.35);
\definecolor{drawColor}{RGB}{0,0,0}

\node[text=drawColor,anchor=base west,inner sep=0pt, outer sep=0pt, scale=  1.40] at (300.38, 14.78) {$10-60$ sec};
\end{scope}
\begin{scope}
\path[clip] (  0.00,  0.00) rectangle (469.76,361.35);
\definecolor{drawColor}{RGB}{0,0,0}

\node[text=drawColor,anchor=base west,inner sep=0pt, outer sep=0pt, scale=  1.40] at (373.22, 14.78) {$>60$ sec};
\end{scope}
\end{tikzpicture}

%% file: tables/benchmark_summary.tex
\begin{longtable}{lcccc|cccc|cccc|c}
	\toprule
	\hiderowcolors
	Method 	& \multicolumn{4}{c|}{Pred. performance} & \multicolumn{4}{c|}{Relevant features} & \multicolumn{4}{c|}{Stability} & Run-\\
	& Noi & Red & Imb & Dim & 	Noi & Red & Imb & Dim & 	Noi & Red & Imb & Dim &  time \\ 
	\showrowcolors
  \midrule
C:Accuracy & + & + & 0 & + & -- & -- & -- & 0 & 0 & 0 & 0 & + & + \\ 
  C:DistAngle & -- & 0 & -- & + & -- & -- & -- & -- & 0 & -- & 0 & 0 & + \\ 
  C:DistAUC & + & + & 0 & + & -- & -- & -- & 0 & + & 0 & -- & + & + \\ 
  C:DistEuclid & + & + & 0 & + & -- & 0 & -- & 0 & + & 0 & -- & ++ & + \\ 
  C:DistHellinger & + & + & + & + & 0 & 0 & 0 & 0 & -- & 0 & + & + & + \\ 
  C:DKM & + & + & + & + & 0 & 0 & 0 & 0 & 0 & + & + & 0 & + \\ 
  C:EqualDKM & -- & -- -- & -- -- & 0 & -- -- & -- -- & -- -- & -- -- & -- -- & -- -- & -- -- & -- & + \\ 
  C:EqualGini & -- & -- -- & -- -- & 0 & -- -- & -- -- & -- -- & -- -- & -- -- & -- -- & -- -- & 0 & + \\ 
  C:EqualHellinger & -- & -- & -- & 0 & -- -- & -- & -- -- & -- -- & -- -- & -- -- & 0 & -- & + \\ 
  C:EqualInf & -- & -- -- & -- -- & 0 & -- -- & -- -- & -- -- & -- -- & -- -- & -- -- & -- -- & 0 & + \\ 
  C:GainRatio & -- & -- & -- & 0 & -- -- & -- -- & -- -- & -- -- & -- -- & -- -- & 0 & -- & + \\ 
  C:Gini & + & + & + & + & + & + & 0 & + & + & + & + & 0 & + \\ 
  C:ImpurityEuclid & -- & -- & -- -- & 0 & -- -- & -- -- & -- -- & -- -- & -- -- & -- -- & -- -- & 0 & + \\ 
  C:ImpurityHellinger & -- -- & -- & -- & 0 & -- -- & -- -- & -- -- & -- -- & -- -- & -- -- & -- -- & -- & + \\ 
  C:InfGain & + & + & + & + & + & + & + & 0 & 0 & + & 0 & 0 & + \\ 
  C:MDL & + & + & + & + & 0 & + & 0 & 0 & 0 & + & 0 & 0 & + \\ 
  C:MyopicReliefF & + & + & 0 & + & -- & -- & -- & 0 & 0 & 0 & + & + & + \\ 
  C:Relief & -- -- & 0 & -- & -- -- & -- -- & -- -- & -- -- & -- -- & -- -- & -- -- & -- -- & -- -- & -- \\ 
  C:ReliefFbestK & -- & 0 & 0 & -- -- & -- -- & -- -- & -- & -- -- & -- -- & -- -- & -- & -- -- & -- \\ 
  C:ReliefFdistance & + & + & + & + & 0 & 0 & + & 0 & 0 & -- & ++ & -- & 0 \\ 
  C:ReliefFequalK & -- & + & 0 & 0 & -- & -- & -- & -- -- & -- -- & -- & -- & -- -- & 0 \\ 
  C:ReliefFexpRank & 0 & + & 0 & 0 & -- & -- & -- & -- & -- & -- & + & -- -- & 0 \\ 
  C:ReliefFmerit & 0 & + & + & 0 & -- & -- & -- & -- & -- & -- & + & -- -- & -- \\ 
  C:ReliefFsqrDistance & + & + & + & + & 0 & 0 & + & 0 & -- & -- & ++ & -- & 0 \\ 
  C:UniformAccuracy & + & + & 0 & + & -- & -- & -- & -- & 0 & 0 & -- & + & + \\ 
  C:UniformDKM & + & + & + & + & 0 & 0 & 0 & 0 & -- & 0 & 0 & 0 & + \\ 
  C:UniformGini & + & + & + & + & + & + & + & + & + & ++ & 0 & 0 & + \\ 
  C:UniformInf & + & + & + & + & + & + & + & + & 0 & + & 0 & 0 & + \\ 
  F:cfs & + & + & + & + & ++ & ++ & ++ & ++ & -- & -- -- & -- -- & -- -- & -- -- \\ 
  F:chi.squared & + & + & + & + & 0 & -- & 0 & -- & -- & 0 & -- & -- -- & 0 \\ 
  F:consistency & 0 & -- & 0 & 0 & -- & -- -- & -- & -- -- & -- -- & -- -- & -- -- & -- -- & -- -- \\ 
  F:gain.ratio & + & + & + & + & 0 & 0 & 0 & 0 & -- & 0 & -- & -- & 0 \\ 
  F:oneR & + & + & 0 & + & -- & -- & -- & -- & 0 & -- & -- -- & -- & 0 \\ 
  F:random.forest.importance & ++ & ++ & ++ & ++ & ++ & ++ & ++ & ++ & ++ & ++ & ++ & -- & -- \\ 
  F:relief & -- -- & -- -- & -- -- & -- -- & -- -- & -- -- & -- -- & -- -- & -- -- & -- -- & -- -- & -- -- & -- -- \\ 
  F:symmetrical.uncertainty & + & + & + & + & 0 & 0 & 0 & -- & -- & 0 & -- & -- & 0 \\ 
  m:anova & -- & -- & -- & + & -- & -- & -- -- & -- & 0 & -- & -- & + & + \\ 
  m:auc & -- & -- & -- & + & -- -- & -- -- & -- -- & -- & + & -- & 0 & ++ & ++ \\ 
  m:kruskal.test & -- & -- & -- & + & -- -- & -- -- & -- -- & -- & + & -- & 0 & ++ & 0 \\ 
  m:ranger\_impurity & ++ & + & + & ++ & ++ & ++ & ++ & ++ & ++ & ++ & ++ & ++ & 0 \\ 
  m:ranger\_permutation & ++ & + & ++ & + & ++ & ++ & ++ & + & ++ & ++ & ++ & -- & -- \\ 
  p:CMIM & + & + & + & + & 0 & 0 & 0 & -- & -- & 0 & -- & -- & + \\ 
  p:DISR & ++ & ++ & ++ & ++ & ++ & ++ & ++ & ++ & ++ & ++ & + & ++ & + \\ 
  p:JIM & + & + & + & + & + & + & + & + & ++ & + & 0 & 0 & ++ \\ 
  p:JMI & + & + & ++ & ++ & + & + & + & ++ & + & + & -- & + & + \\ 
  p:JMIM & + & + & + & + & 0 & 0 & 0 & + & -- & -- & -- -- & -- -- & + \\ 
  p:MIM & + & + & + & + & 0 & 0 & 0 & -- & -- & 0 & -- & -- -- & ++ \\ 
  p:MRMR & -- & -- -- & -- & 0 & -- -- & -- -- & -- -- & -- -- & -- -- & -- -- & -- -- & -- -- & ++ \\ 
  p:NJMIM & + & + & + & + & + & 0 & 0 & + & 0 & -- & -- -- & -- -- & + \\ 
   \bottomrule
\end{longtable}

%% file: tables/models_relev.tex
\begin{table}[h]
\caption{Models for the criterion 'number of relevant features selected' in the four dataset scenarios}
\begin{center}
\begin{tiny}
\begin{tabular}{l D{)}{)}{9)3} D{)}{)}{9)3} D{)}{)}{9)3} D{)}{)}{9)3}}
\toprule
 & \multicolumn{1}{c}{Noise} & \multicolumn{1}{c}{Redundant} & \multicolumn{1}{c}{Imbalanced} & \multicolumn{1}{c}{Dimensionality} \\
\midrule
(Intercept)                      & 1.00 \; (0.00)^{***}  & 1.00                  & 1.00 \; (0.00)^{***}  & 1.00 \; (0.00)^{***}  \\
FSMs                             &                       &                       &                       &                       \\
\quad C:Accuracy                 & -0.58 \; (0.02)^{***} & -0.57 \; (0.03)^{***} & -0.71 \; (0.06)^{***} & -0.63 \; (0.05)^{***} \\
\quad C:DistAngle                & -0.68 \; (0.02)^{***} & -0.66 \; (0.02)^{***} & -0.72 \; (0.03)^{***} & -0.71 \; (0.03)^{***} \\
\quad C:DistAUC                  & -0.59 \; (0.02)^{***} & -0.56 \; (0.03)^{***} & -0.63 \; (0.05)^{***} & -0.63 \; (0.05)^{***} \\
\quad C:DistEuclid               & -0.59 \; (0.02)^{***} & -0.56 \; (0.03)^{***} & -0.63 \; (0.05)^{***} & -0.63 \; (0.05)^{***} \\
\quad C:DistHellinger            & -0.55 \; (0.03)^{***} & -0.52 \; (0.05)^{***} & -0.54 \; (0.06)^{***} & -0.59 \; (0.07)^{***} \\
\quad C:DKM                      & -0.55 \; (0.03)^{***} & -0.52 \; (0.05)^{***} & -0.54 \; (0.06)^{***} & -0.59 \; (0.07)^{***} \\
\quad C:DKMcost                  &                       &                       & -0.62 \; (0.04)^{***} &                       \\
\quad C:EqualDKM                 & -0.83 \; (0.01)^{***} & -0.70 \; (0.00)^{***} & -0.82 \; (0.03)^{***} & -0.84 \; (0.03)^{***} \\
\quad C:EqualGini                & -0.83 \; (0.01)^{***} & -0.70 \; (0.00)^{***} & -0.82 \; (0.03)^{***} & -0.84 \; (0.03)^{***} \\
\quad C:EqualHellinger           & -0.83 \; (0.01)^{***} & -0.70 \; (0.00)^{***} & -0.76 \; (0.02)^{***} & -0.84 \; (0.03)^{***} \\
\quad C:EqualInf                 & -0.83 \; (0.01)^{***} & -0.70 \; (0.00)^{***} & -0.82 \; (0.03)^{***} & -0.84 \; (0.03)^{***} \\
\quad C:GainRatio                & -0.83 \; (0.01)^{***} & -0.70 \; (0.00)^{***} & -0.76 \; (0.02)^{***} & -0.84 \; (0.03)^{***} \\
\quad C:GainRatioCost            &                       &                       & -0.82 \; (0.01)^{***} &                       \\
\quad C:Gini                     & -0.48 \; (0.02)^{***} & -0.49 \; (0.04)^{***} & -0.53 \; (0.05)^{***} & -0.57 \; (0.06)^{***} \\
\quad C:ImpurityEuclid           & -0.83 \; (0.01)^{***} & -0.70 \; (0.00)^{***} & -0.82 \; (0.03)^{***} & -0.84 \; (0.03)^{***} \\
\quad C:ImpurityHellinger        & -0.83 \; (0.01)^{***} & -0.70 \; (0.00)^{***} & -0.82 \; (0.03)^{***} & -0.84 \; (0.03)^{***} \\
\quad C:InfGain                  & -0.50 \; (0.03)^{***} & -0.49 \; (0.04)^{***} & -0.51 \; (0.05)^{***} & -0.57 \; (0.06)^{***} \\
\quad C:MDL                      & -0.52 \; (0.03)^{***} & -0.49 \; (0.04)^{***} & -0.53 \; (0.06)^{***} & -0.58 \; (0.07)^{***} \\
\quad C:MDLsmp                   &                       &                       & -0.74 \; (0.02)^{***} &                       \\
\quad C:MyopicReliefF            & -0.58 \; (0.02)^{***} & -0.56 \; (0.03)^{***} & -0.59 \; (0.03)^{***} & -0.63 \; (0.05)^{***} \\
\quad C:Relief                   & -0.86 \; (0.01)^{***} & -0.75 \; (0.03)^{***} & -0.82 \; (0.01)^{***} & -0.92 \; (0.01)^{***} \\
\quad C:ReliefFavgC              &                       &                       & -0.70 \; (0.03)^{***} &                       \\
\quad C:ReliefFbestK             & -0.85 \; (0.01)^{***} & -0.75 \; (0.02)^{***} & -0.67 \; (0.04)^{***} & -0.91 \; (0.01)^{***} \\
\quad C:ReliefFdistance          & -0.53 \; (0.04)^{***} & -0.56 \; (0.06)^{***} & -0.46 \; (0.04)^{***} & -0.61 \; (0.07)^{***} \\
\quad C:ReliefFequalK            & -0.74 \; (0.02)^{***} & -0.67 \; (0.03)^{***} & -0.61 \; (0.01)^{***} & -0.81 \; (0.03)^{***} \\
\quad C:ReliefFexpC              &                       &                       & -0.70 \; (0.03)^{***} &                       \\
\quad C:ReliefFexpRank           & -0.66 \; (0.02)^{***} & -0.63 \; (0.04)^{***} & -0.58 \; (0.01)^{***} & -0.73 \; (0.04)^{***} \\
\quad C:ReliefFmerit             & -0.66 \; (0.02)^{***} & -0.64 \; (0.04)^{***} & -0.58 \; (0.01)^{***} & -0.73 \; (0.04)^{***} \\
\quad C:ReliefFpa                &                       &                       & -0.60 \; (0.00)^{***} &                       \\
\quad C:ReliefFpe                &                       &                       & -0.60 \; (0.00)^{***} &                       \\
\quad C:ReliefFsmp               &                       &                       & -0.71 \; (0.03)^{***} &                       \\
\quad C:ReliefFsqrDistance       & -0.53 \; (0.04)^{***} & -0.56 \; (0.06)^{***} & -0.46 \; (0.04)^{***} & -0.61 \; (0.07)^{***} \\
\quad C:ReliefKukar              &                       &                       & -0.82 \; (0.01)^{***} &                       \\
\quad C:UniformAccuracy          & -0.59 \; (0.02)^{***} & -0.57 \; (0.03)^{***} & -0.63 \; (0.05)^{***} & -0.63 \; (0.05)^{***} \\
\quad C:UniformDKM               & -0.55 \; (0.03)^{***} & -0.52 \; (0.05)^{***} & -0.54 \; (0.06)^{***} & -0.59 \; (0.07)^{***} \\
\quad C:UniformGini              & -0.48 \; (0.02)^{***} & -0.49 \; (0.04)^{***} & -0.51 \; (0.05)^{***} & -0.57 \; (0.06)^{***} \\
\quad C:UniformInf               & -0.50 \; (0.03)^{***} & -0.49 \; (0.04)^{***} & -0.50 \; (0.05)^{***} & -0.57 \; (0.06)^{***} \\
\quad F:cfs                      & -0.17 \; (0.01)^{***} & -0.44 \; (0.09)^{***} & -0.18 \; (0.03)^{***} & -0.28 \; (0.08)^{***} \\
\quad F:chi.squared              & -0.56 \; (0.02)^{***} & -0.57 \; (0.03)^{***} & -0.57 \; (0.04)^{***} & -0.64 \; (0.06)^{***} \\
\quad F:consistency              & -0.73 \; (0.02)^{***} & -0.78 \; (0.03)^{***} & -0.72 \; (0.03)^{***} & -0.79 \; (0.04)^{***} \\
\quad F:gain.ratio               & -0.54 \; (0.02)^{***} & -0.55 \; (0.04)^{***} & -0.53 \; (0.05)^{***} & -0.63 \; (0.07)^{***} \\
\quad F:oneR                     & -0.62 \; (0.02)^{***} & -0.65 \; (0.03)^{***} & -0.70 \; (0.04)^{***} & -0.69 \; (0.04)^{***} \\
\quad F:random.forest.importance & -0.28 \; (0.04)^{***} & -0.40 \; (0.08)^{***} & -0.18 \; (0.04)^{***} & -0.42 \; (0.10)^{***} \\
\quad F:relief                   & -0.91 \; (0.01)^{***} & -0.91 \; (0.01)^{***} & -0.88 \; (0.01)^{***} & -0.94 \; (0.01)^{***} \\
\quad F:symmetrical.uncertainty  & -0.54 \; (0.02)^{***} & -0.56 \; (0.04)^{***} & -0.54 \; (0.04)^{***} & -0.63 \; (0.06)^{***} \\
\quad m:anova                    & -0.71 \; (0.01)^{***} & -0.68 \; (0.01)^{***} & -0.78 \; (0.02)^{***} & -0.72 \; (0.02)^{***} \\
\quad m:auc                      & -0.74 \; (0.01)^{***} & -0.74 \; (0.01)^{***} & -0.80 \; (0.02)^{***} & -0.74 \; (0.02)^{***} \\
\quad m:kruskal.test             & -0.74 \; (0.01)^{***} & -0.74 \; (0.01)^{***} & -0.80 \; (0.02)^{***} & -0.74 \; (0.02)^{***} \\
\quad m:ranger\_impurity         & -0.30 \; (0.04)^{***} & -0.40 \; (0.07)^{***} & -0.43 \; (0.08)^{***} & -0.41 \; (0.09)^{***} \\
\quad m:ranger\_permutation      & -0.32 \; (0.04)^{***} & -0.42 \; (0.08)^{***} & -0.22 \; (0.04)^{***} & -0.44 \; (0.10)^{***} \\
\quad p:CMIM                     & -0.56 \; (0.02)^{***} & -0.56 \; (0.04)^{***} & -0.55 \; (0.05)^{***} & -0.63 \; (0.06)^{***} \\
\quad p:DISR                     & -0.36 \; (0.04)^{***} & -0.42 \; (0.08)^{***} & -0.39 \; (0.09)^{***} & -0.33 \; (0.12)^{**}  \\
\quad p:JIM                      & -0.46 \; (0.04)^{***} & -0.46 \; (0.06)^{***} & -0.46 \; (0.07)^{***} & -0.43 \; (0.10)^{***} \\
\quad p:JMI                      & -0.45 \; (0.04)^{***} & -0.45 \; (0.07)^{***} & -0.45 \; (0.08)^{***} & -0.40 \; (0.10)^{***} \\
\quad p:JMIM                     & -0.55 \; (0.03)^{***} & -0.52 \; (0.04)^{***} & -0.56 \; (0.06)^{***} & -0.57 \; (0.07)^{***} \\
\quad p:MIM                      & -0.56 \; (0.02)^{***} & -0.56 \; (0.04)^{***} & -0.55 \; (0.05)^{***} & -0.63 \; (0.06)^{***} \\
\quad p:MRMR                     & -0.75 \; (0.01)^{***} & -0.86 \; (0.04)^{***} & -0.78 \; (0.03)^{***} & -0.80 \; (0.03)^{***} \\
\quad p:NJMIM                    & -0.50 \; (0.03)^{***} & -0.50 \; (0.05)^{***} & -0.52 \; (0.07)^{***} & -0.52 \; (0.08)^{***} \\
\midrule
R$^2$                            & 0.65                  & 0.48                  & 0.58                  & 0.41                  \\
Adj. R$^2$                       & 0.64                  & 0.44                  & 0.55                  & 0.37                  \\
Num. obs.                        & 1750                  & 750                   & 840                   & 745                   \\
F statistic                      & 63.71                 & 13.13                 & 18.52                 & 9.76                  \\
\bottomrule
\multicolumn{5}{l}{\tiny{$^{***}p<0.001$; $^{**}p<0.01$; $^{*}p<0.05$}}
\end{tabular}
\end{tiny}
\label{tab:models_relev}
\end{center}
\end{table}

%% file: tables/models_stab.tex
\begin{table}[h]
\caption{Models for the criterion 'stability' in the four dataset scenarios}
\begin{center}
\begin{tiny}
\begin{tabular}{l D{)}{)}{9)3} D{)}{)}{9)3} D{)}{)}{9)3} D{)}{)}{9)3}}
\toprule
 & \multicolumn{1}{c}{Noise} & \multicolumn{1}{c}{Redundant} & \multicolumn{1}{c}{Imbalanced} & \multicolumn{1}{c}{Dimensionality} \\
\midrule
(Intercept)                      & 1.26 \; (0.03)^{***}  & 1.14 \; (0.02)^{***}  & 1.06 \; (0.01)^{***}  & 1.06 \; (0.02)^{***}  \\
FSMs                             &                       &                       &                       &                       \\
\quad C:Accuracy                 & -0.64 \; (0.03)^{***} & -0.31 \; (0.02)^{***} & -0.40 \; (0.02)^{***} & -0.36 \; (0.02)^{***} \\
\quad C:DistAngle                & -0.67 \; (0.03)^{***} & -0.38 \; (0.03)^{***} & -0.40 \; (0.02)^{***} & -0.39 \; (0.03)^{***} \\
\quad C:DistAUC                  & -0.63 \; (0.03)^{***} & -0.31 \; (0.02)^{***} & -0.41 \; (0.02)^{***} & -0.36 \; (0.02)^{***} \\
\quad C:DistEuclid               & -0.63 \; (0.03)^{***} & -0.31 \; (0.02)^{***} & -0.40 \; (0.02)^{***} & -0.36 \; (0.02)^{***} \\
\quad C:DistHellinger            & -0.70 \; (0.03)^{***} & -0.31 \; (0.02)^{***} & -0.34 \; (0.02)^{***} & -0.39 \; (0.02)^{***} \\
\quad C:DKM                      & -0.70 \; (0.03)^{***} & -0.31 \; (0.02)^{***} & -0.34 \; (0.02)^{***} & -0.39 \; (0.02)^{***} \\
\quad C:DKMcost                  &                       &                       & -0.50 \; (0.04)^{***} &                       \\
\quad C:EqualDKM                 & -0.90 \; (0.04)^{***} & -0.51 \; (0.02)^{***} & -0.58 \; (0.02)^{***} & -0.44 \; (0.03)^{***} \\
\quad C:EqualGini                & -0.90 \; (0.04)^{***} & -0.51 \; (0.02)^{***} & -0.58 \; (0.02)^{***} & -0.44 \; (0.03)^{***} \\
\quad C:EqualHellinger           & -0.90 \; (0.04)^{***} & -0.51 \; (0.02)^{***} & -0.40 \; (0.03)^{***} & -0.44 \; (0.03)^{***} \\
\quad C:EqualInf                 & -0.90 \; (0.04)^{***} & -0.51 \; (0.02)^{***} & -0.58 \; (0.02)^{***} & -0.44 \; (0.03)^{***} \\
\quad C:GainRatio                & -0.89 \; (0.04)^{***} & -0.51 \; (0.02)^{***} & -0.40 \; (0.03)^{***} & -0.44 \; (0.03)^{***} \\
\quad C:GainRatioCost            &                       &                       & -0.36 \; (0.02)^{***} &                       \\
\quad C:Gini                     & -0.60 \; (0.03)^{***} & -0.24 \; (0.02)^{***} & -0.33 \; (0.02)^{***} & -0.39 \; (0.03)^{***} \\
\quad C:ImpurityEuclid           & -0.90 \; (0.04)^{***} & -0.51 \; (0.02)^{***} & -0.58 \; (0.02)^{***} & -0.44 \; (0.03)^{***} \\
\quad C:ImpurityHellinger        & -0.90 \; (0.04)^{***} & -0.51 \; (0.02)^{***} & -0.58 \; (0.02)^{***} & -0.44 \; (0.03)^{***} \\
\quad C:InfGain                  & -0.64 \; (0.03)^{***} & -0.26 \; (0.02)^{***} & -0.36 \; (0.03)^{***} & -0.40 \; (0.02)^{***} \\
\quad C:MDL                      & -0.66 \; (0.03)^{***} & -0.27 \; (0.02)^{***} & -0.35 \; (0.03)^{***} & -0.40 \; (0.02)^{***} \\
\quad C:MDLsmp                   &                       &                       & -0.70 \; (0.03)^{***} &                       \\
\quad C:MyopicReliefF            & -0.64 \; (0.03)^{***} & -0.31 \; (0.02)^{***} & -0.32 \; (0.02)^{***} & -0.36 \; (0.02)^{***} \\
\quad C:Relief                   & -0.91 \; (0.04)^{***} & -0.55 \; (0.04)^{***} & -0.54 \; (0.03)^{***} & -0.63 \; (0.02)^{***} \\
\quad C:ReliefFavgC              &                       &                       & -0.40 \; (0.02)^{***} &                       \\
\quad C:ReliefFbestK             & -0.91 \; (0.04)^{***} & -0.55 \; (0.04)^{***} & -0.41 \; (0.05)^{***} & -0.62 \; (0.02)^{***} \\
\quad C:ReliefFdistance          & -0.70 \; (0.03)^{***} & -0.34 \; (0.02)^{***} & -0.26 \; (0.03)^{***} & -0.47 \; (0.03)^{***} \\
\quad C:ReliefFequalK            & -0.87 \; (0.03)^{***} & -0.44 \; (0.04)^{***} & -0.41 \; (0.04)^{***} & -0.58 \; (0.03)^{***} \\
\quad C:ReliefFexpC              &                       &                       & -0.40 \; (0.02)^{***} &                       \\
\quad C:ReliefFexpRank           & -0.81 \; (0.03)^{***} & -0.39 \; (0.04)^{***} & -0.33 \; (0.04)^{***} & -0.50 \; (0.03)^{***} \\
\quad C:ReliefFmerit             & -0.81 \; (0.03)^{***} & -0.40 \; (0.03)^{***} & -0.32 \; (0.04)^{***} & -0.50 \; (0.02)^{***} \\
\quad C:ReliefFpa                &                       &                       & -0.22 \; (0.02)^{***} &                       \\
\quad C:ReliefFpe                &                       &                       & -0.22 \; (0.02)^{***} &                       \\
\quad C:ReliefFsmp               &                       &                       & -0.43 \; (0.02)^{***} &                       \\
\quad C:ReliefFsqrDistance       & -0.70 \; (0.03)^{***} & -0.34 \; (0.01)^{***} & -0.26 \; (0.03)^{***} & -0.47 \; (0.03)^{***} \\
\quad C:ReliefKukar              &                       &                       & -0.54 \; (0.02)^{***} &                       \\
\quad C:UniformAccuracy          & -0.63 \; (0.03)^{***} & -0.31 \; (0.02)^{***} & -0.41 \; (0.02)^{***} & -0.36 \; (0.02)^{***} \\
\quad C:UniformDKM               & -0.70 \; (0.03)^{***} & -0.31 \; (0.02)^{***} & -0.34 \; (0.02)^{***} & -0.39 \; (0.02)^{***} \\
\quad C:UniformGini              & -0.60 \; (0.03)^{***} & -0.24 \; (0.02)^{***} & -0.34 \; (0.03)^{***} & -0.39 \; (0.03)^{***} \\
\quad C:UniformInf               & -0.64 \; (0.03)^{***} & -0.26 \; (0.02)^{***} & -0.37 \; (0.03)^{***} & -0.40 \; (0.02)^{***} \\
\quad F:cfs                      & -0.81 \; (0.03)^{***} & -0.45 \; (0.01)^{***} & -0.46 \; (0.01)^{***} & -0.62 \; (0.05)^{***} \\
\quad F:chi.squared              & -0.71 \; (0.03)^{***} & -0.33 \; (0.03)^{***} & -0.41 \; (0.02)^{***} & -0.50 \; (0.02)^{***} \\
\quad F:consistency              & -0.85 \; (0.04)^{***} & -0.85 \; (0.02)^{***} & -0.79 \; (0.02)^{***} & -0.77 \; (0.03)^{***} \\
\quad F:gain.ratio               & -0.70 \; (0.03)^{***} & -0.31 \; (0.02)^{***} & -0.41 \; (0.02)^{***} & -0.49 \; (0.03)^{***} \\
\quad F:oneR                     & -0.70 \; (0.03)^{***} & -0.34 \; (0.02)^{***} & -0.46 \; (0.04)^{***} & -0.49 \; (0.03)^{***} \\
\quad F:random.forest.importance & -0.42 \; (0.04)^{***} & -0.18 \; (0.04)^{***} & -0.19 \; (0.03)^{***} & -0.48 \; (0.09)^{***} \\
\quad F:relief                   & -1.00 \; (0.04)^{***} & -1.11 \; (0.02)^{***} & -0.99 \; (0.02)^{***} & -0.99 \; (0.03)^{***} \\
\quad F:symmetrical.uncertainty  & -0.71 \; (0.03)^{***} & -0.32 \; (0.03)^{***} & -0.41 \; (0.02)^{***} & -0.48 \; (0.02)^{***} \\
\quad m:anova                    & -0.66 \; (0.03)^{***} & -0.42 \; (0.02)^{***} & -0.41 \; (0.02)^{***} & -0.37 \; (0.03)^{***} \\
\quad m:auc                      & -0.61 \; (0.03)^{***} & -0.40 \; (0.02)^{***} & -0.39 \; (0.02)^{***} & -0.35 \; (0.02)^{***} \\
\quad m:kruskal.test             & -0.61 \; (0.03)^{***} & -0.40 \; (0.02)^{***} & -0.39 \; (0.02)^{***} & -0.35 \; (0.02)^{***} \\
\quad m:ranger\_impurity         & -0.42 \; (0.04)^{***} & -0.18 \; (0.03)^{***} & -0.21 \; (0.02)^{***} & -0.29 \; (0.05)^{***} \\
\quad m:ranger\_permutation      & -0.47 \; (0.04)^{***} & -0.20 \; (0.03)^{***} & -0.25 \; (0.03)^{***} & -0.47 \; (0.07)^{***} \\
\quad p:CMIM                     & -0.71 \; (0.03)^{***} & -0.32 \; (0.03)^{***} & -0.43 \; (0.02)^{***} & -0.49 \; (0.02)^{***} \\
\quad p:DISR                     & -0.51 \; (0.04)^{***} & -0.19 \; (0.04)^{***} & -0.34 \; (0.06)^{***} & -0.26 \; (0.07)^{***} \\
\quad p:JIM                      & -0.59 \; (0.03)^{***} & -0.29 \; (0.03)^{***} & -0.40 \; (0.02)^{***} & -0.40 \; (0.04)^{***} \\
\quad p:JMI                      & -0.60 \; (0.03)^{***} & -0.27 \; (0.03)^{***} & -0.41 \; (0.04)^{***} & -0.36 \; (0.05)^{***} \\
\quad p:JMIM                     & -0.71 \; (0.03)^{***} & -0.44 \; (0.02)^{***} & -0.57 \; (0.03)^{***} & -0.58 \; (0.03)^{***} \\
\quad p:MIM                      & -0.71 \; (0.03)^{***} & -0.32 \; (0.03)^{***} & -0.43 \; (0.02)^{***} & -0.49 \; (0.02)^{***} \\
\quad p:MRMR                     & -0.88 \; (0.03)^{***} & -0.75 \; (0.02)^{***} & -0.69 \; (0.02)^{***} & -0.63 \; (0.04)^{***} \\
\quad p:NJMIM                    & -0.68 \; (0.03)^{***} & -0.40 \; (0.02)^{***} & -0.55 \; (0.03)^{***} & -0.54 \; (0.04)^{***} \\
Controls                         &                       &                       &                       &                       \\
\quad classnoise                 & -1.83 \; (0.03)^{***} &                       &                       &                       \\
\quad attributenoise             & -1.17 \; (0.03)^{***} &                       &                       &                       \\
\quad mean\_informativeFeatures  &                       & -0.01 \; (0.00)^{***} &                       &                       \\
\quad minClassDev                &                       &                       & -0.03 \; (0.00)^{***} &                       \\
\quad relFeatObs                 &                       &                       &                       & -0.09 \; (0.00)^{***} \\
\midrule
R$^2$                            & 0.85                  & 0.83                  & 0.76                  & 0.72                  \\
Adj. R$^2$                       & 0.84                  & 0.82                  & 0.74                  & 0.70                  \\
Num. obs.                        & 1750                  & 750                   & 840                   & 745                   \\
F statistic                      & 181.85                & 68.70                 & 41.50                 & 36.50                 \\
\bottomrule
\multicolumn{5}{l}{\tiny{$^{***}p<0.001$; $^{**}p<0.01$; $^{*}p<0.05$}}
\end{tabular}
\end{tiny}
\label{tab:models_stab}
\end{center}
\end{table}

%% file: FSM_benchmark_arXiv.bbl
\begin{thebibliography}{100}
\providecommand{\natexlab}[1]{#1}
\providecommand{\url}[1]{\texttt{#1}}
\expandafter\ifx\csname urlstyle\endcsname\relax
  \providecommand{\doi}[1]{doi: #1}\else
  \providecommand{\doi}{doi: \begingroup \urlstyle{rm}\Url}\fi

\bibitem[Adams et~al.(2017)Adams, Meekins, and Beling]{adams_empirical_2017}
Stephen Adams, Ryan Meekins, and Peter~A. Beling.
\newblock An {Empirical} {Evaluation} of {Techniques} for {Feature} {Selection}
  with {Cost}.
\newblock In \emph{2017 {IEEE} {International} {Conference} on {Data} {Mining}
  {Workshops} ({ICDMW})}, pages 834--841, New Orleans, LA, November 2017. IEEE.
\newblock ISBN 978-1-5386-3800-2.
\newblock \doi{10.1109/ICDMW.2017.153}.
\newblock URL \url{http://ieeexplore.ieee.org/document/8215748/}.

\bibitem[Alelyani et~al.(2011)Alelyani, Zhao, and Liu]{alelyani_dilemma_2011}
Salem Alelyani, Zheng Zhao, and Huan Liu.
\newblock A {Dilemma} in {Assessing} {Stability} of {Feature} {Selection}
  {Algorithms}.
\newblock In \emph{2011 {IEEE} {International} {Conference} on {High}
  {Performance} {Computing} and {Communications}}, pages 701--707, Banff, AB,
  Canada, September 2011. IEEE.
\newblock ISBN 978-1-4577-1564-8.
\newblock \doi{10.1109/HPCC.2011.99}.
\newblock URL \url{http://ieeexplore.ieee.org/document/6063062/}.

\bibitem[Alonso(2015)]{alonso_challenges_2015}
Omar Alonso.
\newblock Challenges with label quality for supervised learning.
\newblock \emph{Journal of Data and Information Quality}, 6\penalty0
  (1):\penalty0 2:1--2:3, 2015.
\newblock \doi{10.1145/2724721}.
\newblock URL \url{https://doi.org/10.1145/2724721}.

\bibitem[Aphinyanaphongs et~al.(2014)Aphinyanaphongs, Fu, Li, Peskin,
  Efstathiadis, Aliferis, and Statnikov]{aphinyanaphongs_comprehensive_2014}
Yindalon Aphinyanaphongs, Lawrence~D. Fu, Zhiguo Li, Eric~R. Peskin, Efstratios
  Efstathiadis, Constantin~F. Aliferis, and Alexander Statnikov.
\newblock A comprehensive empirical comparison of modern supervised
  classification and feature selection methods for text categorization: {A}
  {Comprehensive} {Empirical} {Comparison} of {Modern} {Supervised}
  {Classification} and {Feature}-{Selection} {Methods} for {Text}
  {Categorization}.
\newblock \emph{Journal of the Association for Information Science and
  Technology}, 65\penalty0 (10):\penalty0 1964--1987, October 2014.
\newblock \doi{10.1002/asi.23110}.
\newblock URL \url{http://doi.wiley.com/10.1002/asi.23110}.

\bibitem[Battiti(1994)]{battiti_using_1994}
R.~Battiti.
\newblock Using mutual information for selecting features in supervised neural
  net learning.
\newblock \emph{{IEEE} Transactions on Neural Networks}, 5\penalty0
  (4):\penalty0 537--550, July 1994.
\newblock \doi{10.1109/72.298224}.

\bibitem[Baumann et~al.(2019)Baumann, Hochbaum, and
  Yang]{baumann_comparative_2019}
P.~Baumann, D.S. Hochbaum, and Y.T. Yang.
\newblock A comparative study of the leading machine learning techniques and
  two new optimization algorithms.
\newblock \emph{European Journal of Operational Research}, 272\penalty0
  (3):\penalty0 1041--1057, 2019.
\newblock \doi{10.1016/j.ejor.2018.07.009}.
\newblock URL
  \url{https://linkinghub.elsevier.com/retrieve/pii/S0377221718306143}.

\bibitem[Bennasar et~al.(2015)Bennasar, Hicks, and
  Setchi]{bennasar_feature_2015}
Mohamed Bennasar, Yulia Hicks, and Rossitza Setchi.
\newblock Feature selection using {Joint} {Mutual} {Information}
  {Maximisation}.
\newblock \emph{Expert Systems with Applications}, 42\penalty0 (22):\penalty0
  8520--8532, December 2015.
\newblock \doi{10.1016/j.eswa.2015.07.007}.
\newblock URL
  \url{https://linkinghub.elsevier.com/retrieve/pii/S0957417415004674}.

\bibitem[Berente et~al.(2019)Berente, Seidel, and
  Safadi]{berente_research_2019}
Nicholas Berente, Stefan Seidel, and Hani Safadi.
\newblock Research commentary—data-driven computationally intensive theory
  development.
\newblock \emph{Information Systems Research}, 30\penalty0 (1):\penalty0
  50--64, 2019.
\newblock \doi{10.1287/isre.2018.0774}.

\bibitem[Bischl et~al.(2016)Bischl, Lang, Kotthoff, Schiffner, Richter,
  Studerus, Casalicchio, and Jones]{mlr}
Bernd Bischl, Michel Lang, Lars Kotthoff, Julia Schiffner, Jakob Richter, Erich
  Studerus, Giuseppe Casalicchio, and Zachary~M. Jones.
\newblock {mlr}: Machine learning in r.
\newblock \emph{Journal of Machine Learning Research}, 17\penalty0
  (170):\penalty0 1--5, 2016.
\newblock URL \url{http://jmlr.org/papers/v17/15-066.html}.

\bibitem[Blum and Langley(1997)]{blum_selection_1997}
Avrim~L. Blum and Pat Langley.
\newblock Selection of relevant features and examples in machine learning.
\newblock \emph{Artificial Intelligence}, 97\penalty0 (1–2):\penalty0
  245--271, December 1997.
\newblock \doi{10.1016/S0004-3702(97)00063-5}.
\newblock URL
  \url{//www.sciencedirect.com/science/article/pii/S0004370297000635}.

\bibitem[Bolón-Canedo et~al.(2013)Bolón-Canedo, Sánchez-Maroño, and
  Alonso-Betanzos]{bolon-canedo_review_2013}
Verónica Bolón-Canedo, Noelia Sánchez-Maroño, and Amparo Alonso-Betanzos.
\newblock A review of feature selection methods on synthetic data.
\newblock \emph{Knowledge and Information Systems}, 34\penalty0 (3):\penalty0
  483--519, March 2013.
\newblock \doi{10.1007/s10115-012-0487-8}.
\newblock URL \url{http://link.springer.com/10.1007/s10115-012-0487-8}.

\bibitem[Bolón-Canedo et~al.(2014)Bolón-Canedo, Sánchez-Maroño,
  Alonso-Betanzos, Benítez, and Herrera]{bolon-canedo_review_2014}
Verónica Bolón-Canedo, Noelia Sánchez-Maroño, Amparo Alonso-Betanzos, J.M.
  Benítez, and F.~Herrera.
\newblock A review of microarray datasets and applied feature selection
  methods.
\newblock \emph{Information Sciences}, 282:\penalty0 111--135, October 2014.
\newblock \doi{10.1016/j.ins.2014.05.042}.
\newblock URL
  \url{https://linkinghub.elsevier.com/retrieve/pii/S0020025514006021}.

\bibitem[Bommert et~al.(2020)Bommert, Sun, Bischl, Rahnenführer, and
  Lang]{bommert_benchmark_2020}
Andrea Bommert, Xudong Sun, Bernd Bischl, Jörg Rahnenführer, and Michel Lang.
\newblock Benchmark for filter methods for feature selection in
  high-dimensional classification data.
\newblock \emph{Computational Statistics \& Data Analysis}, 143:\penalty0
  106839, March 2020.
\newblock \doi{10.1016/j.csda.2019.106839}.
\newblock URL
  \url{https://linkinghub.elsevier.com/retrieve/pii/S016794731930194X}.

\bibitem[Borisov et~al.(2019)Borisov, Haug, and
  Kasneci]{borisov_cancelout_2019}
Vadim Borisov, Johannes Haug, and Gjergji Kasneci.
\newblock {CancelOut}: {A} {Layer} for {Feature} {Selection} in {Deep} {Neural}
  {Networks}.
\newblock In Igor~V. Tetko, Věra Kůrková, Pavel Karpov, and Fabian Theis,
  editors, \emph{Artificial {Neural} {Networks} and {Machine} {Learning} –
  {ICANN} 2019: {Deep} {Learning}}, Lecture {Notes} in {Computer} {Science},
  pages 72--83, Cham, 2019. Springer International Publishing.
\newblock ISBN 978-3-030-30484-3.
\newblock \doi{10.1007/978-3-030-30484-3_6}.

\bibitem[Bosu and {MacDonell}(2013)]{bosu_taxonomy_2013}
Michael~Franklin Bosu and Stephen~G. {MacDonell}.
\newblock A taxonomy of data quality challenges in empirical software
  engineering.
\newblock In \emph{2013 22nd Australian Software Engineering Conference}, pages
  97--106. {IEEE}, 2013.
\newblock ISBN 978-0-7695-4995-8.
\newblock \doi{10.1109/ASWEC.2013.21}.
\newblock URL \url{http://ieeexplore.ieee.org/document/6601297/}.

\bibitem[Branco et~al.(2016)Branco, Torgo, and Ribeiro]{branco_survey_2016}
Paula Branco, Luís Torgo, and Rita~P. Ribeiro.
\newblock A {Survey} of {Predictive} {Modeling} on {Imbalanced} {Domains}.
\newblock \emph{ACM Computing Surveys}, 49\penalty0 (2):\penalty0 31:1--31:50,
  August 2016.
\newblock ISSN 0360-0300.
\newblock \doi{10.1145/2907070}.

\bibitem[Breiman(2001)]{breiman_random_2001}
Leo Breiman.
\newblock Random forests.
\newblock \emph{Machine Learning}, 45\penalty0 (1):\penalty0 5--32, 2001.

\bibitem[Brown et~al.(2012)Brown, Pocock, Zhao, and
  Lujan]{brown_conditional_2012}
Gavin Brown, Adam Pocock, Ming-Jie Zhao, and Mikel Lujan.
\newblock Conditional {Likelihood} {Maximisation}: {A} {Unifying} {Framework}
  for {Information} {Theoretic} {Feature} {Selection}.
\newblock \emph{Journal of Machine Learning Research}, 13:\penalty0 27--66,
  2012.

\bibitem[Chandrashekar and Sahin(2014)]{chandrashekar_survey_2014}
Girish Chandrashekar and Ferat Sahin.
\newblock A survey on feature selection methods.
\newblock \emph{Computers \& Electrical Engineering}, 40\penalty0 (1):\penalty0
  16--28, January 2014.
\newblock \doi{10.1016/j.compeleceng.2013.11.024}.
\newblock URL
  \url{http://www.sciencedirect.com/science/article/pii/S0045790613003066}.

\bibitem[Darshan and Jaidhar(2018)]{darshan_performance_2018}
Shiva Darshan and C.~D. Jaidhar.
\newblock Performance {Evaluation} of {Filter}-based {Feature} {Selection}
  {Techniques} in {Classifying} {Portable} {Executable} {Files}.
\newblock \emph{Procedia Computer Science}, 125:\penalty0 346--356, January
  2018.
\newblock \doi{10.1016/j.procs.2017.12.046}.
\newblock URL
  \url{http://www.sciencedirect.com/science/article/pii/S1877050917328107}.

\bibitem[Dash and Liu(1997)]{dash_feature_1997}
M~Dash and H~Liu.
\newblock Feature {Selection} for {Classification}.
\newblock \emph{Intelligent Data Analysis}, 1:\penalty0 131--167, 1997.

\bibitem[Dash et~al.(2000)Dash, Liu, and Motoda]{dash_consistency_2000}
Manoranjan Dash, Huan Liu, and Hiroshi Motoda.
\newblock Consistency based feature selection.
\newblock In \emph{Knowledge {Discovery} and {Data} {Mining}. {Current}
  {Issues} and {New} {Applications}}, pages 98--109. Springer, 2000.
\newblock ISBN 3-540-67382-2.

\bibitem[Davenport and Harris(2007)]{davenport_competing_2007}
Thomas~H. Davenport and Jeanne~G. Harris.
\newblock \emph{Competing on Analytics: The New Science of Winning}.
\newblock Harvard Business Press, 2007.
\newblock ISBN 978-1-4221-0332-6.

\bibitem[Demšar(2006)]{demsar_statistical_2006}
J.~Demšar.
\newblock Statistical comparisons of classifiers over multiple data sets.
\newblock \emph{Journal of Machine Learning Research}, 7:\penalty0 1--30, 2006.
\newblock 05008.

\bibitem[Dietterich et~al.(1996)Dietterich, Kearns, and
  Mansour]{Dietterich_Applyingweaklearning_1996}
Tom Dietterich, Michael Kearns, and Yishay Mansour.
\newblock Applying the weak learning framework to understand and improve
  {C4}.5.
\newblock In \emph{Proceedings of the 13. {International} {Conference} on
  {Machine} {Learning}}, pages 96--104. Morgan Kaufmann, 1996.

\bibitem[Dougherty et~al.(2009)Dougherty, Hua, and
  Sima]{dougherty_performance_2009}
Edward~R Dougherty, Jianping Hua, and Chao Sima.
\newblock Performance of {Feature} {Selection} {Methods}.
\newblock \emph{Current Genomics}, 10\penalty0 (6):\penalty0 365--374,
  September 2009.
\newblock \doi{10.2174/138920209789177629}.
\newblock URL \url{http://www.ncbi.nlm.nih.gov/pmc/articles/PMC2766788/}.

\bibitem[Dougherty(2013)]{dougherty2013estimating}
Geoff Dougherty.
\newblock Estimating and comparing classifiers.
\newblock In \emph{Pattern Recognition and Classification}, pages 157--176.
  Springer, 2013.

\bibitem[Fawcett(2006)]{fawcett_introduction_2006}
T.~Fawcett.
\newblock An introduction to {ROC} analysis.
\newblock \emph{Pattern Recognition Letters}, 27\penalty0 (8):\penalty0
  861--874, 2006.

\bibitem[Feng and Shanthikumar(2018)]{feng_big_data_production_2017}
Qi~Feng and J.~George Shanthikumar.
\newblock How research in production and operations management may evolve in
  the era of big data.
\newblock \emph{Production and Operations Management}, 27\penalty0
  (9):\penalty0 1670--1684, 2018.
\newblock \doi{10.1111/poms.12836}.
\newblock URL \url{https://onlinelibrary.wiley.com/doi/abs/10.1111/poms.12836}.

\bibitem[Fernández-Delgado et~al.(2014)Fernández-Delgado, Cernadas, Barro,
  and Amorim]{fernandez-delgado_we_2014}
Manuel Fernández-Delgado, Eva Cernadas, Senén Barro, and Dinani Amorim.
\newblock Do we need hundreds of classifiers to solve real world classification
  problems?
\newblock \emph{Journal of Machine Learning Research}, 15\penalty0
  (1):\penalty0 3133--3181, 2014.

\bibitem[Fitzpatrick and Mues(2016)]{fitzpatrick_empirical_2016}
Trevor Fitzpatrick and Christophe Mues.
\newblock An empirical comparison of classification algorithms for mortgage
  default prediction: evidence from a distressed mortgage market.
\newblock \emph{European Journal of Operational Research}, 249\penalty0
  (2):\penalty0 427--439, 2016.
\newblock \doi{10.1016/j.ejor.2015.09.014}.
\newblock URL
  \url{https://linkinghub.elsevier.com/retrieve/pii/S0377221715008383}.

\bibitem[Fleuret(2004)]{fleuret_fast_2004}
François Fleuret.
\newblock Fast binary feature selection with conditional mutual information.
\newblock \emph{Journal of Machine Learning Research}, 5:\penalty0 1531--1555,
  2004.

\bibitem[Forman(2003)]{forman_extensive_2003}
George Forman.
\newblock An extensive empirical study of feature selection metrics for text
  classification.
\newblock \emph{Journal of Machine Learning Research}, 3\penalty0
  (Mar):\penalty0 1289--1305, 2003.

\bibitem[Gartner et~al.(2015)Gartner, Kolisch, Neill, and
  Padman]{gartner_machine_2015}
Daniel Gartner, Rainer Kolisch, Daniel~B. Neill, and Rema Padman.
\newblock Machine learning approaches for early {DRG} classification and
  resource allocation.
\newblock \emph{{INFORMS} Journal on Computing}, 27\penalty0 (4):\penalty0
  718--734, 2015.
\newblock \doi{10.1287/ijoc.2015.0655}.
\newblock URL
  \url{http://search.ebscohost.com/login.aspx?direct=true&AuthType=ip&db=bsu&AN=114699847&site=ehost-live}.

\bibitem[Gómez and Rojas(2016)]{gomez_empirical_2016}
David Gómez and Alfonso Rojas.
\newblock An {Empirical} {Overview} of the {No} {Free} {Lunch} {Theorem} and
  {Its} {Effect} on {Real}-{World} {Machine} {Learning} {Classification}.
\newblock \emph{Neural Computation}, 28\penalty0 (1):\penalty0 216--228,
  January 2016.
\newblock ISSN 0899-7667.
\newblock \doi{10.1162/NECO_a_00793}.
\newblock URL \url{https://doi.org/10.1162/NECO_a_00793}.

\bibitem[Guyon et~al.(2003)Guyon, {André}, and
  {Elisseeff}]{guyon_introduction_2003}
Isabelle Guyon, {André}, and {Elisseeff}.
\newblock An introduction to variable and feature selection.
\newblock \emph{Journal of Machine Learning Research}, 3:\penalty0 1157--1182,
  2003.

\bibitem[Hall and Holmes(2003)]{hall_benchmarking_2003}
M.A. Hall and G.~Holmes.
\newblock Benchmarking attribute selection techniques for discrete class data
  mining.
\newblock \emph{IEEE Transactions on Knowledge and Data Engineering},
  15\penalty0 (6):\penalty0 1437--1447, November 2003.
\newblock \doi{10.1109/TKDE.2003.1245283}.

\bibitem[Hall(1999)]{hall_correlation-based_1999}
Mark~A. Hall.
\newblock \emph{Correlation-based feature selection for machine learning}.
\newblock PhD thesis, The University of Waikato, Hamilton, New Zealand, April
  1999.

\bibitem[Haury et~al.(2011)Haury, Gestraud, and Vert]{haury_influence_2011}
Anne-Claire Haury, Pierre Gestraud, and Jean-Philippe Vert.
\newblock The {Influence} of {Feature} {Selection} {Methods} on {Accuracy},
  {Stability} and {Interpretability} of {Molecular} {Signatures}.
\newblock \emph{PLOS ONE}, 6\penalty0 (12):\penalty0 e28210, December 2011.
\newblock \doi{10.1371/journal.pone.0028210}.
\newblock URL
  \url{http://journals.plos.org/plosone/article?id=10.1371/journal.pone.0028210}.

\bibitem[Holte(1993)]{holte_very_1993}
Robert~C. Holte.
\newblock Very simple classification rules perform well on most commonly used
  datasets.
\newblock \emph{Machine Learning}, 11\penalty0 (1):\penalty0 63--90, 1993.
\newblock URL \url{http://link.springer.com/article/10.1023/A:1022631118932}.

\bibitem[Hopf(2019)]{hopf_predictive_2019}
Konstantin Hopf.
\newblock \emph{Predictive Analytics for Energy Efficiency and Energy
  Retailing}, volume~36 of \emph{Contributions of the Faculty Information
  Systems and Applied Computer Sciences of the Otto-Friedrich-University
  Bamberg}.
\newblock University of Bamberg, 1 edition, 2019.
\newblock ISBN 978-3-86309-669-4.
\newblock URL \url{https://doi.org/10.20378/irbo-54833}.

\bibitem[Hua et~al.(2009)Hua, Tembe, and Dougherty]{hua_performance_2009}
Jianping Hua, Waibhav~D. Tembe, and Edward~R. Dougherty.
\newblock Performance of feature-selection methods in the classification of
  high-dimension data.
\newblock \emph{Pattern Recognition}, 42\penalty0 (3):\penalty0 409--424, March
  2009.
\newblock \doi{10.1016/j.patcog.2008.08.001}.
\newblock URL
  \url{http://linkinghub.elsevier.com/retrieve/pii/S0031320308003142}.

\bibitem[Hunt et~al.(1966)Hunt, Marin, and Stone]{hunt_experiments_1966}
Earl~B Hunt, Janet Marin, and Philip~J Stone.
\newblock \emph{Experiments in induction.}
\newblock Academic Press, Oxford, 1966.

\bibitem[Inza et~al.(2004)Inza, Larrañaga, Blanco, and
  Cerrolaza]{inza_filter_2004}
Iñaki Inza, Pedro Larrañaga, Rosa Blanco, and Antonio~J. Cerrolaza.
\newblock Filter versus wrapper gene selection approaches in {DNA} microarray
  domains.
\newblock \emph{Artificial Intelligence in Medicine}, 31\penalty0 (2):\penalty0
  91--103, June 2004.
\newblock \doi{10.1016/j.artmed.2004.01.007}.
\newblock URL
  \url{http://www.sciencedirect.com/science/article/pii/S0933365704000193}.

\bibitem[Jain and Zongker(1997)]{jain_feature_1997}
A.~Jain and D.~Zongker.
\newblock Feature selection: evaluation, application, and small sample
  performance.
\newblock \emph{IEEE Transactions on Pattern Analysis and Machine
  Intelligence}, 19\penalty0 (2):\penalty0 153--158, 1997.

\bibitem[Kalousis et~al.(2007)Kalousis, Prados, and
  Hilario]{kalousis_stability_2007}
Alexandros Kalousis, Julien Prados, and Melanie Hilario.
\newblock Stability of {Feature} {Selection} {Algorithms}: {A} {Study} on
  {High}-dimensional {Spaces}.
\newblock \emph{Knowledge and Information Systems}, 12\penalty0 (1):\penalty0
  95--116, May 2007.
\newblock \doi{10.1007/s10115-006-0040-8}.

\bibitem[Kaur et~al.(2019)Kaur, Pannu, and Malhi]{kaur_systematic_2019}
Harsurinder Kaur, Husanbir~Singh Pannu, and Avleen~Kaur Malhi.
\newblock A systematic review on imbalanced data challenges in machine
  learning: Applications and solutions.
\newblock \emph{{ACM} Computing Surveys}, 52\penalty0 (4):\penalty0
  79:1--79:36, 2019.
\newblock \doi{10.1145/3343440}.
\newblock URL \url{https://doi.org/10.1145/3343440}.

\bibitem[Keogh and Mueen(2011)]{keogh_curse_2011}
Eamonn Keogh and Abdullah Mueen.
\newblock Curse of {Dimensionality}.
\newblock In Claude Sammut and Geoffrey~I. Webb, editors, \emph{Encyclopedia of
  {Machine} {Learning}}, pages 257--258. Springer, Boston, MA, 2011.
\newblock ISBN 978-0-387-30768-8.
\newblock \doi{10.1007/978-0-387-30164-8_192}.
\newblock URL
  \url{http://link.springer.com/referenceworkentry/10.1007/978-0-387-30164-8_192}.

\bibitem[Khoshgoftaar et~al.(2015)Khoshgoftaar, Gao, Chen, and
  Napolitano]{khoshgoftaar_comparing_2015}
Taghi~M. Khoshgoftaar, Kehan Gao, Ye~Chen, and Amri Napolitano.
\newblock Comparing {Feature} {Selection} {Techniques} for {Software} {Quality}
  {Estimation} {Using} {Data}-{Sampling}-{Based} {Boosting} {Algorithms}.
\newblock \emph{International Journal of Reliability, Quality and Safety
  Engineering}, 22\penalty0 (03):\penalty0 1550013, May 2015.
\newblock \doi{10.1142/S0218539315500138}.

\bibitem[Kira and Rendell(1992)]{kira_practical_1992}
Kenji Kira and Larry~A. Rendell.
\newblock A practical approach to feature selection.
\newblock In \emph{Proceedings of the ninth international workshop on {Machine}
  learning}, pages 249--256. Morgan Kaufmann, Aberdeen, Scotland, United
  Kingdom, 1992.

\bibitem[Kohavi(1995)]{kohavi_study_1995}
Ron Kohavi.
\newblock A study of cross-validation and bootstrap for accuracy estimation and
  model selection.
\newblock In \emph{The proceedings of the 14th International Conference in
  {AI}}, volume~14, pages 1137--1145. Morgan Kaufmann, 1995.

\bibitem[Kononenko(1994)]{kononenko_estimating_1994}
Igor Kononenko.
\newblock Estimating attributes: {Analysis} and extensions of {RELIEF}.
\newblock In \emph{Machine {Learning}: {ECML}-94}, pages 171--182, Berlin,
  Heidelberg, April 1994. Springer.
\newblock \doi{10.1007/3-540-57868-4_57}.
\newblock URL \url{https://link.springer.com/chapter/10.1007/3-540-57868-4_57}.

\bibitem[Kou et~al.(2020)Kou, Yang, Peng, Xiao, Chen, and
  Alsaadi]{kou_2020_evaluation}
Gang Kou, Pei Yang, Yi~Peng, Feng Xiao, Yang Chen, and Fawaz~E. Alsaadi.
\newblock Evaluation of feature selection methods for text classification with
  small datasets using multiple criteria decision-making methods.
\newblock \emph{Applied Soft Computing}, 86:\penalty0 105836, 2020.
\newblock ISSN 1568-4946.
\newblock \doi{10.1016/j.asoc.2019.105836}.

\bibitem[Kraus et~al.(2020)Kraus, Feuerriegel, and Oztekin]{kraus_deep_2020}
Mathias Kraus, Stefan Feuerriegel, and Asil Oztekin.
\newblock Deep learning in business analytics and operations research: Models,
  applications and managerial implications.
\newblock \emph{European Journal of Operational Research}, 281\penalty0
  (3):\penalty0 628--641, 2020.
\newblock \doi{10.1016/j.ejor.2019.09.018}.
\newblock URL
  \url{http://www.sciencedirect.com/science/article/pii/S0377221719307581}.

\bibitem[Kudo and Sklansky(2000)]{kudo_comparison_2000}
Mineichi Kudo and Jack Sklansky.
\newblock Comparison of {Algorithms} that {Select} {Features} for {Pattern}
  {Classifiers}.
\newblock \emph{Pattern Recognition}, 33\penalty0 (1):\penalty0 25--41, 2000.

\bibitem[Kukar et~al.(1999)Kukar, Kononenko, Groselj, Kralj, and
  Fettich]{kukar_analysing_1999}
M.~Kukar, I.~Kononenko, C.~Groselj, K.~Kralj, and J.~Fettich.
\newblock Analysing and improving the diagnosis of ischaemic heart disease with
  machine learning.
\newblock \emph{Artificial Intelligence in Medicine}, 16\penalty0 (1):\penalty0
  25--50, May 1999.

\bibitem[Kursa(2020)]{praznik}
Miron~B. Kursa.
\newblock \emph{praznik: Tools for Information-Based Feature Selection}, 2020.
\newblock URL \url{https://CRAN.R-project.org/package=praznik}.
\newblock R package version 8.0.0.

\bibitem[Lai et~al.(2006)Lai, Reinders, van't Veer, and
  Wessels]{lai_comparison_2006}
Carmen Lai, Marcel~JT Reinders, Laura~J van't Veer, and Lodewyk~FA Wessels.
\newblock A comparison of univariate and multivariate gene selection techniques
  for classification of cancer datasets.
\newblock \emph{BMC Bioinformatics}, 7\penalty0 (1):\penalty0 235, 2006.
\newblock \doi{10.1186/1471-2105-7-235}.
\newblock URL
  \url{http://bmcbioinformatics.biomedcentral.com/articles/10.1186/1471-2105-7-235}.

\bibitem[LeCun et~al.(2015)LeCun, Bengio, and Hinton]{lecun_deep_2015}
Yann LeCun, Yoshua Bengio, and Geoffrey Hinton.
\newblock Deep learning.
\newblock \emph{Nature}, 521\penalty0 (7553):\penalty0 436--444, May 2015.
\newblock \doi{10.1038/nature14539}.
\newblock URL \url{https://www.nature.com/articles/nature14539}.

\bibitem[Lee and Shin(2020)]{lee_machine_2020}
In~Lee and Yong~Jae Shin.
\newblock Machine learning for enterprises: Applications, algorithm selection,
  and challenges.
\newblock \emph{Business Horizons}, 63\penalty0 (2):\penalty0 157--170, 2020.
\newblock \doi{10.1016/j.bushor.2019.10.005}.
\newblock URL
  \url{http://www.sciencedirect.com/science/article/pii/S0007681319301521}.

\bibitem[L'Heureux et~al.(2017)L'Heureux, Grolinger, Elyamany, and
  Capretz]{lheureux_machine_2017}
Alexandra L'Heureux, Katarina Grolinger, Hany~F. Elyamany, and Miriam A.~M.
  Capretz.
\newblock Machine {Learning} {With} {Big} {Data}: {Challenges} and
  {Approaches}.
\newblock \emph{IEEE Access}, 5:\penalty0 7776--7797, 2017.
\newblock \doi{10.1109/ACCESS.2017.2696365}.
\newblock URL \url{http://ieeexplore.ieee.org/document/7906512/}.

\bibitem[Li et~al.(2017)Li, Cheng, Wang, Morstatter, Trevino, Tang, and
  Liu]{li_feature_2017}
Jundong Li, Kewei Cheng, Suhang Wang, Fred Morstatter, Robert~P. Trevino,
  Jiliang Tang, and Huan Liu.
\newblock Feature {Selection}: {A} {Data} {Perspective}.
\newblock \emph{ACM Computing Surveys}, 50\penalty0 (6):\penalty0 94:1--94:45,
  December 2017.
\newblock ISSN 0360-0300.
\newblock \doi{10.1145/3136625}.

\bibitem[Liaw and Wiener(2002)]{randomForest}
Andy Liaw and Matthew Wiener.
\newblock Classification and regression by randomforest.
\newblock \emph{R News}, 2\penalty0 (3):\penalty0 18--22, 2002.
\newblock URL \url{https://CRAN.R-project.org/doc/Rnews/}.

\bibitem[Liu and Setiono(1995)]{liu_chi2_1995}
Huan Liu and Rudy Setiono.
\newblock Chi2: {Feature} selection and discretization of numeric attributes.
\newblock In \emph{{TAI} '95 {Proceedings} of the {Seventh} {International}
  {Conference} on {Tools} with {Artificial} {Intelligence}}, page~88. IEEE,
  1995.
\newblock ISBN 0-8186-7312-5.

\bibitem[Liu et~al.(2010)Liu, Motoda, Setiono, and Zhao]{liu_feature_2010}
Huan Liu, Hiroshi Motoda, Rudy Setiono, and Zheng Zhao.
\newblock Feature {Selection}: {An} {Ever} {Evolving} {Frontier} in {Data}
  {Mining}.
\newblock In \emph{Proceedings of the {Fourth} {International} {Workshop} on
  {Feature} {Selection} in {Data} {Mining}, {Journal} of {Machine} {Learning}
  {Research}}, volume~10, pages 4--13, Hyderabad, India, 2010.
\newblock URL \url{http://www.jmlr.org/proceedings/papers/v10/liu10b/liu10b}.

\bibitem[Liu et~al.(2002)Liu, Li, and Wong]{liu_comparative_2002}
Huiqing Liu, Jinyan Li, and Limsoon Wong.
\newblock A {Comparative} {Study} on {Feature} {Selection} and {Classification}
  {Methods} {Using} {Gene} {Expression} {Profiles} and {Proteomic} {Patterns}.
\newblock \emph{Genome Informatics}, 13:\penalty0 51--60, 2002.
\newblock \doi{10.11234/gi1990.13.51}.

\bibitem[Liu et~al.(2018)Liu, Xu, Luo, Xu, Wen, and
  Tao]{liu_cost-sensitive_2018}
Meng Liu, Chang Xu, Yong Luo, Chao Xu, Yonggang Wen, and Dacheng Tao.
\newblock Cost-{Sensitive} {Feature} {Selection} by {Optimizing}
  {F}-{Measures}.
\newblock \emph{IEEE Transactions on Image Processing}, 27\penalty0
  (3):\penalty0 1323--1335, March 2018.
\newblock \doi{10.1109/TIP.2017.2781298}.
\newblock URL \url{https://ieeexplore.ieee.org/document/8170306/}.

\bibitem[Marcus(2018)]{marcus2018deep}
Gary Marcus.
\newblock Deep learning: A critical appraisal, 2018.

\bibitem[Meyer et~al.(2019)Meyer, Dimitriadou, Hornik, Weingessel, and
  Leisch]{e1071}
David Meyer, Evgenia Dimitriadou, Kurt Hornik, Andreas Weingessel, and
  Friedrich Leisch.
\newblock \emph{e1071: Misc Functions of the Department of Statistics,
  Probability Theory Group (Formerly: E1071), TU Wien}, 2019.
\newblock URL \url{https://CRAN.R-project.org/package=e1071}.
\newblock R package version 1.7-3.

\bibitem[Meyer and Bontempi(2006)]{meyer_use_2006}
Patrick~E. Meyer and Gianluca Bontempi.
\newblock On the use of variable complementarity for feature selection in
  cancer classification.
\newblock In Franz Rothlauf, Jürgen Branke, Stefano Cagnoni, Ernesto Costa,
  Carlos Cotta, Rolf Drechsler, Evelyne Lutton, Penousal Machado, Jason~H.
  Moore, Juan Romero, George~D. Smith, Giovanni Squillero, and Hideyuki Takagi,
  editors, \emph{Applications of Evolutionary Computing}, Lecture Notes in
  Computer Science, pages 91--102. Springer, 2006.
\newblock ISBN 978-3-540-33238-1.
\newblock \doi{10.1007/11732242_9}.

\bibitem[Meyer et~al.(2008)Meyer, Schretter, and
  Bontempi]{meyer_informationtheoretic_2008}
Patrick~Emmanuel Meyer, Colas Schretter, and Gianluca Bontempi.
\newblock Information-{Theoretic} {Feature} {Selection} in {Microarray} {Data}
  {Using} {Variable} {Complementarity}.
\newblock \emph{IEEE Journal of Selected Topics in Signal Processing},
  2\penalty0 (3):\penalty0 261--274, June 2008.
\newblock \doi{10.1109/JSTSP.2008.923858}.
\newblock URL \url{http://ieeexplore.ieee.org/document/4550559/}.

\bibitem[Müller et~al.(2018)Müller, Fay, and Brocke]{muller_effect_2018}
Oliver Müller, Maria Fay, and Jan~vom Brocke.
\newblock The effect of big data and analytics on firm performance: An
  econometric analysis considering industry characteristics.
\newblock \emph{Journal of Management Information Systems}, 35\penalty0
  (2):\penalty0 488--509, 2018.
\newblock \doi{10.1080/07421222.2018.1451955}.

\bibitem[Nogueira et~al.(2018)Nogueira, Sechidis, and
  Brown]{nogueira_stability_2018}
Sarah Nogueira, Konstantinos Sechidis, and Gavin Brown.
\newblock On the stability of feature selection algorithms.
\newblock \emph{Journal of Machine Learning Research}, 18:\penalty0 1--54,
  2018.

\bibitem[Pedregosa et~al.(2011)Pedregosa, Varoquaux, Gramfort, Michel, Thirion,
  Grisel, Blondel, Prettenhofer, Weiss, Dubourg, and
  {others}]{pedregosa_scikit-learn_2011}
Fabian Pedregosa, Gaël Varoquaux, Alexandre Gramfort, Vincent Michel, Bertrand
  Thirion, Olivier Grisel, Mathieu Blondel, Peter Prettenhofer, Ron Weiss,
  Vincent Dubourg, and {others}.
\newblock Scikit-learn.
\newblock \emph{Journal of Machine Learning Research}, 12\penalty0
  (Oct):\penalty0 2825--2830, 2011.

\bibitem[Peng et~al.(2005)Peng, Long, and Ding]{peng_feature_2005}
Hanchuan Peng, Fuhui Long, and C.~Ding.
\newblock Feature selection based on mutual information criteria of
  max-dependency, max-relevance, and min-redundancy.
\newblock \emph{IEEE Transactions on Pattern Analysis and Machine
  Intelligence}, 27\penalty0 (8):\penalty0 1226--1238, August 2005.
\newblock \doi{10.1109/TPAMI.2005.159}.

\bibitem[Quinlan(1986)]{quinlan_induction_1986}
J.~R. Quinlan.
\newblock Induction of decision trees.
\newblock \emph{Machine Learning}, 1\penalty0 (1):\penalty0 81--106, 1986.
\newblock \doi{10.1007/BF00116251}.
\newblock URL \url{https://link.springer.com/article/10.1007/BF00116251}.

\bibitem[Rawlings et~al.(1998)Rawlings, Pantula, and
  Dickey]{rawlings_applied_1998}
John~O. Rawlings, Sastry~G. Pantula, and David~A. Dickey.
\newblock \emph{Applied regression analysis: a research tool}.
\newblock Springer texts in statistics. Springer, New York, 2nd ed edition,
  1998.
\newblock ISBN 978-0-387-98454-4.

\bibitem[Reunanen(2003)]{reunanen_overfitting_2003}
Juha Reunanen.
\newblock Overfitting in {Making} {Comparisons} {Between} {Variable}
  {Selection} {Methods}.
\newblock \emph{Journal of Machine Learning Research}, 3:\penalty0 1371--1382,
  March 2003.
\newblock URL \url{http://dl.acm.org/citation.cfm?id=944919.944978}.

\bibitem[Robnik-Šikonja(2003)]{robnik-sikonja_experiments_2003}
Marko Robnik-Šikonja.
\newblock Experiments with {Cost}-{Sensitive} {Feature} {Evaluation}.
\newblock In \emph{Machine {Learning}: {ECML} 2003}, pages 325--336, Berlin,
  Heidelberg, September 2003. Springer.
\newblock \doi{10.1007/978-3-540-39857-8_30}.
\newblock URL
  \url{https://link.springer.com/chapter/10.1007/978-3-540-39857-8_30}.

\bibitem[Robnik-Šikonja and Kononenko(2003)]{robnik-sikonja_theoretical_2003}
Marko Robnik-Šikonja and Igor Kononenko.
\newblock Theoretical and {Empirical} {Analysis} of {ReliefF} and {RReliefF}.
\newblock \emph{Machine Learning}, 53:\penalty0 23--69, 2003.
\newblock \doi{10.1023/A:1025667309714}.

\bibitem[Robnik-Sikonja and Savicky(2020)]{corelearn}
Marko Robnik-Sikonja and Petr Savicky.
\newblock \emph{CORElearn: Classification, Regression and Feature Evaluation},
  2020.
\newblock URL \url{https://CRAN.R-project.org/package=CORElearn}.
\newblock R package version 1.54.2.

\bibitem[Romanski and Kotthoff(2018)]{fselector}
Piotr Romanski and Lars Kotthoff.
\newblock \emph{FSelector: Selecting Attributes}, 2018.
\newblock URL \url{https://CRAN.R-project.org/package=FSelector}.
\newblock R package version 0.31.

\bibitem[Roy et~al.(2019)Roy, Qureshi, Pande, Nair, Gairola, Jain, Singh,
  Sharma, Jagadale, Lin, Sharma, Gotety, Zhang, Tang, Mehta, Sindhanuru,
  Okafor, Das, Gopal, Rudraraju, and Kakarlapudi]{roy_performance_2019}
Asim Roy, Shiban Qureshi, Kartikeya Pande, Divitha Nair, Kartik Gairola, Pooja
  Jain, Suraj Singh, Kirti Sharma, Akshay Jagadale, Yi-Yang Lin, Shashank
  Sharma, Ramya Gotety, Yuexin Zhang, Ji~Tang, Tejas Mehta, Hemanth Sindhanuru,
  Nonso Okafor, Santak Das, Chidambara~N. Gopal, Srinivasa~B. Rudraraju, and
  Avinash~V. Kakarlapudi.
\newblock Performance comparison of machine learning platforms.
\newblock \emph{{INFORMS} Journal on Computing}, 31\penalty0 (2):\penalty0
  207--225, 2019.
\newblock \doi{10.1287/ijoc.2018.0825}.
\newblock URL \url{https://pubsonline.informs.org/doi/10.1287/ijoc.2018.0825}.

\bibitem[Russell and Norvig(2016)]{russell_artificial_2016}
Stuart~J. Russell and Peter Norvig.
\newblock \emph{Artificial Intelligence: A Modern Approach}.
\newblock Pearson education, 2016.

\bibitem[Saeys et~al.(2007)Saeys, Inza, and Larrañaga]{saeys_review_2007}
Yvan Saeys, Iñaki Inza, and Pedro Larrañaga.
\newblock A review of feature selection techniques in bioinformatics.
\newblock \emph{Bioinformatics}, 23\penalty0 (19):\penalty0 2507--2517, January
  2007.
\newblock \doi{10.1093/bioinformatics/btm344}.
\newblock URL
  \url{http://bioinformatics.oxfordjournals.org/content/23/19/2507}.

\bibitem[Saeys et~al.(2008)Saeys, Abeel, and van~de Peer]{saeys_robust_2008}
Yvan Saeys, Thomas Abeel, and Yves van~de Peer.
\newblock Robust {Feature} {Selection} {Using} {Ensemble} {Feature} {Selection}
  {Techniques}.
\newblock In \emph{Machine {Learning} and {Knowledge} {Discovery} in
  {Databases}}, pages 313--325, Berlin, Heidelberg, September 2008. Springer.
\newblock \doi{10.1007/978-3-540-87481-2_21}.
\newblock URL
  \url{https://link.springer.com/chapter/10.1007/978-3-540-87481-2_21}.

\bibitem[S{\'a}ez et~al.(2013)S{\'a}ez, Galar, Luengo, and
  Herrera]{saez2013tackling}
Jos{\'e}~A S{\'a}ez, Mikel Galar, Juli{\'a}N Luengo, and Francisco Herrera.
\newblock Tackling the problem of classification with noisy data using multiple
  classifier systems: Analysis of the performance and robustness.
\newblock \emph{Information Sciences}, 247:\penalty0 1--20, 2013.

\bibitem[Semwal et~al.(2017)Semwal, Mondal, and Nandi]{semwal_robust_2017}
Vijay~Bhaskar Semwal, Kaushik Mondal, and G.~C. Nandi.
\newblock Robust and accurate feature selection for humanoid push recovery and
  classification: deep learning approach.
\newblock \emph{Neural Computing and Applications}, 28\penalty0 (3):\penalty0
  565--574, March 2017.
\newblock ISSN 0941-0643, 1433-3058.
\newblock \doi{10.1007/s00521-015-2089-3}.

\bibitem[Shmueli and Koppius(2011)]{shmueli_predictive_2011}
Galit Shmueli and Otto~R. Koppius.
\newblock Predictive analytics in information systems research.
\newblock \emph{{MIS} Quarterly}, 35\penalty0 (3):\penalty0 553--572, 2011.

\bibitem[Sánchez-Maroño et~al.(2007)Sánchez-Maroño, Alonso-Betanzos, and
  Tombilla-Sanromán]{sanchez_filter_2007}
Noelia Sánchez-Maroño, Amparo Alonso-Betanzos, and María Tombilla-Sanromán.
\newblock Filter {Methods} for {Feature} {Selection} – {A} {Comparative}
  {Study}.
\newblock In Hujun Yin, Peter Tino, Emilio Corchado, Will Byrne, and Xin Yao,
  editors, \emph{Intelligent {Data} {Engineering} and {Automated} {Learning} -
  {IDEAL} 2007}, volume 4881, pages 178--187. Springer, Berlin, Heidelberg,
  2007.
\newblock ISBN 978-3-540-77225-5.
\newblock \doi{10.1007/978-3-540-77226-2_19}.
\newblock URL \url{http://link.springer.com/10.1007/978-3-540-77226-2_19}.

\bibitem[Vapnik and Vapnik(1998)]{vapnik_statistical_1998}
Vladimir~Naumovich Vapnik and Vlamimir Vapnik.
\newblock \emph{Statistical learning theory}, volume~1.
\newblock Wiley, New York, 1998.

\bibitem[Wah et~al.(2018)Wah, Ibrahim, Hamid, Abdul-Rahman, and
  Fong]{wah_feature_2018}
Yap~Bee Wah, Nurain Ibrahim, Hamzah~Abdul Hamid, Shuzlina Abdul-Rahman, and
  Simon Fong.
\newblock Feature {Selection} {Methods}: {Case} of {Filter} and {Wrapper}
  {Approaches} for {Maximising} {Classification} {Accuracy}.
\newblock \emph{Pertanika Journal of Science \& Technology}, 26\penalty0 (1),
  2018.

\bibitem[White(1980)]{white_heteroskedasticity-consistent_1980}
H.~White.
\newblock A {Heteroskedasticity}-{Consistent} {Covariance} {Matrix} and a
  {Direct} {Test} {for} {Heteroskedasticity}.
\newblock \emph{Econometrica}, \penalty0 (48):\penalty0 817--383, 1980.

\bibitem[Wolpert(1996)]{wolpert_lack_1996}
David~H. Wolpert.
\newblock The {Lack} of {A} {Priori} {Distinctions} {Between} {Learning}
  {Algorithms}.
\newblock \emph{Neural Computation}, 8\penalty0 (7):\penalty0 1341--1390,
  October 1996.
\newblock ISSN 0899-7667, 1530-888X.
\newblock \doi{10.1162/neco.1996.8.7.1341}.
\newblock URL
  \url{https://www.mitpressjournals.org/doi/abs/10.1162/neco.1996.8.7.1341}.

\bibitem[Wu et~al.(2019)Wu, Hitt, and Lou]{wu_data_2019}
Lynn Wu, Lorin Hitt, and Bowen Lou.
\newblock Data analytics, innovation, and firm productivity.
\newblock \emph{Management Science}, 66\penalty0 (5):\penalty0 2017--2039,
  2019.
\newblock \doi{10.1287/mnsc.2018.3281}.

\bibitem[Xue et~al.(2015)Xue, Zhang, and Browne]{xue_comprehensive_2015}
Bing Xue, Mengjie Zhang, and Will~N. Browne.
\newblock A {Comprehensive} {Comparison} on {Evolutionary} {Feature}
  {Selection} {Approaches} to {Classification}.
\newblock \emph{International Journal of Computational Intelligence and
  Applications}, 14\penalty0 (02):\penalty0 1550008, June 2015.
\newblock \doi{10.1142/S146902681550008X}.
\newblock URL
  \url{https://www.worldscientific.com/doi/abs/10.1142/S146902681550008X}.

\bibitem[Yang and Moody(2000)]{yang_data_2000}
Howard~Hua Yang and John Moody.
\newblock Data visualization and feature selection: New algorithms for
  nongaussian data.
\newblock In \emph{Advances in neural information processing systems}, pages
  687--693, 2000.
\newblock URL
  \url{http://papers.nips.cc/paper/1779-data-visualization-and-feature-selection-new-algorithms-for-nongaussian-data.pdf}.

\bibitem[Yu and Liu(2003)]{yu_feature_2003}
Lei Yu and Huan Liu.
\newblock Feature selection for high-dimensional data: {A} fast
  correlation-based filter solution.
\newblock In \emph{{ICML}}, volume~3, pages 856--863, 2003.

\bibitem[Zeileis(2004)]{zeileis_econometric_2004}
Achim Zeileis.
\newblock Econometric {Computing} with {HC} and {HAC} {Covariance} {Matrix}
  {Estimators}.
\newblock \emph{Journal of Statistical Software}, 11\penalty0 (1):\penalty0
  1--17, November 2004.
\newblock \doi{10.18637/jss.v011.i10}.
\newblock URL
  \url{https://www.jstatsoft.org/index.php/jss/article/view/v011i10}.
\newblock Number: 1.

\bibitem[Zhu and Wu(2004)]{zhu_class_2004}
Xingquan Zhu and Xindong Wu.
\newblock Class noise vs. attribute noise: A quantitative study.
\newblock \emph{Artificial Intelligence Review}, 22\penalty0 (3):\penalty0
  177--210, 2004.
\newblock \doi{10.1007/s10462-004-0751-8}.
\newblock URL \url{http://link.springer.com/article/10.1007/s10462-004-0751-8}.

\end{thebibliography}
